\pdfoutput=1

\documentclass[11pt]{article}

\usepackage[final]{acl}

\usepackage{times}
\usepackage{latexsym}

\usepackage[T1]{fontenc}

\usepackage[utf8]{inputenc}

\usepackage{microtype}

\usepackage{inconsolata}

\usepackage{graphicx}
\usepackage{float}
\usepackage{dsfont}
\usepackage{tabularx}

\usepackage{enumitem}
\usepackage{rotating}

\usepackage{booktabs}
\usepackage{multirow}
\usepackage{tcolorbox}
\usepackage[table]{xcolor}
\usepackage{array}
\usepackage{subcaption}
\usepackage{pifont}
\usepackage{arydshln}

\usepackage{amsthm}
\usepackage{amsmath}
\usepackage{amssymb}
\usepackage{xspace}
\newcommand{\narrowtextsc}[1]{\textls[-50]{\textsc{#1}}}
\newcommand{\lm}[1]{\texttt{#1}}
\newcommand{\sys}[1]{\narrowtextsc{#1}}
\newcommand{\data}[1]{\textsf{#1}}

\newcommand{\our}{\sys{Macro}\xspace}

\usepackage[table,dvipsnames]{xcolor}
\usepackage{colortbl}

\usepackage{soul}
\definecolor{lightblue}{RGB}{239,247,250}
\definecolor{blue}{RGB}{224,230,255}
\definecolor{lighpurple}{RGB}{226, 218, 246}
\definecolor{purple}{RGB}{193, 175, 236}
\definecolor{deepred}{RGB}{183,26,26}
\definecolor{deepgreen}{RGB}{4,98,10}
\definecolor{lightred}{RGB}{242,207, 194}

\usepackage[normalem]{ulem}

\definecolor{methodblue}{RGB}{230,240,250}
\definecolor{methodgreen}{RGB}{235,245,235}
\definecolor{methodorange}{RGB}{252,239,224}

\definecolor{BestSLFR}{RGB}{217,237,255} 
\definecolor{BestPPL}{RGB}{230,216,242}   
\definecolor{BestSS}{RGB}{255,231,214}  

\definecolor{AvgGray}{HTML}{F2F2F2}

\definecolor{bestCell}{HTML}{DCD3F2}   
\definecolor{secondCell}{HTML}{ECECEC} 

\definecolor{sibColor}{HTML}{E8F4F8}   
\definecolor{taxiColor}{HTML}{EEF8E8}  

\newcommand{\bestresult}[1]{\cellcolor{bestCell}#1}
\newcommand{\secondresult}[1]{\cellcolor{secondCell}#1}

\newcommand{\bestlegend}{\colorbox{bestCell}{\textbf{best}}}
\newcommand{\secondlegend}{\colorbox{secondCell}{second-best}}

\newcommand{\Up}{\ensuremath{\uparrow}}
\newcommand{\Down}{\ensuremath{\downarrow}}
\newcommand{\Same}{\ensuremath{\rightarrow}}

\newcommand{\dneg}[1]{\textcolor{LimeGreen}{\Down\,#1}}
\newcommand{\dpos}[1]{\textcolor{WildStrawberry}{\Up\,#1}}
\newcommand{\dzer}[1]{\textcolor{RoyalBlue}{\Same\,#1}}

\newcommand{\cell}[3]{%
  #1\,(%
  \ifx#3+\dpos{#2}\else%
    \ifx#3-\dneg{#2}\else%
      \dzer{#2}%
    \fi%
  \fi%
  )%
}

\usepackage{fontawesome5}       

\colorlet{FlipCol}{RubineRed}
\colorlet{EditCol}{BurntOrange}
\colorlet{AugCol}{RoyalPurple}

\newcommand{\Rflip}{$\mathcal{R}_{\text{flip}}$\xspace}
\newcommand{\Redit}{$\mathcal{R}_{\text{edit}}$\xspace}
\newcommand{\Raug}{$\mathcal{R}_{\text{aug}}$\xspace}

\usepackage{verbatim}

\title{\our: Enhancing Multilingual Counterfactual Explanations through Alignment-as-Preference Optimization}

\newcommand{\affilsup}[1]{\rlap{\textsuperscript{\normalfont#1}}}

\author{
    Yilong Wang\affilsup{1,\footnotemark[1]}
    \qquad
    Qianli Wang\affilsup{1,2,\footnotemark[1],\footnotemark[2]}
    \qquad
    \textbf{Bohao Chu}\affilsup{3}
    \\
    \textbf{Yihong Liu}\affilsup{4,5}
    \qquad
    \textbf{Jing Yang}\affilsup{1,7}
    \qquad
    \textbf{Simon Ostermann}\affilsup{2,6,8}
    \\
    $^1$Technische Universit\"at Berlin
    \quad
    $^2$German Research Center for Artificial Intelligence (DFKI)
    \\
    $^3$University of Duisburg-Essen
    \quad
    $^4$LMU Munich
    \quad
    $^5$Munich Center for Machine Learning (MCML)
    \\
    $^6$Saarland Informatics Campus
    \quad
    $^7$BIFOLD – Berlin Institute for the Foundations of Learning and Data
    \\
    $^8$Centre for European Research in Trusted AI (CERTAIN)\\
    \small{\footnotemark[2]\textbf{Correspondence:} \texttt{\href{mailto:qianli.wang@tu-berlin.de}{qianli.wang@tu-berlin.de}}
  }
}

\begin{document}
\maketitle

\renewcommand{\thefootnote}{\fnsymbol{footnote}}
\footnotetext[1]{Equal contribution and shared first authorship.}
\renewcommand*{\thefootnote}{\arabic{footnote}}

\begin{abstract}
Self-generated counterfactual explanations (SCEs) are minimally modified inputs (\textit{minimality}) generated by large language models (LLMs) that flip their own predictions (\textit{validity}), offering a causally grounded approach to unraveling black-box LLM behavior. Yet extending them beyond English remains challenging: existing methods struggle to produce valid SCEs in non-dominant languages, and a persistent trade-off between \textit{validity} and \textit{minimality} undermines explanation quality. We introduce \our\footnote{\url{https://github.com/YilongWangBerlin/MACRO}}, a preference alignment framework that applies Direct Preference Optimization (DPO) to multilingual SCE generation, using a composite scoring function to construct preference pairs that effectively translate the trade-off into measurable preference signals. Experiments across four LLMs and seven typologically diverse languages show that \our improves \textit{validity} by 12.55\% on average over the chain-of-thought baseline without degrading \textit{minimality}, while avoiding the severe \textit{minimality} violations of the translation-based baseline. Compared to supervised fine-tuning, \our achieves superior performance on both metrics, confirming that explicit preference optimization is essential for balancing this trade-off. Further analyses reveal that \our increases cross-lingual perturbation alignment and mitigates common generation errors. Our results highlight preference optimization as a promising direction for enhancing multilingual model explanations.


\end{abstract}

\begin{figure*}[t]
  \centering
  \includegraphics[width=\textwidth]{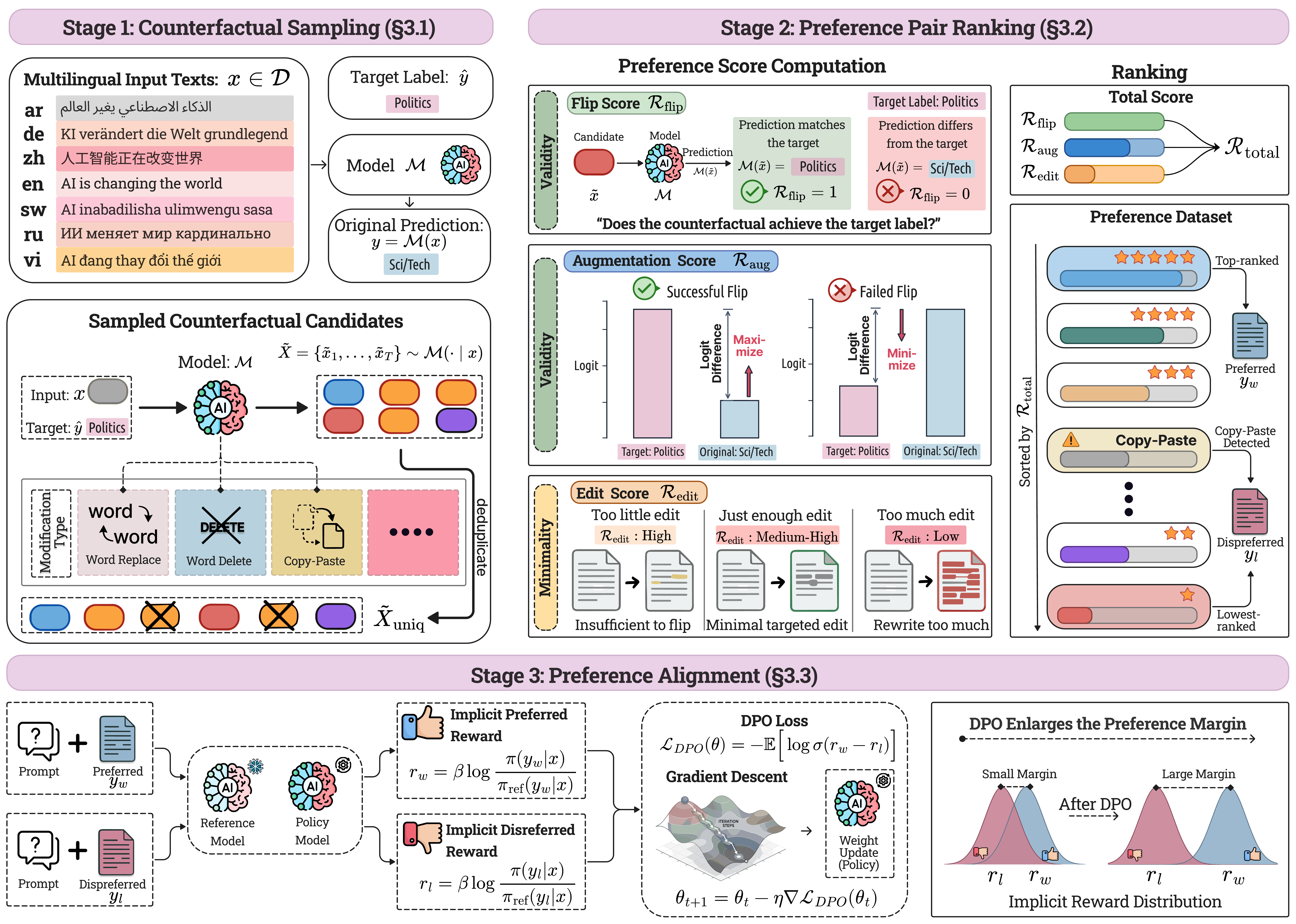}
  \caption{ Overview of our three-stage framework (\our). \textbf{Stage 1} samples counterfactual candidates via word-level perturbations across multilingual inputs.\textbf{ Stage 2} ranks candidates using $\mathcal{R}_{\text{flip}}$, $\mathcal{R}_{\text{aug}}$, and $\mathcal{R}_{\text{edit}}$ to construct preference pairs. \textbf{Stage 3} applies DPO to align the model toward generating minimal, effective counterfactuals.}
  \label{fig:pipeline}
\end{figure*}

\section{Introduction}
The necessity of multilingual explanations for interpreting black-box LLM behavior has recently been recognized \cite{resck-etal-2025-explainability, kastrati-etal-2025-multilingual, zhao-etal-2025-multilingual, wang-etal-2025-multilingual-datasets},
driven by both empirical evidence of systematic behavioral differences between English and non-English contexts \cite{lai-etal-2023-chatgpt} and equitable access requirements for non-English-speaking populations \cite{steigerwald-etal-2022-equal}. Among various self-explanations, SCEs offer a causally grounded probe of model behavior: given an input, an LLM generates a minimally modified version (\textit{minimality}) that changes its own prediction (\textit{validity}) \cite{ross-etal-2021-explaining, bhan-et-al-2023-tictec, zhao-etal-2024-survey}. However, extending SCEs to multilingual settings poses two key difficulties: \ding{192} existing methods largely fail to generate valid SCEs in non-dominant languages \cite{wang-etal-2026-parallel}; \ding{193} a trade-off between \textit{validity} and \textit{minimality} may persist \cite{mayne-etal-2025-llms}.\footnote{SCEs could be more valid and effective when more modifications are applied. However, deviating too far from the original inputs makes SCEs unreliable as model explanations.} 
We argue that preference optimization (PO) offers a natural solution to these challenges, as it can directly align model outputs toward \textit{validity} while penalizing excessive edits to preserve \textit{minimality}.
Despite this potential, applying PO to SCE generation is non-trivial, primarily due to the absence of high-quality multilingual preference data and the difficulty of mathematically formulating the trade-off into measurable preference signals.

To this end, we introduce \our, an automated preference alignment framework that leverages DPO \cite{rafailov-etal-2023-dpo} to optimize multilingual SCE generation (Figure~\ref{fig:pipeline}), turning the need for high \textit{validity} and \textit{minimality} directly into scoring components (\S\ref{subsec:preference_ranking}) that are subsequently used for preference pair construction. Our experiments across four LLMs and seven typologically diverse languages demonstrate that \our effectively mitigates the \textit{validity}–\textit{minimality} trade-off that persists in prior approaches \cite{mayne-etal-2025-llms}. Our main contributions are as follows:

\textbf{First, we show that \our consistently improves \textit{validity}} over three representative baselines: DG-CF, TB-CF \cite{wang-etal-2026-parallel}, and supervised fine-tuning (SFT). This improvement is achieved without degrading \textit{minimality}, marking a pronounced distinction from the translation-based approach TB-CF and SFT baselines.


\textbf{Second, we confirm that each component in \our plays a distinct and complementary role in balancing \textit{validity} and \textit{minimality}} by conducting systematic ablation studies on three preference scoring components: flip score ($\mathcal{R}_{\text{flip}}$), augmentation score ($\mathcal{R}_{\text{aug}}$), and edit score ($\mathcal{R}_{\text{edit}}$). 


\textbf{Third, we find that \our improves the alignment of cross-lingual input perturbation strategies} across all language pairs, by examining whether SCEs across different languages rely on similar types of modifications to change the semantics of the original input.

\textbf{Lastly, we show that two common issues, \textit{copy-paste} and \textit{language confusion}, can be largely mitigated by \our.} This effect is especially pronounced for lower-resource languages.


\section{Background and Related Work}

\paragraph{Multilingual Counterfactuals.} \citet{liu-etal-2021-counterfactual} enhance the machine translation performance via multilingual counterfactual data augmentation. \citet{barriere-cifuentes-2024-text, barriere-cifuentes-2024-study} evaluate nationality bias in diverse languages using multilingual counterfactual examples. \citet{roy-etal-2025-rag} utilize multilingual counterfactuals to perform answer attribution in multilingual retrieval-augmented generation systems. Nonetheless, the counterfactuals used in the aforementioned work differ from the counterfactual explanations investigated in this paper: while the former flip ground-truth labels, the latter aim to alter the model's predictions. \citet{wang-etal-2026-parallel} introduce the first multilingual counterfactual explanation benchmark and propose two CoT-based approaches \cite{wei-etal-2022-cot}, DG-CF and TB-CF; however, both methods exhibit limited effectiveness, particularly for low-resource languages. Consequently, we explore the potential of PO to enhance multilingual SCE generation and largely mitigate the \textit{validity}-\textit{minimality} trade-off.

\paragraph{Preference Optimization for Explainability.}
Preference optimization methods have recently been employed to guide LLMs toward generating superior outputs compared to inferior alternatives across various tasks \cite{rafailov-etal-2023-dpo, shao2024deepseekmathpushinglimitsmathematical, ethayarajh-etal-2024-kto, liu2026gdpogrouprewarddecouplednormalization}. Among these approaches, Direct Preference Optimization \cite{rafailov-etal-2023-dpo} has gained widespread adoption for LLM alignment, owing to its simplicity. DPO aligns model outputs with human preferences by utilizing both preferred responses ($y_w$) and dispreferred responses ($y_l$), with the objective function formulated as:
\begin{align}
\begin{split}
    &\mathcal{L}_{DPO}(\theta) = -\mathbb{E}_{(x,y_w,y_l)\sim \mathcal{D}_{\text{DPO}}}\Bigr[\log \sigma\big( \\
    & \beta\log\frac{\pi(y_w|x)}{\pi_{\text{ref}}(y_w|x)}-\beta\log\frac{\pi(y_l|x)}{\pi_{\text{ref}}(y_l|x)}\big)\Bigr]
\end{split}
\label{eq:dpo_loss}
\end{align}
where $\sigma(\cdot)$ represents the sigmoid function and $\beta$ denotes the weighting factor of the Kullback-Leibler divergence \cite{kullback1951information}, which constrains the policy $\pi_{\theta}$ to remain close to the reference policy $\pi_{\text{ref}}$.\footnote{More information about DPO is detailed in Appendix~\ref{app:dpo}.} While DPO has proven effective in enhancing natural language explanation quality \cite{villaarenas2024anchoredalignmentselfexplanationsenhancement, trivedi2024selfrationalizationimprovesllmfinegrained, bansal2026fragmentsfactscurriculumdrivendpo, sun2026investigatinginterplaycontextualparametric}, its use in (multilingual) counterfactual generation is still unexplored. Counterfactual generation demands objective measures of both \textit{minimality} and \textit{validity}, which may involve a trade-off between one another \cite{mayne-etal-2025-llms}. This makes it more challenging than other explainability methods.

\section{Methodology: \our}
\label{sec:marco}
Given a model $\mathcal{M}$ and an input text $x^{(\ell)} \in \mathcal{D}$ predicted as $y^{(\ell)} = \mathcal{M}(x^{(\ell)})$, where $\mathcal{D}$ is a multilingual dataset with languages $\ell \in \mathcal{L}$ and parallel data, our objective is to construct a counterfactual $\tilde{x}^{(\ell)}$ that causes the model to output the target label $\hat{y}^{(\ell)}$,\footnote{The target label is sampled uniformly among labels differing from both the gold label and the model prediction. Appendix~\ref{app:transition} analyzes how difficulty varies across transitions.} while maintaining similarity to $x^{(\ell)}$. 
To address this, we propose \our (Figure~\ref{fig:pipeline}), which applies DPO to multilingual SCE generation, treating \textit{validity} as the preferred outcome while penalizing excessive edits to preserve \textit{minimality}. \our operates in three stages: counterfactual sampling, preference pair ranking, and preference alignment.

\subsection{Counterfactual Sampling}
\label{subsec:counterfactual_sampling}
In the first step, for each data point $x^{(\ell)} \in \mathcal{D}$ across all investigated languages $\ell \in \mathcal{L}$, the model $\mathcal{M}$ samples a set of $T$ diverse counterfactuals, denoted $\tilde{X}^{(\ell)}=\{\tilde{x}_1^{(\ell)},...,\tilde{x}_T^{(\ell)}\}$, by applying the CoT-based approach DG-CF \cite{wang-etal-2026-parallel} to perturb the original input $x^{(\ell)}$ stepwise (for more details, see Section~\ref{subsec:baseline}; the prompt is shown in Figure~\ref{fig:prompt_cf}): 
\begin{align}
    \tilde{X}^{(\ell)} \sim \mathcal{M}(\cdot \mid x^{(\ell)})
\end{align}

\subsection{Preference Pair Ranking}
\label{subsec:preference_ranking}
After sampling, we deduplicate the sampled counterfactuals $\tilde{X}^{(\ell)}$, as we find that LLMs often produce identical outputs under stochastic decoding, yielding $\tilde{X}_{\mathrm{uniq}}^{(\ell)}$. We then compute the scores for each candidate and rank them accordingly. The score comprises three components: $\mathcal{R}_{\text{flip}}$, $\mathcal{R}_{\text{aug}}$, and $\mathcal{R}_{\text{edit}}$, covering the \textit{validity} and \textit{minimality} perspectives. $\mathcal{R}_{\text{flip}}$ provides the
primary binary signal driving label change (\S\ref{subsubsec:flip}), $\mathcal{R}_{\text{aug}}$ refines this signal with continuous logit-margin
feedback to consolidate and strengthen flips (\S\ref{subsubsec:aug}), and $\mathcal{R}_{\text{edit}}$ constrains the edit path to ensure that
validity is achieved through minimal modifications (\S\ref{subsubsec:edit}).
Only $\mathcal{R}_{\text{aug}}$ requires logit access, so \our also applies to black-box models (Appendix~\ref{app:blackbox}).

\subsubsection{Flip Score (\Rflip)}
\label{subsubsec:flip}
One of the core objectives is to ensure that the counterfactual $\tilde{x}$ effectively shifts the model's output from the original prediction $y$ to the desired target label $\hat{y}$ (\textit{validity}). We therefore design $\mathcal{R}_{\text{flip}}$ as a sparse binary score:
\begin{equation}
    \mathcal{R}_{\text{flip}}(\tilde{x},\hat{y}) = \mathbb{I}[\mathcal{M}(\tilde{x}) = \hat{y}] 
\end{equation}
where $\mathbb{I}(\cdot)$ is the indicator function, returning 1 when the condition is satisfied and 0 otherwise.

\subsubsection{Augmentation Score (\Raug)}
\label{subsubsec:aug}
For \textit{validity}, we further introduce an augmentation score to encourage the generated counterfactual either
(i) to separate the target label $\hat{y}$ from the original label $y$ by a large margin when the label change succeeds, or (ii) to reduce the model's confidence in the original label when the label change fails, making the candidate closer to a successful label change. The logit vector $\mathbf{z}(x)\in\mathbb{R}^{|V|}$ is the output of the model $\mathcal{M}$ for an input text $x$, where $|V|$ represents the vocabulary size and $z_c(x)$ denotes the logit of class $c$. 
\begin{equation}
\mathcal{R}_{\text{aug}}(x,\tilde{x};\hat{y},y)=
\begin{cases}
\sigma\big(z_{\hat{y}}(\tilde{x})-z_{y}(\tilde{x})\big), \\
\quad \text{if } \mathcal{M}(\tilde{x}) = \hat{y},\\
\sigma (z_{y}(x) - z_{y}(\tilde{x})), \\
\quad \text{otherwise.}
\end{cases}
\end{equation}

\subsubsection{Edit Score (\Redit)}
\label{subsubsec:edit}
The other objective beyond validity is to minimize modifications to the original input $x$. Following \citet{dehghanighobadi-etal-2025-llms, wang-etal-2025-fitcf, bhattacharjee-etal-2024-zero}, we employ a normalized Levenshtein distance $d$, which captures all applied edits, converted to a similarity score\footnote{Although Levenshtein distances are not mathematically comparable across languages due to differing script systems, they remain comparable within a given language.}:
\begin{align}
    \mathcal{R}_{\text{edit}}(x,\tilde{x}) = 1 - \frac{d(x,\tilde{x})}{|x|}
\end{align}

This will penalize $\tilde{x}$ with extensive modifications to the input $x$, which may even result in a negative edit score.

\subsubsection{Total Score ($\mathcal{R}_{\text{total}}$)}
\label{subsubsec:total_reward}
We compute a scalar total score by a weighted sum over available components:
\begin{equation}
    \mathcal{R}_{\text{total}} = \sum_{k} w_k \mathcal{R}_k, \quad w_k > 0
\end{equation}
where $k$ indexes applicable scores with weights $w_k$ and 
scores $\mathcal{R}_k \in \{ \mathcal{R}_{\text{flip}}, \mathcal{R}_{\text{aug}}, \mathcal{R}_{\text{edit}}\}$ (App.~\ref{app:reward_weight}).

\subsection{Preference Alignment}
\label{subsec:dataset_optimization}
We construct preference pairs only for examples where at least one sampled candidate changes the model prediction; otherwise, the example is discarded.\footnote{Appendix~\ref{app:retention} reports the retention rate.}
The preferred candidate $y_w^{(\ell)}$ and the dispreferred candidate $y_l^{(\ell)}$ for input $x^{(\ell)}$ are determined by $\mathcal{R}_{\text{total}}$:
\begin{align}
    y_w^{(\ell)} &= \mathop{\arg\max}_{\tilde{x}_j^{(\ell)} \in \tilde{X}_{\mathrm{uniq}}^{(\ell)}} \mathcal{R}_{\text{total}}(\tilde{x}_j^{(\ell)}) 
    \label{eq:yw_def} \\
    y_l^{(\ell)} &= \mathop{\arg\min}_{\tilde{x}_j^{(\ell)} \in \tilde{X}_{\mathrm{uniq}}^{(\ell)}} \mathcal{R}_{\text{total}}(\tilde{x}_j^{(\ell)})
\end{align}
Additionally, if any candidate $\tilde{x}^{(\ell)}$ exactly matches the original input $x^{(\ell)}$, we explicitly set $y_l^{(\ell)} = x^{(\ell)}$ while keeping $y_w^{(\ell)}$ as defined in Eq.~(\ref{eq:yw_def}), thereby discouraging the model from copying the input and facilitating meaningful edits. This yields the DPO training dataset for a given language $\ell$, $\mathcal{D}_{\text{DPO}}^{(\ell)} = \{(x^{(\ell)}, y_w^{(\ell)}, y_l^{(\ell)})_i\}_{i=1}^N$, of size $N$. 
Leveraging the multilingual DPO dataset $\mathcal{D}_{\text{DPO}}=\cup_{\ell \in \mathcal{L}} \mathcal{D}_{\text{DPO}}^{(\ell)}$, we optimize the model $\mathcal{M}$ with respect to $\mathcal{R}_{\text{total}}$, employing the loss function defined in Eq.~(\ref{eq:dpo_loss}).\footnote{Details about DPO training are presented in Appendix~\ref{app:training}.}

\section{Experimental Setup}

\subsection{Datasets}
\label{subsec:dataset}
Following prior work \cite{wang-etal-2026-parallel, bhattacharjee-etal-2024-zero, nguyen-etal-2024-llms} and owing to the inherent definition of counterfactual explanation (\S\ref{sec:marco}), we focus on two classification tasks and employ the following two large-scale multilingual datasets with sufficient language coverage accordingly for model preference alignment.\footnote{Dataset information is detailed in Appendix \ref{app:dataset}.}


\paragraph{\data{SIB200}} \cite{adelani-etal-2024-sib} is a dataset for \textit{topic classification} in 205 languages, designed to evaluate natural language understanding across diverse languages. Sentences are categorized into 7 classes: \textit{Science/Technology}, \textit{Travel}, \textit{Politics}, \textit{Sports}, \textit{Health}, \textit{Entertainment}, and \textit{Geography}.

\paragraph{\data{TAXI1500}} \cite{ma-etal-2025-taxi1500} is a massively multilingual \textit{sentence classification} dataset covering 1504 languages. It is built from parallel Bible translations by annotating English New Testament verses with 6 topics and projecting labels to other languages via verse-level alignment. The 6 classes are \textit{Recommendation}, \textit{Faith}, \textit{Description}, \textit{Sin}, \textit{Grace}, and \textit{Violence}.

\paragraph{Language Selection.} We identify seven overlapping languages between \data{SIB200} and \data{TAXI1500}: \textit{English}, \textit{German}, \textit{Chinese}, \textit{Arabic}, \textit{Swahili}, \textit{Vietnamese}, and \textit{Russian}. This selection exemplifies typological diversity across language families, encompassing both high- and low-resource languages across diverse scripts.

\subsection{Models}
In our experiments, we evaluate \our on four state-of-the-art open-source multilingual LLMs ranging from 4B to 12B parameters across two model families: \lm{Qwen3-\{4B,8B\}} \cite{yang2025qwen3technicalreport} and \lm{Gemma3-\{4B,12B\}} \cite{gemmateam2025gemma3technicalreport}, all of which support the languages investigated in this paper (\S\ref{subsec:dataset}).\footnote{Model information is reported in Appendix \ref{app:model}.}

\subsection{Baselines}
\label{subsec:baseline}
\citet{wang-etal-2026-parallel} propose two baselines for multilingual SCE generation: DG-CF and TB-CF. \ding{192} DG-CF is based on one-shot CoT prompting and generates SCEs directly in the target language by (1) identifying the most important words and (2) finding appropriate replacement words and substituting them (Appendix Figure~\ref{fig:prompt_cf}). As \citet{wang-etal-2026-parallel} report that SCEs in English achieve the highest validity, \ding{193} TB-CF first generates English SCEs using DG-CF, and then translates them into the target language using the same LLM. 
\ding{194} For a fair comparison, we further apply SFT with $\mathcal{D}_{\text{SFT}}=\cup_{\ell \in \mathcal{L}} \mathcal{D}_{\text{SFT}}^{(\ell)}$, where $\mathcal{D}_{\text{SFT}}^{(\ell)} = \{(x^{(\ell)}, y_w^{(\ell)})_i\}_{i=1}^N$, i.e, using only preferred answers (\S\ref{subsec:dataset_optimization}), \textit{without} dispreferred answers. \our uses richer (internal) signals from the base model (\S\ref{subsec:preference_ranking}) to explicitly optimize \textit{validity} while employing \textit{minimality} regularization, whereas the other baselines either do not use these signals or do not exploit them to the same extent.



\subsection{Automatic Evaluation Metrics}
Following prior work on counterfactual explanations \cite{nguyen-etal-2024-llms,wang2026iflipiterativefeedbackdrivencounterfactual,ross-etal-2021-explaining,dehghanighobadi-etal-2025-llms}, we conduct automatic evaluation to assess the generated multilingual SCEs along three dimensions: (i) \textit{validity}; (ii) \textit{fluency}; and (iii) \textit{minimality}.

\paragraph{Soft Label Flipping Rate (SLFR)} measures validity, defined as the fraction of SCEs that successfully change the model prediction.\footnote{We emphasize that SLFR checks whether $\tilde{x}$ has a different prediction $\mathcal{M}(\tilde{x})$ than $y$ and is thus different from the flip score (\S\ref{subsubsec:flip}), which checks whether $\tilde{x}$ reaches the target label $\hat{y}$, where $\hat{y}$ is not necessarily equal to $\mathcal{M}(\tilde{x})$.} Due to the self-explanation property, the same model $\mathcal{M}$ is used both for generating SCE $\tilde{x}$ and for evaluating whether the predicted label is shifted. Formally, over $N$ evaluation instances, SLFR is computed as:
\begin{align*}
\mathrm{SLFR}
= \frac{1}{N}\sum_{i=1}^{N}\mathbb{I}\big[\mathcal{M}(\tilde{x}_{i}) \neq \mathcal{M}(x_{i})\big]
\end{align*}

\paragraph{Fluency}
We assess fluency using perplexity score  (PPL) assigned by \lm{mGPT}\footnote{\url{https://huggingface.co/ai-forever/mGPT}} \cite{shliazhko-etal-2024-mgpt}, an open-source multilingual autoregressive language model. \lm{mGPT} is selected because it excels at modeling text distributions and covers the support of all investigated languages. For a token sequence $\tilde{x}=(w_1,\dots,w_T)$, PPL is defined as follows:
\begin{align*}
\mathrm{PPL}(\tilde{x})
= \exp\left\{-\frac{1}{T}\sum_{t=1}^{T}\log p_{\theta}(w_t \mid w_{<t})\right\}
\end{align*}
where $p_{\theta}$ denotes the next-token probability.

\begin{table*}[t!]
\centering
\renewcommand*{\arraystretch}{1}
\resizebox{\textwidth}{!}{%
\begin{tabular}{
  cc @{\hspace{1.2em}} 
  *{3}{c} @{\hspace{1.2em}} 
  *{3}{c} @{\hspace{1.2em}} 
  *{3}{c} @{\hspace{2.5em}} 
  *{3}{c} @{\hspace{1.2em}} 
  *{3}{c} @{\hspace{1.2em}} 
  *{3}{c}
}
\toprule[1.5pt]

\multicolumn{2}{c}{\textbf{Dataset}} & 
\multicolumn{9}{c}{\textbf{\data{SIB200}}} & 
\multicolumn{9}{c}{\textbf{\data{TAXI1500}}}\\

\cmidrule(r){1-2} \cmidrule(lr){3-11} \cmidrule(l){12-20}

\multirow{2}{*}{\rotatebox[origin=c]{90}{\scriptsize{\textbf{Model}}}} & \textbf{Lang-} 
& \multicolumn{3}{c}{\textbf{Validity/SLFR} (\(\uparrow\))} 
& \multicolumn{3}{c}{\textbf{Fluency/PPL} (\(\downarrow\))} 
& \multicolumn{3}{c}{\textbf{Minimality/SS} (\(\uparrow\))} 
& \multicolumn{3}{c}{\textbf{Validity/SLFR} (\(\uparrow\))} 
& \multicolumn{3}{c}{\textbf{Fluency/PPL} (\(\downarrow\))} 
& \multicolumn{3}{c}{\textbf{Minimality/SS} (\(\uparrow\))}\\

 & \textbf{uage} 
 & \textbf{DG-CF} & \textbf{SFT} & \textbf{\our}
 & \textbf{DG-CF} & \textbf{SFT} & \textbf{\our}
 & \textbf{DG-CF} & \textbf{SFT} & \textbf{\our}
 & \textbf{DG-CF} & \textbf{SFT} & \textbf{\our}
 & \textbf{DG-CF} & \textbf{SFT} & \textbf{\our}
 & \textbf{DG-CF} & \textbf{SFT} & \textbf{\our}\\
\midrule

\centering \multirow{8}{*}{\rotatebox[origin=c]{90}{\lm{Gemma3-4B}}}
 & \textsf{en} & 0.63 & \bestresult{0.70} & \secondresult{0.64} & 48.8 & \bestresult{31.4} & \secondresult{31.7} & \secondresult{0.61} & 0.55 & \bestresult{0.65} & 0.53 & \secondresult{0.56} & \bestresult{0.60} & 36.2 & \bestresult{23.4} & \secondresult{23.5} & 0.51 & \bestresult{0.59} & \secondresult{0.59}\\
 & \textsf{de} & 0.59 & \bestresult{0.65} & \secondresult{0.64} & 28.5 & \bestresult{24.0} & \secondresult{24.4} & \secondresult{0.75} & 0.71 & \bestresult{0.78} & 0.56 & \secondresult{0.59} & \bestresult{0.66} & 33.9 & \secondresult{25.2} & \bestresult{23.0} & 0.64 & \secondresult{0.69} & \bestresult{0.73}\\
 & \textsf{zh} & 0.66 & \secondresult{0.76} & \bestresult{0.80} & 35.1 & \bestresult{28.5} & \secondresult{30.1} & \bestresult{0.69} & 0.65 & \secondresult{0.66} & 0.52 & \bestresult{0.61} & \secondresult{0.60} & 39.5 & \bestresult{30.3} & \secondresult{32.4} & 0.61 & \secondresult{0.65} & \bestresult{0.67}\\
 & \textsf{ar} & 0.61 & \bestresult{0.74} & \secondresult{0.66} & 27.7 & \bestresult{24.6} & \secondresult{26.3} & \bestresult{0.66} & 0.60 & \secondresult{0.65} & 0.52 & \secondresult{0.54} & \bestresult{0.57} & 41.9 & \secondresult{32.8} & \bestresult{30.5} & \secondresult{0.63} & 0.60 & \bestresult{0.66}\\
 & \textsf{vi} & 0.59 & \bestresult{0.65} & \secondresult{0.65} & 18.9 & \bestresult{16.5} & \secondresult{18.7} & \secondresult{0.71} & 0.67 & \bestresult{0.71} & \secondresult{0.50} & 0.51 & \bestresult{0.51} & 18.5 & \secondresult{17.5} & \bestresult{16.3} & 0.61 & \secondresult{0.64} & \bestresult{0.66}\\
 & \textsf{sw} & 0.68 & \secondresult{0.72} & \bestresult{0.74} & 18.7 & \secondresult{18.2} & \bestresult{17.6} & \secondresult{0.68} & \bestresult{0.69} & 0.67 & \secondresult{0.57} & 0.57 & \bestresult{0.64} & 25.6 & \secondresult{20.6} & \bestresult{19.9} & 0.67 & \secondresult{0.75} & \bestresult{0.76}\\
 & \textsf{ru} & 0.61 & \bestresult{0.72} & \secondresult{0.64} & 20.3 & \secondresult{19.3} & \bestresult{19.1} & \secondresult{0.74} & 0.70 & \bestresult{0.74} & 0.53 & \secondresult{0.54} & \bestresult{0.60} & 21.5 & \bestresult{17.9} & \secondresult{19.0} & \secondresult{0.66} & 0.65 & \bestresult{0.68}\\
\cdashline{2-20}
 & \textsf{avg} & 0.62 & \bestresult{0.71} & \secondresult{0.68} & 28.3 & \bestresult{23.2} & \secondresult{24.0} & \secondresult{0.69} & 0.65 & \bestresult{0.69} & 0.53 & \secondresult{0.56} & \bestresult{0.60} & 31.0 & \secondresult{24.0} & \bestresult{23.5} & 0.62 & \secondresult{0.65} & \bestresult{0.68}\\

\midrule
\addlinespace[0.5em]

\centering \multirow{8}{*}{\rotatebox[origin=c]{90}{\lm{Qwen3-4B}}}
 & \textsf{en} & 0.54 & \bestresult{0.68} & \secondresult{0.66} & 53.4 & \bestresult{32.8} & \secondresult{34.8} & \bestresult{0.79} & 0.68 & \secondresult{0.75} & 0.62 & \secondresult{0.63} & \bestresult{0.66} & 32.6 & \bestresult{25.1} & \secondresult{26.1} & \secondresult{0.61} & 0.60 & \bestresult{0.68}\\
 & \textsf{de} & 0.62 & \secondresult{0.65} & \bestresult{0.66} & 44.4 & \bestresult{26.9} & \secondresult{27.9} & \secondresult{0.71} & 0.71 & \bestresult{0.76} & 0.49 & \bestresult{0.54} & \secondresult{0.53} & 40.2 & \bestresult{26.8} & \secondresult{27.0} & 0.66 & \secondresult{0.67} & \bestresult{0.70}\\
 & \textsf{zh} & \bestresult{0.81} & \secondresult{0.78} & 0.72 & 47.7 & \bestresult{31.7} & \secondresult{33.5} & 0.60 & \secondresult{0.61} & \bestresult{0.73} & 0.59 & \bestresult{0.66} & \secondresult{0.60} & 45.8 & \bestresult{36.9} & \secondresult{39.1} & \secondresult{0.67} & 0.65 & \bestresult{0.71}\\
 & \textsf{ar} & 0.64 & \bestresult{0.72} & \secondresult{0.66} & 43.3 & \bestresult{29.8} & \secondresult{32.2} & \secondresult{0.70} & 0.59 & \bestresult{0.70} & 0.68 & \secondresult{0.69} & \bestresult{0.72} & 86.2 & \bestresult{43.6} & \secondresult{48.6} & \secondresult{0.58} & 0.53 & \bestresult{0.60}\\
 & \textsf{vi} & 0.71 & \bestresult{0.73} & \secondresult{0.71} & 25.0 & \bestresult{18.9} & \secondresult{21.7} & \secondresult{0.69} & 0.65 & \bestresult{0.71} & \secondresult{0.65} & 0.60 & \bestresult{0.66} & 25.7 & \secondresult{21.5} & \bestresult{20.7} & 0.66 & \secondresult{0.67} & \bestresult{0.67}\\
 & \textsf{sw} & \bestresult{0.49} & \secondresult{0.47} & 0.41 & 31.2 & \secondresult{21.6} & \bestresult{21.2} & 0.73 & \secondresult{0.78} & \bestresult{0.80} & 0.46 & \secondresult{0.48} & \bestresult{0.51} & 25.6 & \secondresult{18.8} & \bestresult{17.7} & \bestresult{0.84} & \secondresult{0.84} & 0.83\\
 & \textsf{ru} & 0.45 & \bestresult{0.69} & \secondresult{0.62} & 26.8 & \bestresult{21.2} & \secondresult{22.0} & \bestresult{0.84} & 0.72 & \secondresult{0.77} & 0.53 & \secondresult{0.57} & \bestresult{0.64} & 27.7 & \bestresult{23.1} & \secondresult{23.2} & \secondresult{0.68} & 0.67 & \bestresult{0.71}\\
\cdashline{2-20}
 & \textsf{avg} & 0.61 & \bestresult{0.67} & \secondresult{0.63} & 38.8 & \bestresult{26.1} & \secondresult{27.6} & \secondresult{0.72} & 0.68 & \bestresult{0.75} & 0.57 & \secondresult{0.60} & \bestresult{0.62} & 40.5 & \bestresult{28.0} & \secondresult{28.9} & \secondresult{0.67} & 0.66 & \bestresult{0.70}\\

\midrule
\addlinespace[0.5em]

\centering \multirow{8}{*}{\rotatebox[origin=c]{90}{\lm{Qwen3-8B}}}
 & \textsf{en} & 0.59 & \secondresult{0.70} & \bestresult{0.74} & 53.2 & \bestresult{33.0} & \secondresult{37.9} & \bestresult{0.73} & 0.62 & \secondresult{0.66} & 0.56 & \secondresult{0.70} & \bestresult{0.72} & 31.0 & \bestresult{20.7} & \secondresult{25.1} & \secondresult{0.68} & 0.61 & \bestresult{0.69}\\
 & \textsf{de} & 0.65 & \bestresult{0.75} & \secondresult{0.73} & 32.1 & \bestresult{24.6} & \secondresult{27.8} & \secondresult{0.69} & 0.65 & \bestresult{0.71} & 0.50 & \secondresult{0.62} & \bestresult{0.71} & 25.9 & \bestresult{21.4} & \secondresult{22.9} & 0.62 & \secondresult{0.65} & \bestresult{0.66}\\
 & \textsf{zh} & 0.62 & \bestresult{0.72} & \secondresult{0.70} & 38.6 & \bestresult{30.0} & \secondresult{34.9} & \secondresult{0.68} & 0.67 & \bestresult{0.70} & \secondresult{0.63} & \secondresult{0.63} & \bestresult{0.66} & 38.9 & \bestresult{30.2} & \secondresult{33.6} & \secondresult{0.66} & 0.63 & \bestresult{0.71}\\
 & \textsf{ar} & 0.58 & \bestresult{0.77} & \secondresult{0.76} & 34.9 & \bestresult{26.8} & \secondresult{30.5} & \bestresult{0.74} & 0.60 & \secondresult{0.65} & 0.66 & \bestresult{0.84} & \secondresult{0.76} & 50.3 & \bestresult{37.0} & \secondresult{37.8} & \bestresult{0.64} & 0.59 & \secondresult{0.63}\\
 & \textsf{vi} & 0.68 & \bestresult{0.73} & \secondresult{0.72} & \secondresult{22.7} & \bestresult{18.3} & 23.1 & \secondresult{0.69} & 0.65 & \bestresult{0.71} & 0.58 & \secondresult{0.70} & \bestresult{0.78} & 18.8 & \bestresult{16.0} & \secondresult{18.0} & \secondresult{0.61} & 0.60 & \bestresult{0.64}\\
 & \textsf{sw} & 0.53 & \secondresult{0.56} & \bestresult{0.57} & 22.9 & \bestresult{18.5} & \secondresult{19.7} & \bestresult{0.84} & 0.78 & \secondresult{0.79} & 0.11 & \bestresult{0.32} & \secondresult{0.27} & 19.0 & \secondresult{17.4} & \bestresult{16.8} & \bestresult{0.83} & 0.79 & \secondresult{0.83}\\
 & \textsf{ru} & 0.64 & \bestresult{0.72} & \secondresult{0.66} & 22.1 & \bestresult{19.6} & \secondresult{21.5} & \secondresult{0.70} & 0.64 & \bestresult{0.71} & 0.45 & \secondresult{0.56} & \bestresult{0.61} & \secondresult{18.7} & \bestresult{15.6} & 20.2 & \secondresult{0.68} & 0.66 & \bestresult{0.73}\\
\cdashline{2-20}
 & \textsf{avg} & 0.61 & \bestresult{0.71} & \secondresult{0.70} & 32.4 & \bestresult{24.4} & \secondresult{27.9} & \bestresult{0.72} & 0.66 & \secondresult{0.70} & 0.50 & \secondresult{0.62} & \bestresult{0.64} & 28.9 & \bestresult{22.6} & \secondresult{24.9} & \secondresult{0.67} & 0.65 & \bestresult{0.70}\\

\midrule
\addlinespace[0.5em]

\centering \multirow{8}{*}{\rotatebox[origin=c]{90}{\lm{Gemma3-12B}}}
 & \textsf{en} & 0.83 & \bestresult{0.90} & \secondresult{0.84} & 53.9 & \bestresult{34.3} & \secondresult{38.3} & \bestresult{0.53} & 0.49 & \secondresult{0.51} & \secondresult{0.83} & 0.81 & \bestresult{0.86} & 41.5 & \bestresult{26.7} & \secondresult{27.9} & 0.44 & \secondresult{0.45} & \bestresult{0.49}\\
 & \textsf{de} & 0.77 & \secondresult{0.84} & \bestresult{0.85} & 28.1 & \bestresult{24.1} & \secondresult{27.0} & 0.57 & \bestresult{0.58} & \secondresult{0.58} & \secondresult{0.79} & 0.77 & \bestresult{0.81} & 26.5 & \secondresult{21.8} & \bestresult{21.7} & 0.54 & \secondresult{0.59} & \bestresult{0.60}\\
 & \textsf{zh} & \secondresult{0.84} & 0.81 & \bestresult{0.92} & \secondresult{31.6} & \bestresult{29.7} & 32.6 & \secondresult{0.55} & \bestresult{0.57} & 0.51 & \secondresult{0.68} & 0.60 & \bestresult{0.76} & \secondresult{27.9} & \bestresult{26.0} & 28.9 & 0.54 & \secondresult{0.56} & \bestresult{0.59}\\
 & \textsf{ar} & 0.87 & \secondresult{0.87} & \bestresult{0.92} & \secondresult{25.5} & \bestresult{25.1} & 26.6 & \secondresult{0.52} & \bestresult{0.55} & 0.49 & \secondresult{0.78} & 0.79 & \bestresult{0.79} & 37.3 & \bestresult{28.1} & \secondresult{36.3} & 0.54 & \secondresult{0.55} & \bestresult{0.59}\\
 & \textsf{vi} & \secondresult{0.79} & 0.79 & \bestresult{0.80} & \secondresult{18.5} & \bestresult{17.7} & 21.9 & 0.56 & \bestresult{0.59} & \secondresult{0.57} & \secondresult{0.74} & 0.69 & \bestresult{0.80} & \secondresult{16.2} & \bestresult{15.1} & 17.1 & 0.52 & \secondresult{0.55} & \bestresult{0.56}\\
 & \textsf{sw} & 0.81 & \secondresult{0.82} & \bestresult{0.83} & \secondresult{14.9} & \bestresult{14.7} & 15.9 & 0.65 & \secondresult{0.68} & \bestresult{0.71} & \secondresult{0.69} & 0.62 & \bestresult{0.73} & 17.5 & \secondresult{16.7} & \bestresult{15.9} & 0.68 & \secondresult{0.73} & \bestresult{0.75}\\
 & \textsf{ru} & 0.78 & \secondresult{0.79} & \bestresult{0.80} & \secondresult{21.0} & \bestresult{19.6} & 22.1 & \secondresult{0.59} & 0.57 & \bestresult{0.61} & \secondresult{0.75} & 0.69 & \bestresult{0.81} & 20.5 & \bestresult{16.3} & \secondresult{18.5} & 0.60 & \secondresult{0.61} & \bestresult{0.61}\\
\cdashline{2-20}
 & \textsf{avg} & 0.81 & \secondresult{0.83} & \bestresult{0.85} & 27.6 & \bestresult{23.6} & \secondresult{26.3} & 0.57 & \bestresult{0.58} & \secondresult{0.57} & \secondresult{0.75} & 0.71 & \bestresult{0.79} & 26.8 & \bestresult{21.5} & \secondresult{23.8} & 0.55 & \secondresult{0.58} & \bestresult{0.60}\\
\bottomrule[1.5pt]
\end{tabular}%
}
\caption{Automatic evaluation results on \data{SIB200} and \data{TAXI1500}. We compare DG-CF, SFT, and \our across three metrics: soft label flipping rate (SLFR (\(\uparrow\))), perplexity (PPL (\(\downarrow\))), and semantic similarity (SS (\(\uparrow\))). The \bestlegend\ and \secondlegend\ results for each language-metric pair are highlighted accordingly. Results for TB-CF are provided in Appendix Table~\ref{tab:dgcf_tbcf_with_deltas} and discussed in \S\ref{subsec:automatic_evaluation}.}
\label{tab:automatic_evaluation_sib200_taxi1500_main}
\end{table*}

\paragraph{Semantic Similarity (SS)}
Minimality can be assessed by quantifying the semantic similarity between the original input $x$ and its counterfactual $\tilde{x}$ with multilingual sentence embeddings, thereby enabling mathematically grounded cross-lingual comparability across the languages studied in this work.
We compute SS as the average cosine similarity between $x$ and $\tilde{x}$ over $N$ instances:
\begin{align*}
\mathrm{SS}
= \frac{1}{N}\sum_{i=1}^{N}
\frac{\mathbf{e}(x_{i})^\top \mathbf{e}(\tilde{x}_{i})}
{\lVert \mathbf{e}(x_{i})\rVert_2 \, \lVert \mathbf{e}(\tilde{x}_{i})\rVert_2}
\end{align*}
where $\mathbf{e}(\cdot)$ denotes the sentence embedding.\footnote{\url{https://huggingface.co/sentence-transformers/paraphrase-multilingual-MiniLM-L12-v2}}

\subsection{Cross-lingual Edit Similarity}

Cross-lingual Edit Similarity (CES) measures the degree to which semantic modifications in counterfactuals $\tilde{x}^{(\ell)}$ are consistently applied across language pairs $(\ell_1, \ell_2)$ relative to the original input \cite{wang-etal-2026-parallel}:
\begin{align*}
    \text{CES}(\ell_1, \ell_2) = \frac{1}{N} \sum_{i=1}^{N} \frac{\mathbf{e}(\tilde{x}_{i}^{(\ell_1)})\top \mathbf{e}(\tilde{x}_{i}^{(\ell_2)})}{\lVert \mathbf{e}(\tilde{x}_i^{(\ell_1)}) \rVert_2 \, \lVert \mathbf{e}(\tilde{x}_i^{(\ell_2)}) \rVert_2}
\end{align*}
Higher values indicate that edits are transferred more uniformly between the two languages. 
We adopt CES to examine how perturbation consistency in multilingual SCE generation evolves through \our.



\section{Results}

\subsection{Multilingual Counterfactual Quality}
\label{subsec:automatic_evaluation}

\begin{figure}[t]
    \centering
    \includegraphics[width=\columnwidth]{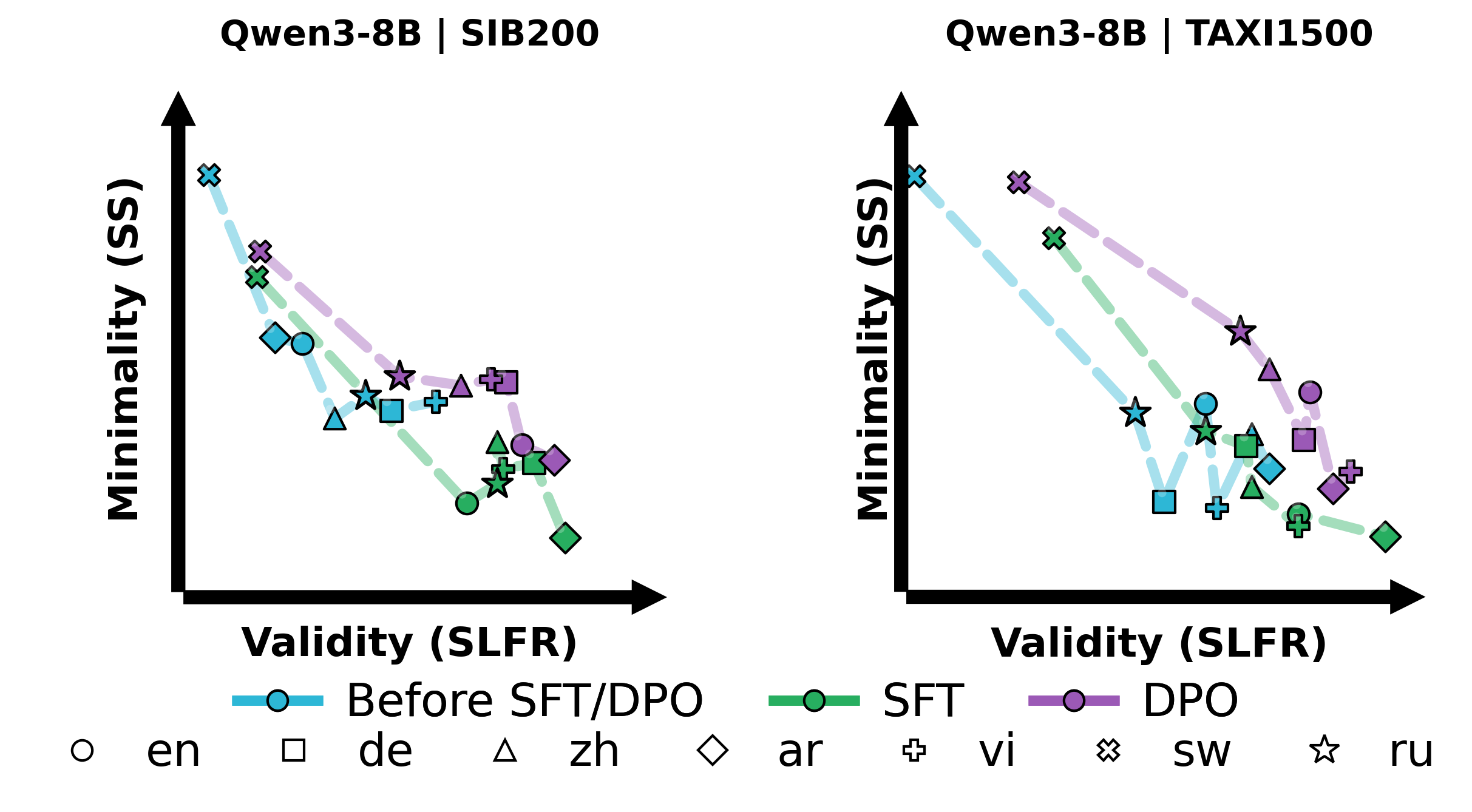}
    \caption{The validity-minimality trade-off across languages with \lm{Qwen3-8B} on \data{SIB200} and \data{Taxi1500}.}
    \label{fig:tradeoff_qwen8}
\end{figure}

\paragraph{\our generally outperforms DG-CF and SFT.} Table~\ref{tab:automatic_evaluation_sib200_taxi1500_main} shows that \our achieves average relative improvements of 12.55\% in \textit{validity}, 18.63\% in \textit{fluency}, and 3.13\% in \textit{minimality} over DG-CF. This highlights \our's ability to mitigate the \textit{validity-minimality} trade-off effectively (Figure~\ref{fig:tradeoff_qwen8}). Notably, improvements are more substantial on \data{TAXI1500}, likely because its parallel Bible verses construction provides a more predictable syntactic and semantic structure that facilitates counterfactual optimization.
While larger models generally outperform smaller counterparts within families, performance differences between high- and relatively low-resource languages (\textsf{ar}, \textsf{vi}, \textsf{sw}) remain surprisingly marginal. Moreover, no single language invariably yields optimal SCEs across all metrics. Regarding cross-lingual stability, \lm{Gemma3-12B} exhibits the lowest \textit{validity} variation across seven languages, whereas \lm{Qwen3-8B} shows the poorest cross-lingual stability, driven primarily by its suboptimal performance on Swahili. 
Additionally, \our outperforms SFT on both \textit{validity} (+1.3\%) and \textit{minimality} (+3.5\%) on average. While SFT occasionally surpasses \our on \textit{validity} on \data{SIB200}, its consistent underperformance on \textit{minimality} (frequently even worse than the DG-CF baseline) makes it less suitable for faithful SCE generation. This confirms that the trade-off cannot be easily resolved through standard supervised learning alone and preference optimization proves essential for achieving Pareto-optimal SCE quality.\footnote{Appendix~\ref{app:impact} further demonstrates that, compared to SFT, \our better preserves or even enhances the model's general and multilingual capabilities while simultaneously enhancing cross-lingual generalizability in terms of SCE validity. Appendix~\ref{app:cls_degradation} additionally shows that performance on the original classification task is preserved.} \looseness=-1

\paragraph{\our produces more reliable and stable SCEs than TB-CF.} \label{par:translation} Appendix Table~\ref{tab:dgcf_tbcf_with_deltas} shows that, while TB-CF achieves a higher SLFR than DG-CF, SFT, and \our, it requires substantially more modifications, as indicated by markedly reduced \textit{minimality}. Specifically, its Edit Score ($\mathcal{R}_{\text{edit}}$; \S\ref{subsubsec:edit}) is frequently below 0, suggesting that the edit length surpasses that of the original input. Manual inspection reveals that these excessive modifications primarily 
stem from translation artifacts rather than semantically 
meaningful changes (Tables~\ref{fig:taxi1500-macro-example} and \ref{fig:sib200-macro-example}). \textit{This violates the minimality constraint of counterfactual explanations} (\S\ref{sec:marco}). 
Consequently, despite its high SLFR (\textit{validity}), TB-CF strays noticeably from the original input without pinpointing the nuanced feature changes that account for the explained model's decisions (\textit{minimality}), severely undermining their reliability as SCEs.

\twocolumn[{%
\centering
\includegraphics[width=\textwidth]{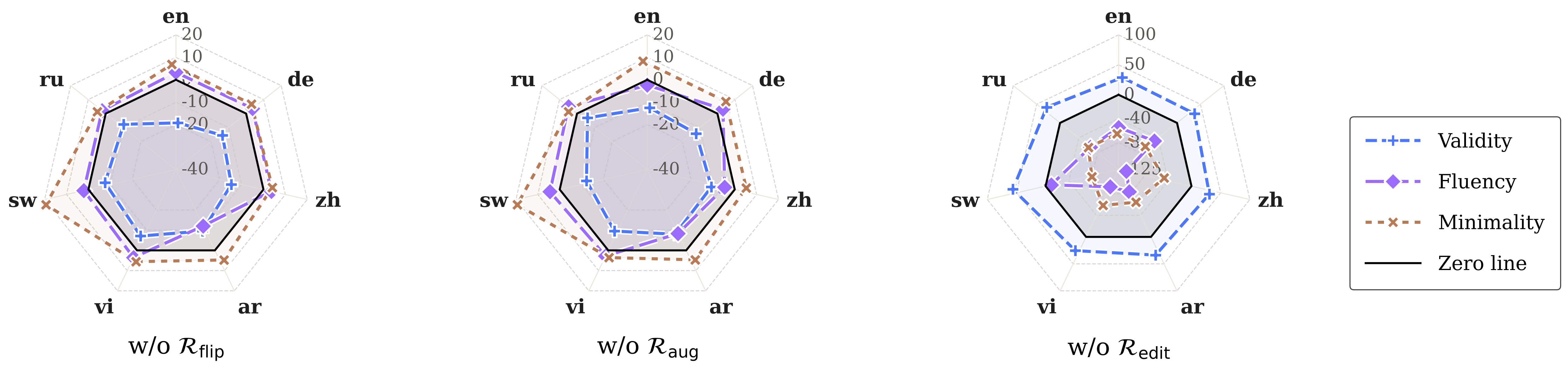}
\captionof{figure}{Relative performance change across languages for \lm{Qwen3-8B} under three ablation settings. We evaluate the removal of $\mathcal{R}_{\text{flip}}$, $\mathcal{R}_{\text{aug}}$, and $\mathcal{R}_{\text{edit}}$ across \textit{validity}, \textit{fluency}, and \textit{minimality}. Each radar plot illustrates results for a specific ablation, with scores averaged over \data{SIB200} and \data{TAXI1500}. \textit{Positive} values denote a percentage \textit{improvement} in performance, while \textit{negative} values indicate a \textit{degradation} relative to the full \our configuration.}
\label{fig:additional_analysis_radar}
\par\vspace{1.5em}
}]

\paragraph{Validity-Minimality Trade-off.}

The trade-off between \textit{validity} and \textit{minimality} observed by \citet{mayne-etal-2025-llms} can be ascribable to the imperfect SCE generation approaches, which are mostly prompting-based without robust regularization to ensure the \textit{minimality} \cite{cardiel2026drivexcounterfactualexplanationsdriving}, as also evidenced by our ablation study (\S\ref{subsec:ablation_study}). Given that we incorporate three meticulously designed scoring components aimed at optimizing \textit{validity} and \textit{minimality} (\S\ref{subsec:preference_ranking}), \our is able to enhance \textit{validity} by 12.55\% without sacrificing \textit{minimality} (even enhancing \textit{minimality} by 3.13\% on average), which largely mitigates the previously observed trade-off \cite{mayne-etal-2025-llms} (Figure \ref{fig:tradeoff_qwen8}, Appendix~\ref{app:trade-off}).

\paragraph{English is not necessarily the optimal language for counterfactual generation.} Table~\ref{tab:automatic_evaluation_sib200_taxi1500_main} reveals that English counterfactuals do not consistently achieve the highest validity across all experimental settings. This observation contradicts the findings of \citet{wang-etal-2026-parallel}, who report that counterfactuals in high-resource European languages generally exhibit superior quality. We argue that the optimal language is highly dependent on the explained model and its internal decision boundaries for different languages, which vary across models.

\subsection{Ablation Study}
\label{subsec:ablation_study}
\our employs three scoring components for preference pair ranking (\S\ref{subsec:preference_ranking}): $\mathcal{R}_{\text{flip}}, \mathcal{R}_{\text{aug}}, \mathcal{R}_{\text{edit}}$, which collectively aim to optimize for \textit{validity} and \textit{minimality} of counterfactuals. Accordingly, to demonstrate their individual contributions, we conduct an ablation study for each component.\footnote{The discussion of the exclusion of multilingual alignment is provided in Appendix~\ref{app:multilingual_align}.} Figure~\ref{fig:additional_analysis_radar} and Appendix Table~\ref{tab:automatic_evaluation_ablation} present the ablation study results for \lm{Qwen3-8B}, obtained by removing one scoring component (\S\ref{subsec:preference_ranking}) at a time.\footnote{Further results with other models appear in Appendix~\ref{subsec:ablation_full_app}; example SCEs from each ablation condition are shown in Appendix~\ref{app:ablation_examples}.} 

\paragraph{Effect of Removing $\mathcal{R}_{\text{flip}}$.} Removing $\mathcal{R}_{\text{flip}}$ leads to the most pronounced drop in \textit{validity}, with an average absolute decline of 11.65\% compared to the full model  (Appendix Table~\ref{tab:flip}). 
Meanwhile, the removal of $\mathcal{R}_{\text{flip}}$ yields moderate drops in fluency and improvement in \textit{minimality}. This occurs because removing $\mathcal{R}_{\text{flip}}$ allows $\mathcal{R}_{\text{edit}}$ to dominate the ranking, preferentially selecting candidates with fewer modifications regardless of whether they achieve a successful label flip. The result confirms that an explicit label-flip constraint is essential; without it, the model gravitates toward the original input and fails at the primary objective of SCE generation.



\paragraph{Effect of Removing $\mathcal{R}_{\text{aug}}$.} \textit{Validity} decreases notably without $\mathcal{R}_{\text{aug}}$, although the degradation is less severe than w/o $\mathcal{R}_{\text{flip}}$ (Appendix Table~\ref{tab:aug}), confirming that $\mathcal{R}_{\text{aug}}$ serves as a complementary objective to $\mathcal{R}_{\text{flip}}$. Mechanistically, $\mathcal{R}_{\text{aug}}$ is designed to facilitate margin maximization (\S\ref{subsubsec:aug}), effectively ``nudging'' borderline SCEs across the decision boundary. In its absence, these near-miss candidates often fail to achieve a successful label flip. Similar to w/o $\mathcal{R}_{\text{flip}}$, \textit{minimality} is generally enhanced without $\mathcal{R}_{\text{aug}}$.

\paragraph{Effect of Removing $\mathcal{R}_{\text{edit}}$.} The removal of $\mathcal{R}_{\text{edit}}$ results in a markedly different failure mode. \textit{Validity} consistently increases across all languages relative to the full model. However, this apparent improvement
in \textit{validity} comes at a severe cost to \textit{minimality} (Appendix Table~\ref{tab:edit}), revealing that the model achieves label flips by making excessively large modifications and deviating significantly from the original semantics. 
These findings demonstrate that, without $\mathcal{R}_{\text{edit}}$, the model degenerates into producing heavily modified outputs that trivially flip labels but fail to satisfy the \textit{minimality} criterion central to counterfactual explanations.

\paragraph{Language-Specific Observations.} Appendix Table~\ref{tab:avg} illustrates that \textbf{Swahili} emerges as the \textit{most} sensitive language overall, particularly in its extreme minimality and validity sensitivity. In comparison, \textbf{Russian} is the \textit{least} sensitive language, with the lowest validity sensitivity. \textbf{English} and \textbf{German} display the most balanced sensitivity profiles, with all three dimensions falling within a relatively narrow range, confirming that their label-flip success depends most critically on the collaboration among all three scoring components (Appendix~\ref{subsec:additional_analysis}).

\begin{figure}[t]
  \centering
  \includegraphics[width=\columnwidth]{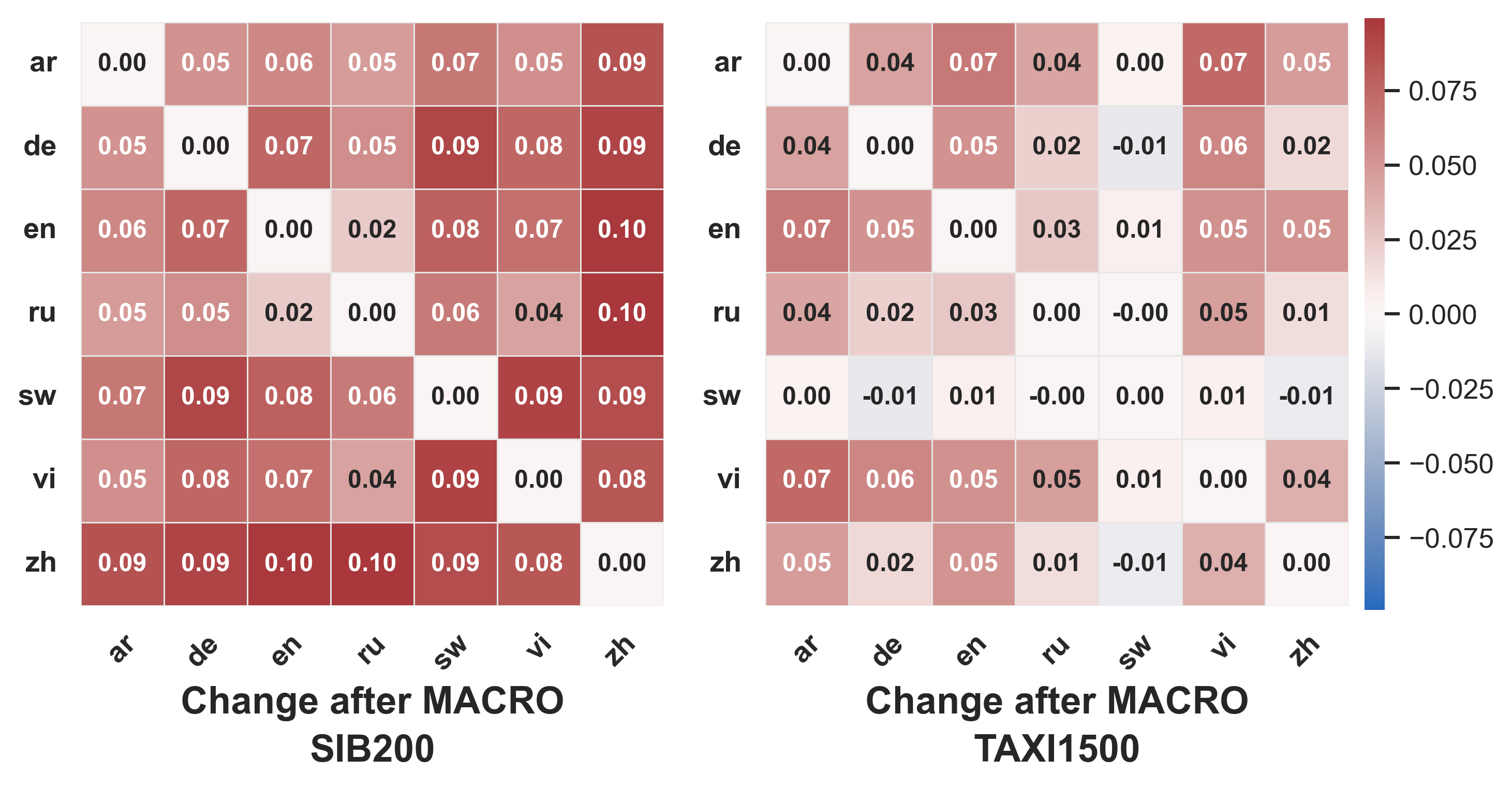}
  \caption{Cross-lingual edit similarity score changes ($\Delta$) for \data{SIB200} and \data{TAXI1500} counterfactuals across language pairs before and after applying \our.}
  \label{fig:ces_compare}
\end{figure}

\subsection{Cross-lingual Edit Similarity}
\label{subsec:ces}

\paragraph{Before applying \our.} Figures~\ref{fig:ces_compare} and \ref{fig:ces_score} reveal that, in alignment with the findings of \citet{wang-etal-2026-parallel}, closely related languages (German, English, and Russian) achieve moderately higher CES, suggesting that semantic edits in counterfactuals are applied with considerable consistency across these language pairs, whereas Swahili scores notably low across the board. 
Surprisingly, although Vietnamese, Arabic, and Chinese are not typologically close to English, German, or Russian, they achieve relatively high CES scores with them. This indicates that topological similarity between language pairs is not the only influential factor; the resource availability of languages also plays a role.

\paragraph{After applying \our.} CES improvements are observed across all language pairs on \data{SIB200}, with the most substantial gains on high-resource languages. On \data{Taxi1500}, the CES of the majority of language pairs is enhanced, but a few pairs involving Swahili exhibit marginal decreases. This may be attributed to the fact that \data{Taxi1500} likely demands finer-grained semantic edits and that \our primarily benefits well-represented languages by aligning their counterfactual editing behavior. This optimization may subtly shift the model's internal representations in a direction that marginally disadvantages Swahili pairs. \looseness=-1

\begin{figure}[t]
    \centering
    \includegraphics[width=\columnwidth]{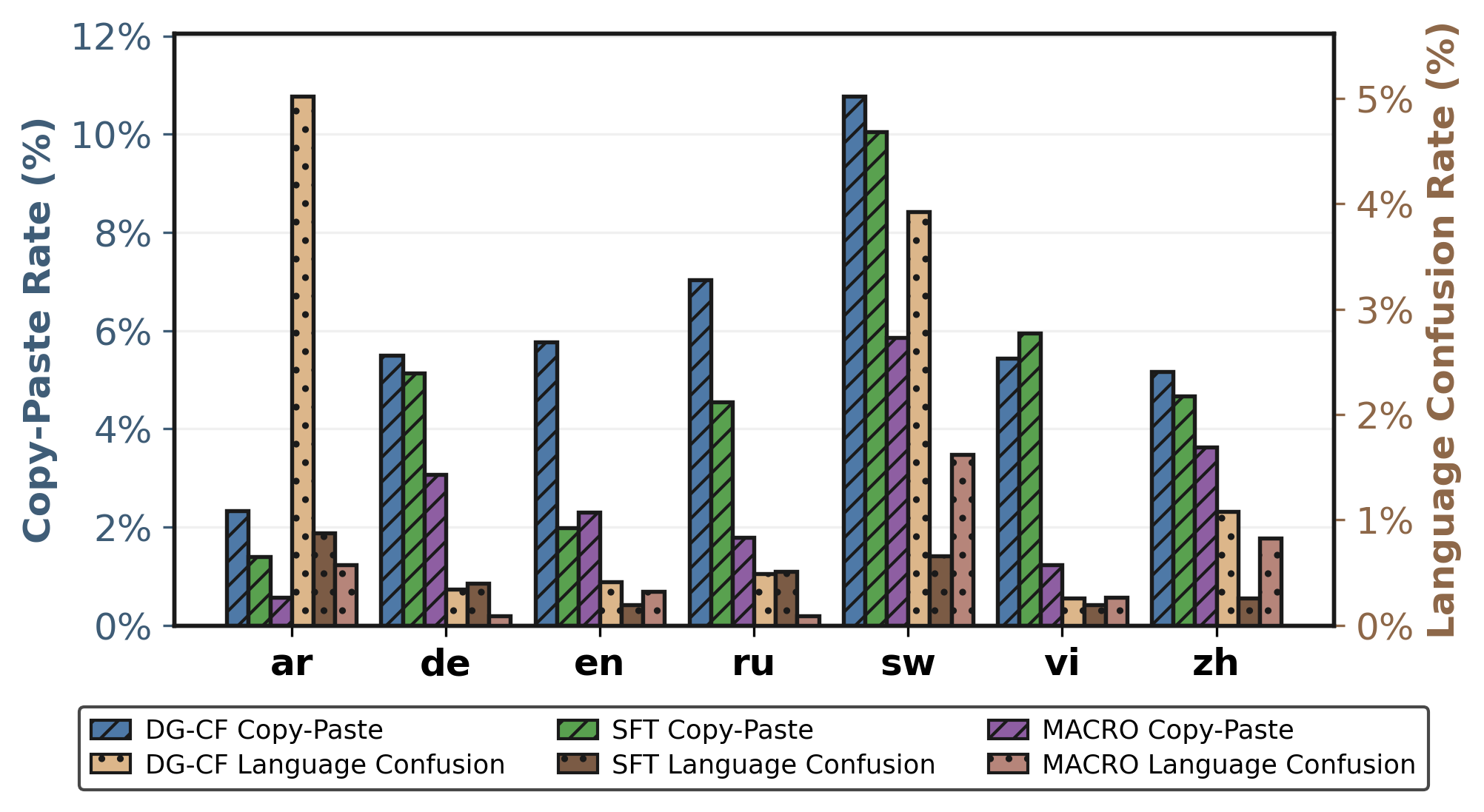}
    \caption{Comparison of \textit{copy-paste} and \textit{language confusion} rates between DG-CF and \our.}
    \label{fig:copy_paste_and_confusion}
\end{figure}

\subsection{Error Analysis}
\label{subsec:error_analysis}

We further report two main failure cases
: \textit{copy-paste} and \textit{language confusion}.

\begin{table*}[ht]
\centering
\small
\setlength{\tabcolsep}{6pt}
\begin{tabular}{l cc|ccc @{\hspace{2.2em}} cc|ccc}
\toprule[1.5pt]
\multirow{2}{*}{\textbf{Lang.}} & \multicolumn{5}{c}{\data{SIB200}} & \multicolumn{5}{c}{\data{TAXI1500}} \\
\cmidrule(lr){2-6} \cmidrule(lr){7-11}
 & \textbf{Zero-shot} & \textbf{SFT$_{\text{orig}}$} & \textbf{\our} & \textbf{DG-CF} & \textbf{SFT}
 & \textbf{Zero-shot} & \textbf{SFT$_{\text{orig}}$} & \textbf{\our} & \textbf{DG-CF} & \textbf{SFT} \\
\midrule
 \textsf{en} & 87.7 & 92.6 & \bestresult{94.6} & 92.2 & 92.6 & 50.5 & \bestresult{79.3} & 78.4 & 75.7 & 78.4 \\
 \textsf{de} & 89.7 & 93.6 & \bestresult{94.6} & 93.6 & 93.6 & 51.4 & \bestresult{75.7} & 73.9 & 73.0 & 71.2 \\
 \textsf{zh} & 85.8 & 92.2 & \bestresult{93.1} & 91.7 & 92.6 & 42.3 & \bestresult{76.6} & 75.7 & 74.8 & 74.8 \\
 \textsf{ar} & 88.2 & \bestresult{93.1} & \bestresult{93.1} & 92.2 & \bestresult{93.1} & 59.5 & 76.6 & 77.5 & 76.6 & \bestresult{79.3} \\
 \textsf{vi} & 89.2 & 92.2 & 91.2 & 90.7 & \bestresult{93.1} & 46.8 & 76.6 & 79.3 & 76.6 & \bestresult{80.2} \\
 \textsf{sw} & 85.3 & 89.7 & 92.2 & 91.2 & \bestresult{92.6} & 48.6 & \bestresult{78.4} & 77.5 & 74.8 & \bestresult{78.4} \\
 \textsf{ru} & 88.7 & \bestresult{94.1} & 93.1 & 91.2 & 93.6 & 44.1 & 74.8 & \bestresult{78.4} & 74.8 & 73.9 \\
\midrule
 \textbf{Avg} & 87.8 & \textbf{92.5} & \bestresult{\textbf{93.1}} & \textbf{91.8} & \textbf{93.0} & 49.0 & \textbf{76.9} & \bestresult{\textbf{77.2}} & \textbf{75.2} & \textbf{76.6} \\
 $\Delta$ & -- & -- & $+0.6$ & $-0.7$ & $+0.5$ & -- & -- & $+0.4$ & $-1.7$ & $-0.3$ \\
\bottomrule[1.5pt]
\end{tabular}
\caption{Counterfactual data augmentation with \lm{Gemma3-12B}. Accuracy (\%) on the original test inputs. $\Delta$ is relative to SFT$_{\text{orig}}$. \bestlegend marks the best fine-tuned condition per language.}
\label{tab:cda_gemma12}
\end{table*}

\paragraph{Copy-Paste.} When asked to generate a counterfactual, the LLM may tend to produce the identical text as the original input, either due to inertia or inability to provide an actionable label flip. Figure~\ref{fig:copy_paste_and_confusion} shows that \our reduces the copy-paste rate by an average of 58.21\% across all models and languages compared to DG-CF, since we intentionally designate such SCEs as dispreferred answers (\S\ref{subsec:dataset_optimization}). This improvement is substantially larger than that of SFT, which achieves only a 19.64\% relative reduction over DG-CF. At the language level, the contrast is especially clear in Swahili, where \our yields a pronounced reduction in copy-paste rate, while SFT brings only a comparatively modest change. 

\paragraph{Language Confusion.} We further analyze the language in which SCEs $\tilde{x}$ are generated and assess whether it corresponds to the desired target language.\footnote{\url{https://github.com/cisnlp/GlotLID}}
Figure~\ref{fig:copy_paste_and_confusion} illustrates that \our yields clear reductions in language confusion, with an average relative decline of 49.65\% across languages, which is particularly pronounced for Arabic and Swahili. Moreover, \our noticeably reduces the tendency to generate SCEs in non-targeted languages (most often English, given the models' English-centric properties), thereby improving consistency in targeted language generation.





\subsection{Counterfactual Data Augmentation} 
Beyond mere explanation, we further examine whether and to what extent the counterfactual data augmentation (CDA) performed by \our using \lm{Gemma3-12B} can improve the model's downstream task performance (the detailed setup is described in Appendix~\ref{app:cda}) \cite{kaushik2020learning, wang-etal-2025-truth}.  Table~\ref{tab:cda_gemma12} reports classification performance on the original test set after fine-tuning either on the original training data $\mathcal{D}_{\text{train}}$ alone or on $\mathcal{D}_{\text{CDA}}$, which augments $\mathcal{D}_{\text{train}}$ with counterfactuals from \our, DG-CF, or SFT:
\begin{align*}
    \mathcal{D}_{\text{train}}^{(\ell)} &= \{(x_i^{(\ell)}, y_i^{(\ell)})\}_{i=1}^{N} \\
    \mathcal{D}_{\text{CDA}}^{(\ell)} &= \mathcal{D}_{\text{train}}^{(\ell)} \cup \\
    &\{(\tilde{x}_i^{(\ell)}, \hat{y}_i^{(\ell)}) \mid \mathcal{M}(\tilde{x}_i^{(\ell)}) = \hat{y}_i^{(\ell)}\}_{i=1}^{N}
\end{align*}
Fine-tuning alone already recovers most of the gain over the zero-shot model, by $5.4\%$ on \data{SIB200} and $56.9\%$ on \data{TAXI1500} in relative terms, which leaves limited room for augmentation. Nevertheless, within that room, \our is the only source that consistently improves model performance on both datasets, whereas DG-CF lowers accuracy on both by $0.8\%$ and $2.2\%$, respectively. Augmentation, therefore, helps only when the generated counterfactuals are of sufficient quality, indicating that \our's SCEs are more reliable as training data than those of the baselines.

\section{Conclusion}
In this work, we proposed \our, a preference alignment framework that leverages DPO to optimize multilingual SCE generation. Our approach addresses two persistent challenges in extending SCEs beyond English: the failure of existing methods to produce valid SCEs in non-dominant languages and the inherent trade-off between \textit{validity} and \textit{minimality}. Experiments across four LLMs and seven languages demonstrate that \our consistently improves \textit{validity} over DG-CF by an average of 12.55\% without compromising \textit{minimality}, and avoids the excessive modifications introduced by TB-CF. Compared to SFT, \our achieves superior performance on both metrics, confirming that explicit preference optimization is essential to mitigate this trade-off. Ablation studies further confirm the complementary roles of the three preference scoring components in shaping effective preference signals. In addition, we showed that \our promotes greater cross-lingual alignment in input perturbation strategies and substantially mitigates common failure modes such as \textit{copy-paste} and \textit{language confusion} errors. Our work highlights that preference optimization is a promising paradigm for improving the quality of model self-explanations in multilingual settings. 


\section*{Limitations}
We acknowledge several limitations of this work:

\paragraph{Limited Exploration of Language.} This paper focuses only on seven selected languages (\S\ref{subsec:dataset}). While these languages represent diverse typological properties and resource levels, the effectiveness of \our on unexplored languages remains to be established. Extending \our to additional languages constitutes an important direction for future work.

\paragraph{Limited Exploration RL Approach.} In this paper, we investigate only the potential of DPO to enhance (multilingual) counterfactual explanation generation. Extending the investigation to different post-training strategies (e.g., online vs. offline learning) and their impact on counterfactual quality enhancement remains a promising direction for future work.

\paragraph{No Human Evaluation.}  Human evaluation is naturally appealing but challenging for us due to budget constraints and the difficulty of recruiting native speakers of low-resource languages. Nevertheless, we do consider it a promising future direction to assess the coherence or trustworthiness of multilingual counterfactual explanations \cite{wang2026largelanguagemodelsexplain}.

\section*{Acknowledgments}
We thank Van Bach Nguyen for his valuable feedback and constructive criticism, which greatly helped improve the paper. Additionally, we are indebted to the anonymous reviewers of EMNLP 2026 for their helpful and rigorous feedback.

\bibliography{custom}

\appendix

\section{Direct Preference Optimization}
\label{app:dpo}
\paragraph{Reward Modeling.} Given a dataset $\mathcal{D}=\{(x, y_w, y_l)_i\}_{i=1}^{N}$ with $N$ samples, where $x$ denotes the input prompt and $y_w$ and $y_l$ represent the preferred and dispreferred answers, respectively. $r_\phi$ is the reward model trained with the Bradley-Terry model \cite{bradley1952rank} by minimizing the following loss, where $\sigma(\cdot)$ is the sigmoid function: 
\begin{align}
\begin{split}
    & \mathcal{L}_{R(r_\phi,\mathcal{D})} = \\
    & - \mathbb{E}_{(x,y_w,y_l)\sim \mathcal{D}}\Bigr[\log\sigma\big(r_\phi(x,y_w)-r_\phi(x,y_l)\big)\Bigr]
\end{split}
\label{eq:loss}
\end{align}

\paragraph{Policy Optimization.} The policy $\pi_\theta$ is then optimized to maximize expected rewards, guided by the reward function $r_\phi$:
\begin{align}
    \begin{split}
        & \mathcal{J}(\theta)  =  \\
        & \max_{\pi} \mathbb{E}_{\substack{x \sim \mathcal{D} \\ y \sim \pi(y|x)}}\left[ r_\phi(x,y) - \beta \log \frac{\pi_\theta(y|x)}{\pi_{\text{ref}}(y|x)} \right]
    \end{split}
\end{align}
where $\beta$ denotes the weighting factor of the Kullback-Leibler divergence \cite{kullback1951information}, which constrains the policy $\pi_{\theta}$ to remain close to the reference policy $\pi_{\text{ref}}$.

\paragraph{Direct Preference Optimization.} DPO is built on a key theoretical insight: under the optimal policy that maximizes the constrained reward objective \cite{rafailov-etal-2023-dpo}, there exists a closed-form relationship between the reward function and the policy. Specifically, the optimal reward function can be expressed as: \looseness=-1
\begin{align}
\begin{split}
    r_\theta(x,y) = \beta \ln \frac{\pi(y|x)}{\pi_{\text{ref}}(y|x)} + \beta \ln(Z(x))
\end{split}
\label{eq:reward}
\end{align}
where $Z(x) = \sum_y \pi_{ref}(y|x) \cdot \exp (\frac{1}{\beta} r_\phi(x,y))$ is the partition function that ensures the relationship holds across all possible completions. The DPO loss function is derived by substituting the reward function Eq.~\eqref{eq:reward} into the reward modeling objective Eq.~\eqref{eq:loss}:
\begin{align}
\begin{split}
    &\mathcal{L}_{DPO}(\theta) = -\mathbb{E}_{(x,y_w,y_l)\sim \mathcal{D}}\Bigr[\log \sigma\big( \\
    & \beta\log\frac{\pi(y_w|x)}{\pi_{\text{ref}}(y_w|x)}-\beta\log\frac{\pi(y_l|x)}{\pi_{\text{ref}}(y_l|x)}\big)\Bigr]
\end{split}
\end{align}

\section{Direct Preference Optimization Training}
\label{app:training}
\subsection{Training Time}
\label{sec:training_time}
\begin{table}[t]
\centering
\small
\begin{tabular}{l cc}
\toprule
Model & \data{SIB200} & \data{TAXI1500} \\
\midrule
\lm{Gemma3-4B}  & 0:32:47 & 0:23:11 \\
\lm{Qwen3-4B}   & 0:32:21 & 0:23:43 \\
\lm{Qwen3-8B}   & 0:43:22 & 0:32:56 \\
\lm{Gemma3-12B} & 1:12:23 & 0:52:09 \\
\bottomrule
\end{tabular}
\caption{Training time on an NVIDIA H100 GPU for the four models, reported separately on \data{SIB200} and \data{TAXI1500}. Times are formatted as \(\mathrm{hh:mm:ss}\).}
\label{tab:training_time}
\end{table}

Table~\ref{tab:training_time} summarizes the total training time for the four base models on \data{SIB200} and \data{TAXI1500}.

\subsection{Training Detail}

For \data{SIB200} and \data{Taxi1500}, we construct 600 and 800 preference pairs per language, respectively.
We fine-tune the base model using DPO on prompt--chosen--rejected triples. 
To reduce training cost, we adopt LoRA and insert trainable low-rank adapters into both attention and MLP projection layers. 
For multilingual training, we employ a language-level round-robin batch sampler to maintain balanced updates across languages. 
Table~\ref{tab:training-hparams} summarizes the main hyperparameters.

\begin{table}[t]
\centering
\small
\begin{tabular}{lc}
\toprule
\textbf{Hyperparameter} & \textbf{Value} \\
\midrule
DPO learning rate & $1\times10^{-6}$ \\
DPO $\beta$ & 0.2 \\
DPO epochs & 3 \\
Per-device batch size & 7 \\
Gradient accumulation & 7 \\
Warmup ratio & 0.03 \\
LoRA rank & 128 \\
LoRA alpha & 256 \\
LoRA dropout & 0.05 \\
Weight decay & 0.0 \\
\bottomrule
\end{tabular}
\caption{Default hyperparameter settings.}
\label{tab:training-hparams}
\end{table}

\subsection{Total Score Changes by Direct Preference Optimization}
\label{app:score_shift}

\begin{figure}[t]
    \centering
    \begin{subfigure}[b]{\columnwidth}
    \centering
    \includegraphics[width=\columnwidth]{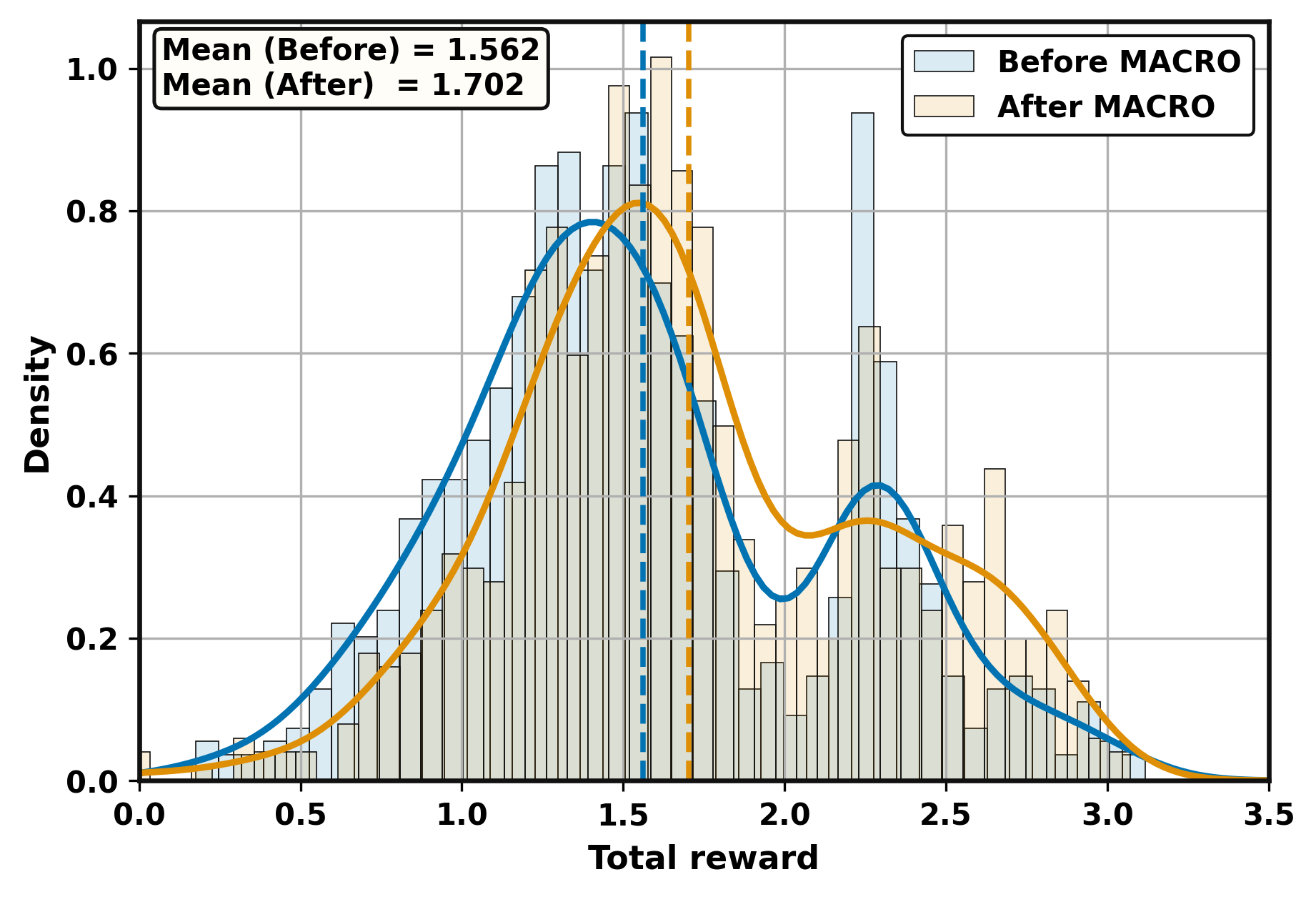}
    \subcaption{\data{TAXI1500}}
    \label{fig:total-reward-qwen3-4b-taxi1500}
    \end{subfigure}
    \begin{subfigure}[b]{\columnwidth}
    \centering
    \includegraphics[width=\columnwidth]{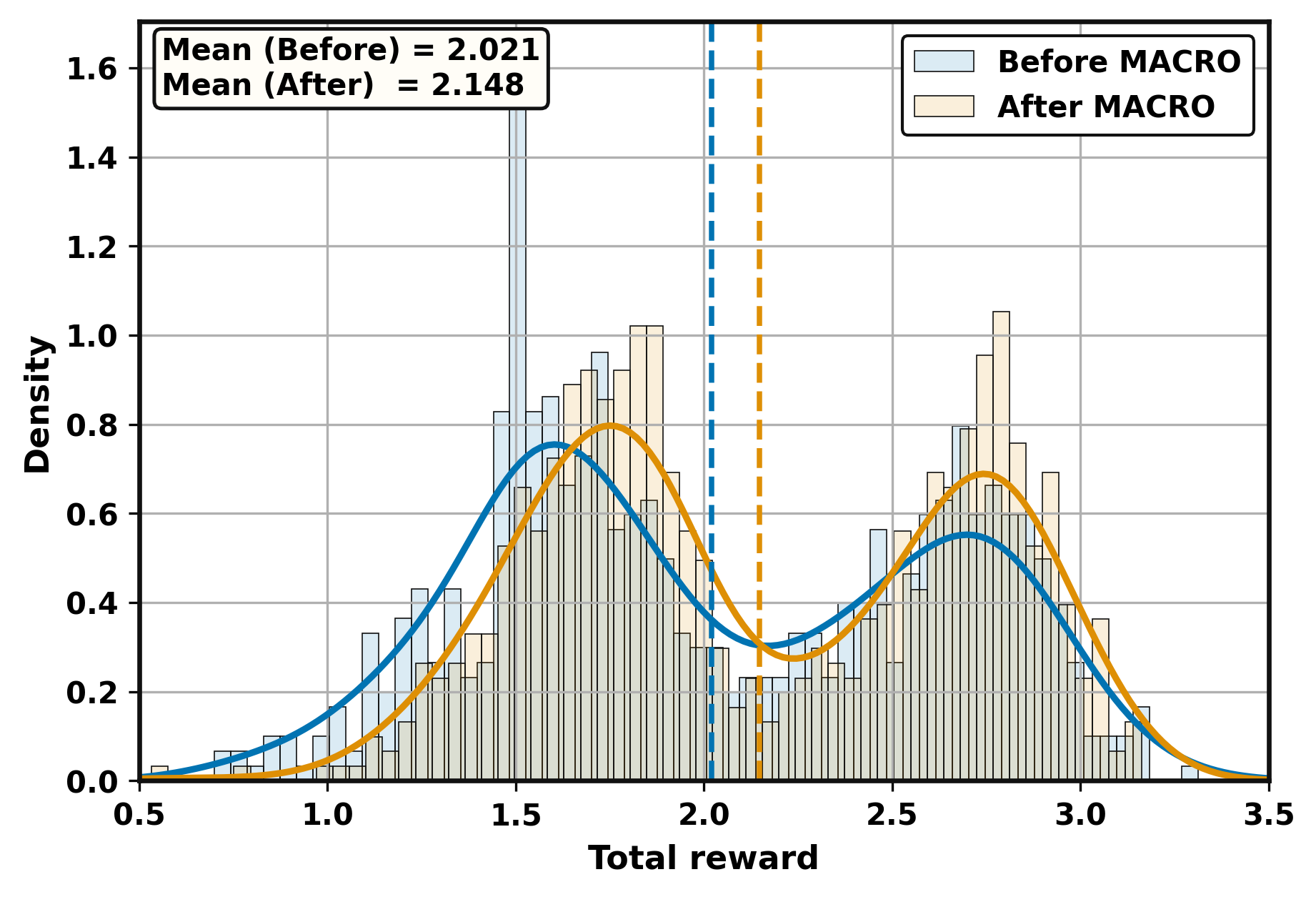}
    \subcaption{\data{SIB200}}
    \label{fig:total-reward-qwen3-4b-sib200}
    \end{subfigure}
    \caption{Total score distributions before and after applying \our using \lm{Qwen3-4B}.}
    \label{fig:total_reward}
\end{figure}

Figure~\ref{fig:total_reward} displays the changes in total score distribution after applying \our. \our can increase overall scores and reduce the variance of the scores while modestly increasing the mean scores after post-training by approximately 9\%, indicating that the model becomes more consistent by elevating lower-performance cases without necessarily improving peak performance.

\subsection{Scoring Weight Configuration}
\label{app:reward_weight}

To compute the total preference score in \S \ref{subsubsec:total_reward}, we assign weights to the three scoring components, i.e., \Rflip, \Raug, and \Redit. Table~\ref{tab:reward_weight} summarizes the scoring weights used for each model on \data{SIB200} and \data{TAXI1500}.

\begin{table}[t]
    \centering
    \small
    \setlength{\tabcolsep}{5pt}
    \begin{tabular}{llccc}
        \toprule
        Dataset & Model & $w_{\text{flip}}$ & $w_{\text{aug}}$ & $w_{\text{edit}}$ \\
        \midrule
        \data{SIB200}   & \lm{Qwen3-4B}   & 1.0 & 0.6 & 1.0 \\
        \data{SIB200}   & \lm{Qwen3-8B}   & 1.0 & 2.0 & 1.0 \\
        \data{SIB200}   & \lm{Gemma3-4B}  & 1.0 & 1.35 & 1.0 \\
        \data{SIB200}   & \lm{Gemma3-12B} & 1.0 & 1.2 & 1.0 \\
        \midrule
        \data{TAXI1500} & \lm{Qwen3-4B}   & 1.0 & 1.0 & 1.0 \\
        \data{TAXI1500} & \lm{Qwen3-8B}   & 1.0 & 1.0 & 1.0 \\
        \data{TAXI1500} & \lm{Gemma3-4B}  & 1.0 & 1.0 & 1.0 \\
        \data{TAXI1500} & \lm{Gemma3-12B} & 1.0 & 0.5 & 1.0 \\
        \bottomrule
    \end{tabular}
    \caption{Scoring weights used to compute the total preference score for each model and dataset.}
    \label{tab:reward_weight}
\end{table}

\section{Dataset Information}
\label{app:dataset}

\subsection{Dataset Distribution}
\begin{figure}[t]
    \centering
    \begin{subfigure}[t]{\linewidth}
        \centering
        \includegraphics[width=\linewidth]{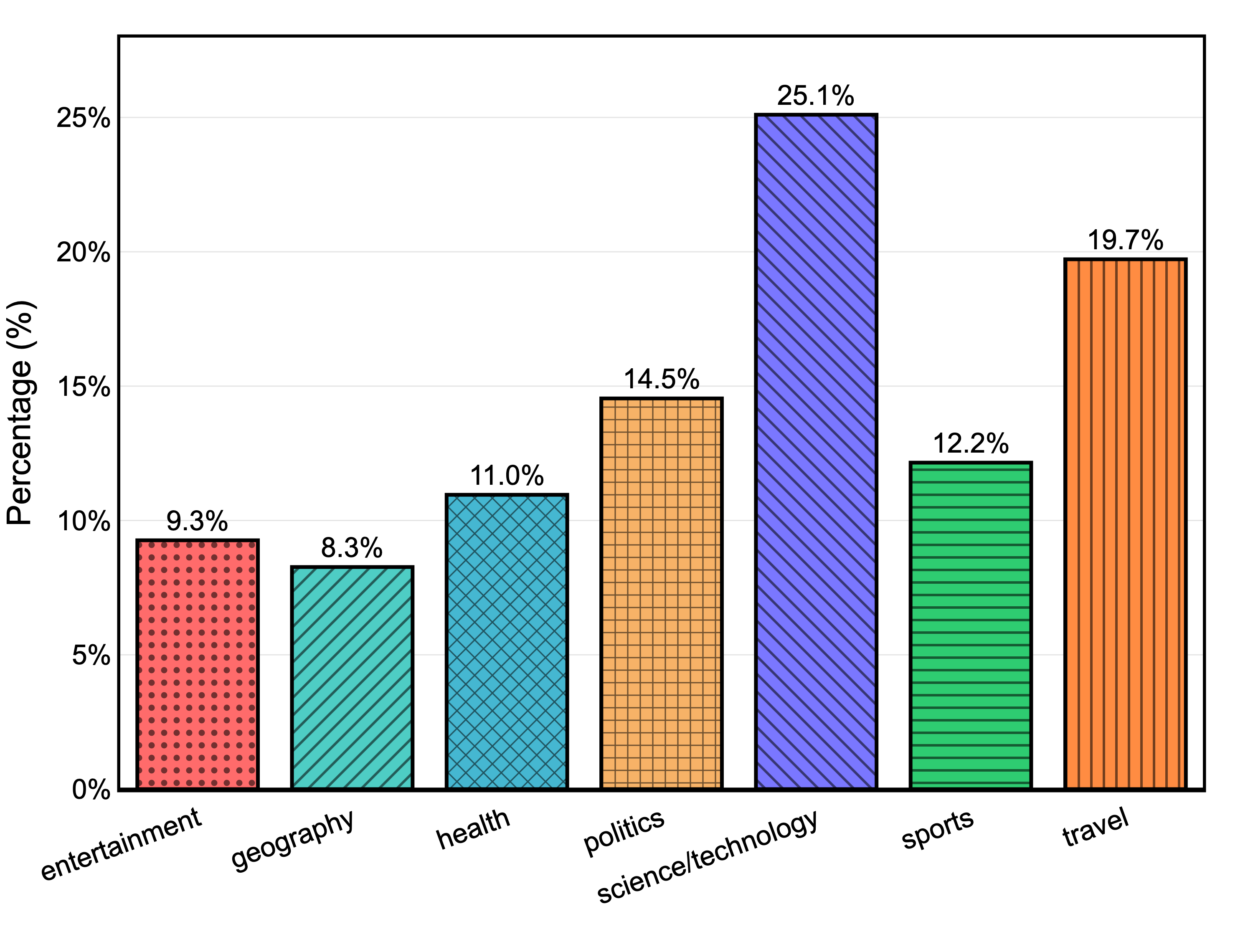}
        \caption{\data{SIB200} label distribution.}
        \label{fig:sib200_distribution}
    \end{subfigure}
    
    \vspace{0.5em}
    
    \begin{subfigure}[t]{\linewidth}
        \centering
        \includegraphics[width=\linewidth]{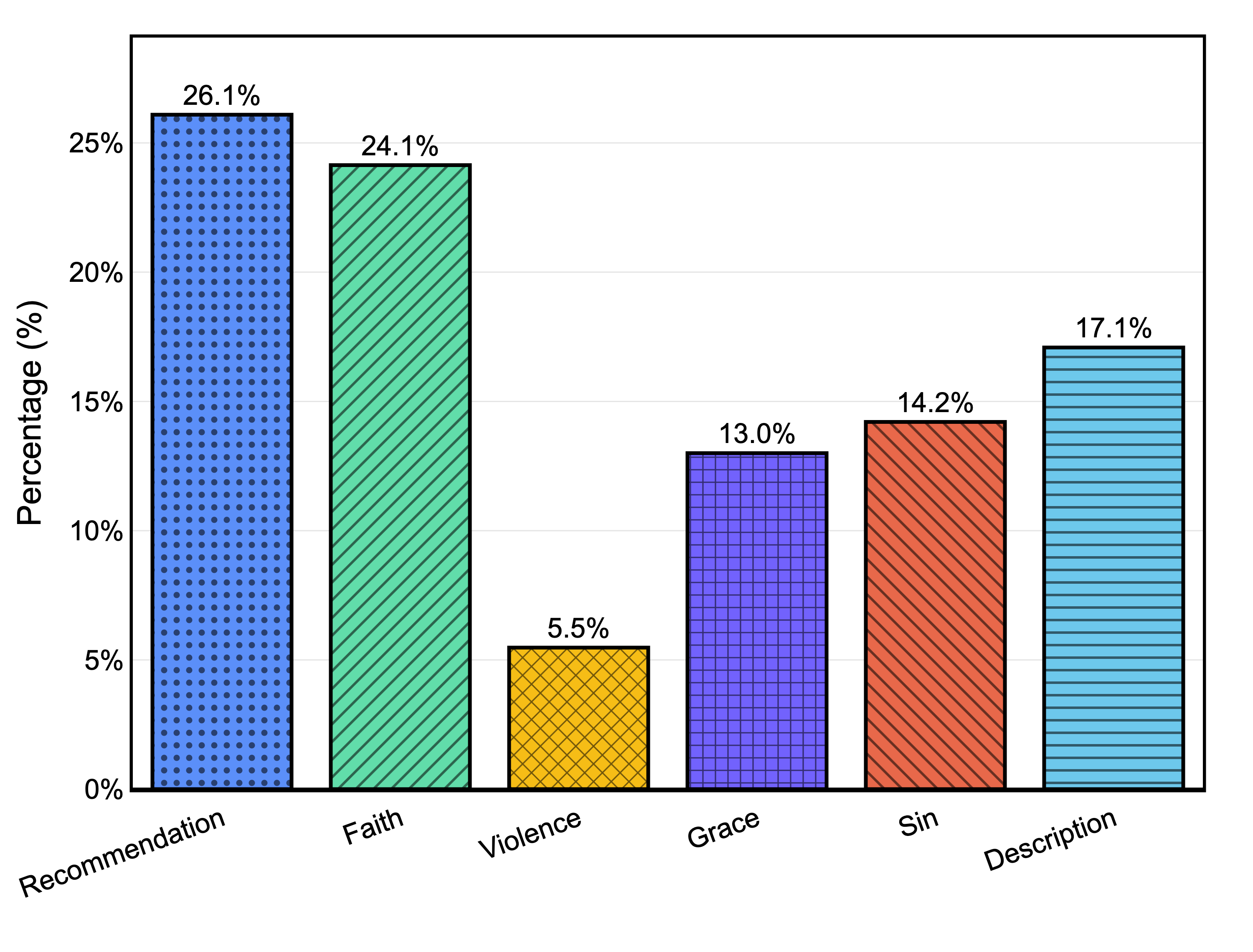}
        \caption{\data{TAXI1500} label distribution.}
        \label{fig:taxi1500_distribution}
    \end{subfigure}
    
    \caption{Label distributions of the two evaluation datasets. Subfigure (a) shows \data{SIB200} and subfigure (b) shows \data{TAXI1500}.}
    \label{fig:dataset_distribution}
\end{figure}

Figure~\ref{fig:dataset_distribution} reports the label distributions of \data{SIB200} and \data{TAXI1500}, where percentages are computed over all splits combined.

\subsection{Dataset Example}
\begin{figure*}[t]
    \centering
    \begin{subfigure}{\linewidth}
        \centering
        \includegraphics[width=0.9\linewidth]{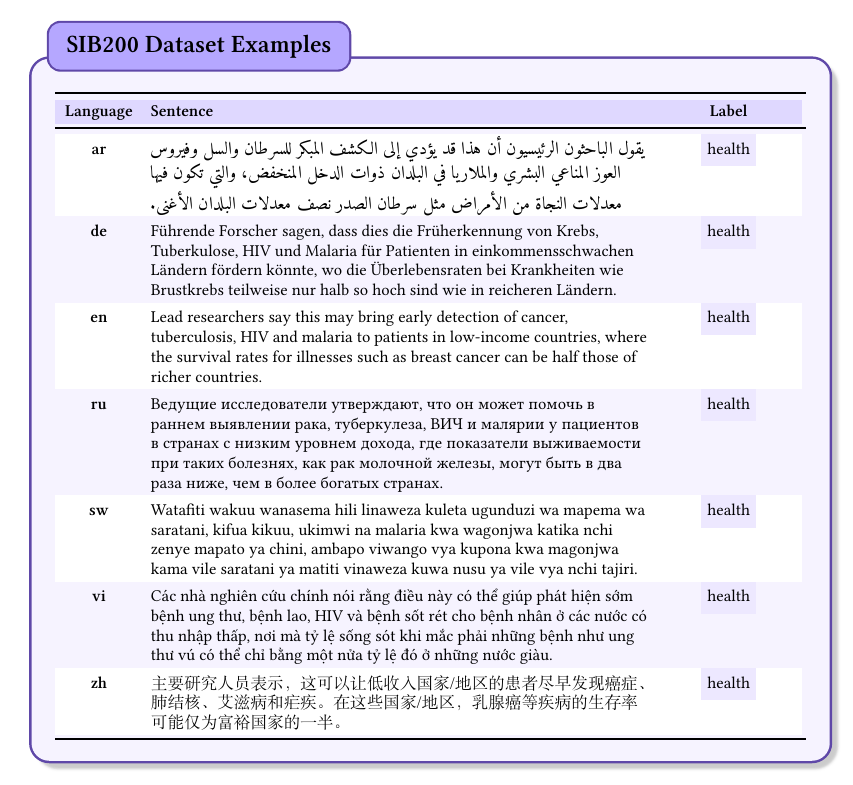}
        \caption{Dataset Examples of \data{SIB200}.}
        \label{fig:sib200_examples}
    \end{subfigure}
        
    \begin{subfigure}{\linewidth}
        \centering
        \includegraphics[width=0.9\linewidth]{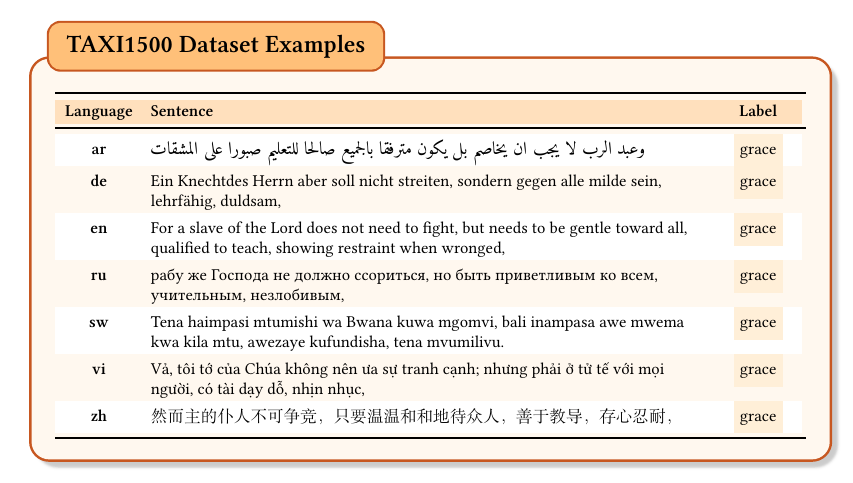}
        \caption{Dataset Examples of \data{TAXI1500}.}
        \label{fig:taxi1500_examples}
    \end{subfigure}
    \caption{Dataset examples.}
\end{figure*}

Figures~\ref{fig:sib200_examples} and \ref{fig:taxi1500_examples} report representative examples from \data{SIB200} and \data{TAXI1500}, respectively.

\section{Model}
\label{app:model}
\begin{table*}[t!]
    \centering
    \begin{tabular}{cccc}

    \toprule
    \textbf{Name}& \textbf{Citation} & \textbf{Size} & \textbf{Link}\\

    \midrule
    \lm{Qwen3} & \citet{yang2025qwen3technicalreport} & 4B & \url{https://huggingface.co/Qwen/Qwen3-4B} \\
    \lm{Qwen3} & \citet{yang2025qwen3technicalreport} & 8B & \url{https://huggingface.co/Qwen/Qwen3-8B} \\
    \lm{Gemma3} & \citet{gemmateam2025gemma3technicalreport} & 4B & \url{https://huggingface.co/google/gemma-3-4b-it}\\
    \lm{Gemma3} & \citet{gemmateam2025gemma3technicalreport} & 12B & \url{https://huggingface.co/google/gemma-3-12b-it}\\

    \bottomrule
    \end{tabular}
        
    \caption{
    Four open-source LLMs are used for counterfactual generation.
    }
    \label{tab:used_model}
\end{table*}
\subsection{Model Information}
Table~\ref{tab:used_model} shows information about the models we employ, including model parameter sizes, citations, and their corresponding Hugging Face link.  

\subsection{Inference Time}
Table~\ref{tab:inference_time} reports the inference time of \our using \lm{Qwen3-8B} on \data{SIB200} and \data{TAXI1500}, broken down by language.

\begin{table}[t]
\centering
\small
\begin{tabular}{lcc}
\toprule
Language & SIB200 & TAXI1500 \\
\midrule
en & 00:06:50 & 00:03:09 \\
de & 00:09:35 & 00:04:07 \\
zh & 00:07:11 & 00:03:22 \\
ar & 00:10:12 & 00:03:43 \\
vi & 00:09:44 & 00:04:08 \\
sw & 00:10:30 & 00:04:35 \\
ru & 00:11:47 & 00:04:47 \\
\bottomrule
\end{tabular}
\caption{Inference time of \our{} using \lm{Qwen3-8B} on \data{SIB200} and \data{TAXI1500}, reported separately by language. Times are formatted as hh:mm:ss.}
\label{tab:inference_time}
\end{table}

\section{Prompt}
\label{sec:prompt}
\subsection{Prediction Prompt}
\begin{figure*}[t]
    \centering
    \begin{subfigure}[b]{\textwidth}
    \centering
    \includegraphics[width=\textwidth]{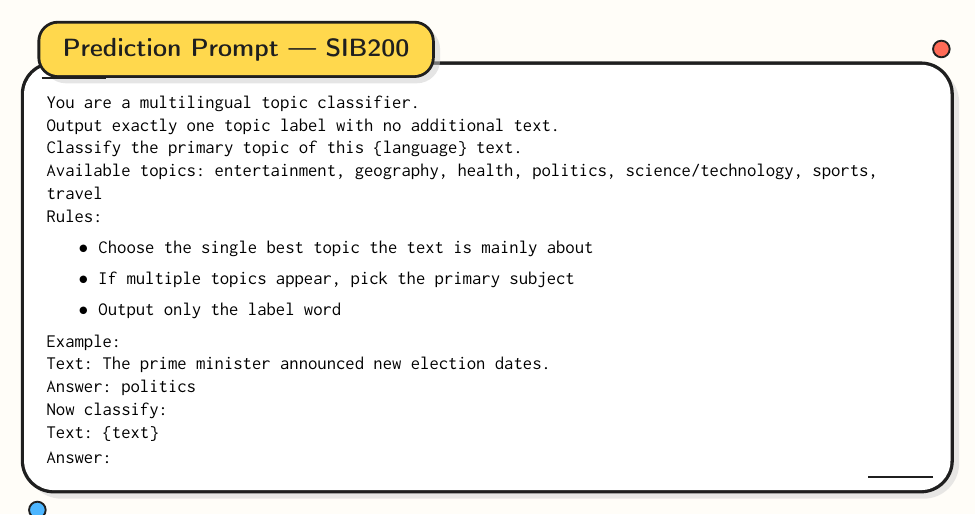}
    \subcaption{\data{SIB200}}
    \label{fig:prompt_pred_sib200}
    \end{subfigure}
    \begin{subfigure}[b]{\textwidth}
    \centering
    \includegraphics[width=\textwidth]{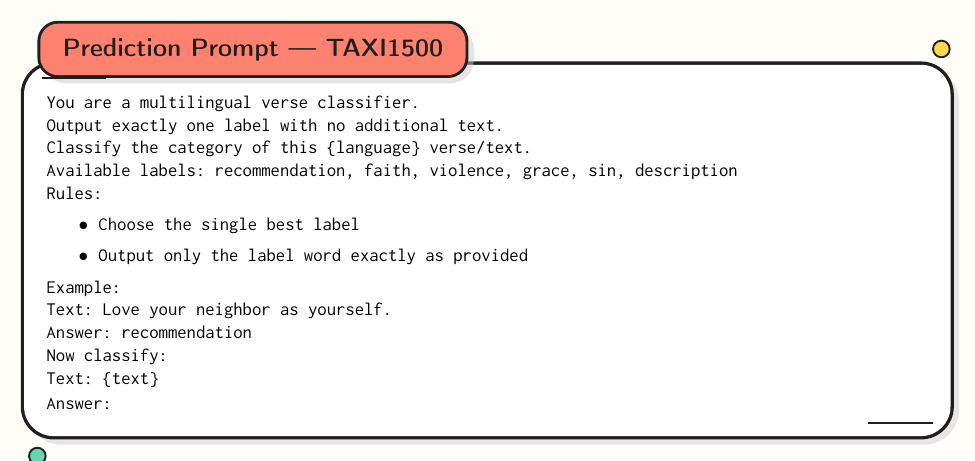}
    \subcaption{\data{TAXI1500}}
    \label{fig:prompt_pred_taxi1500}
    \end{subfigure}
    \caption{Prediction prompts used for the two evaluation datasets: \data{SIB200} and \data{TAXI1500}. Since the tasks differ across datasets, we adopt dataset-specific prompts for label prediction.}
    \label{fig:prompt_pred}
\end{figure*}

Figure \ref{fig:prompt_pred} reports the prediction prompts used for \data{SIB200} and \data{TAXI1500}.

\subsection{Counterfactual Generation Prompt}
\begin{figure*}[t]
    \centering
    \begin{subfigure}[b]{\textwidth}
    \centering
    \includegraphics[width=\textwidth]{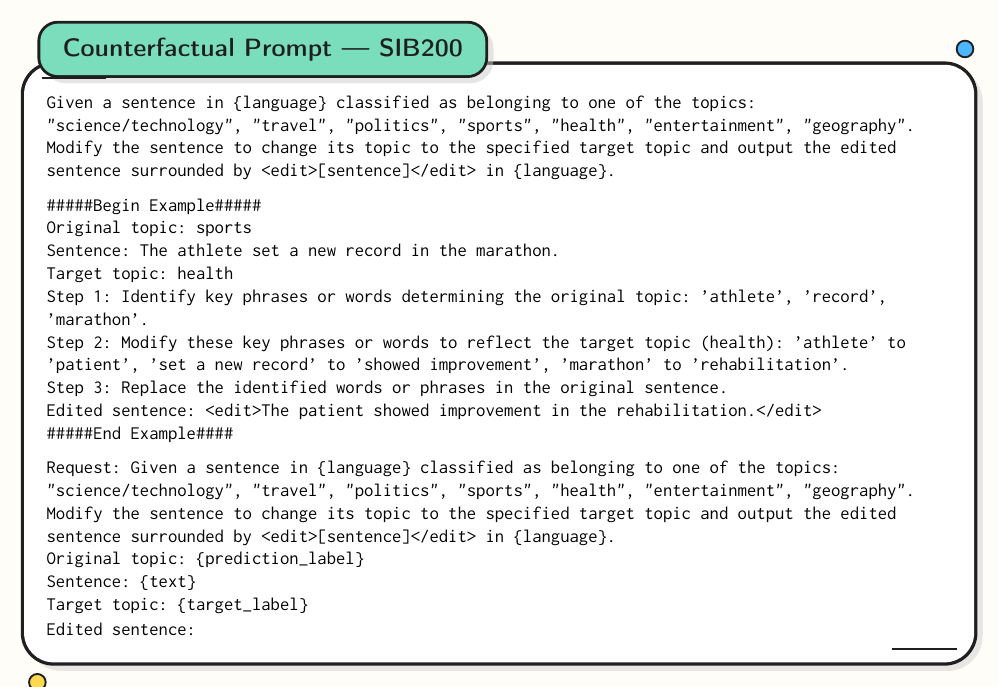}
    \subcaption{\data{SIB200}}
    \label{fig:prompt_cf_sib200}
    \end{subfigure}
    \begin{subfigure}[b]{\textwidth}
    \centering
    \includegraphics[width=\textwidth]{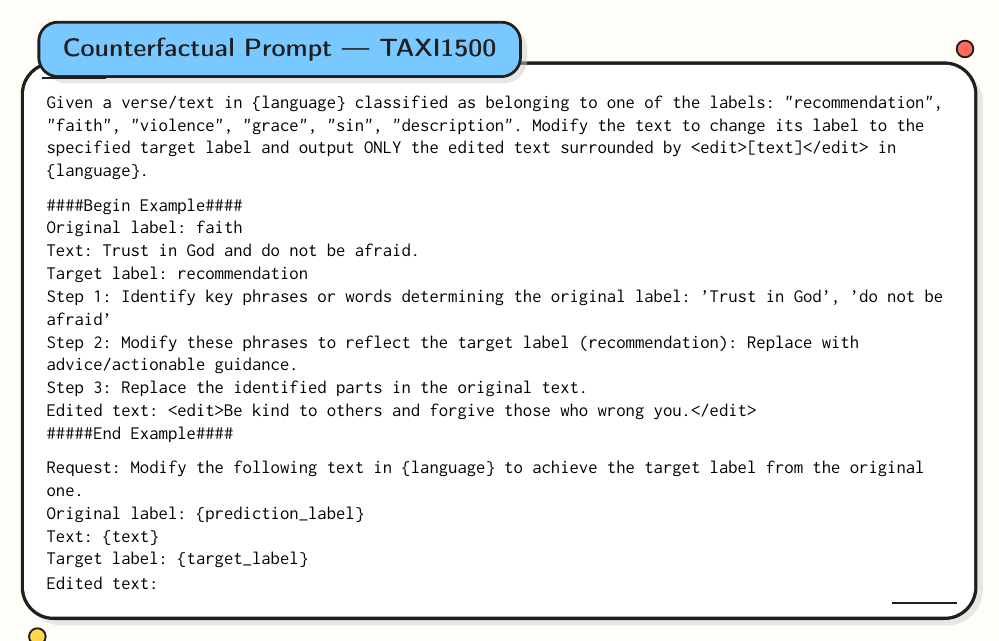}
    \subcaption{\data{TAXI1500}}
    \label{fig:prompt_cf_taxi1500}
    \end{subfigure}
    \caption{Counterfactual generation prompts used for \data{SIB200} and \data{TAXI1500}. For each dataset, both DG-CF and \our use the same generation prompt.}
    \label{fig:prompt_cf}
\end{figure*}

Figure~\ref{fig:prompt_cf} reports counterfactual generation prompts used in our experiments. For a fair comparison, DG-CF and \our use the same prompt template.

\section{Why Not Multilingual Alignment?}
\label{app:multilingual_align}
While one might intuitively consider apply multilingual alignment to enhance performance in non-dominant languages, it is challenging to identify a target dominant language with which other languages align, since English is not necessarily the optimal languages (\S\ref{subsec:automatic_evaluation}) unlike in other tasks, e.g., reasoning \cite{qi-etal-2025-models} or translation \cite{etxaniz-etal-2024-multilingual}. Nevertheless, we follow MAPO \cite{she-etal-2024-mapo} to perform cross-lingual alignment using the conditional generation probability from a translation model. We conduct a small-scale ablation study by \textit{including} an additional reward for multilingual alignment and check how it affects the overall counterfactual quality. 

\subsection{Alignment Score ($\mathcal{R}_{\text{align}}$)}
To address the performance gap between counterfactuals in dominant and non-dominant languages \cite{wang-etal-2026-parallel}, we optimize the latter relative to the former. We designate English as the dominant language throughout, given its prevalence in both counterfactual generation and translation. To ensure semantic consistency in multilingual settings, we measure cross-lingual alignment using the conditional generation probability from an off-the-shelf multilingual translation model \lm{NLLB-200}\footnote{\url{https://huggingface.co/facebook/nllb-200-distilled-1.3B}} \cite{koishekenov-etal-2023-memory}, following MAPO \cite{she-etal-2024-mapo}. Concretely, let $\tilde{x}_{\textsf{en}}$ denote the English counterfactual
anchor and $\tilde{x}_{\textsf{non}}$ the generated non-English counterfactual. We compute the token-level
log-likelihood of generating $\tilde{x}_{\textsf{en}}$ conditioned on $\tilde{x}_{\textsf{non}}$ via teacher forcing:
\begin{align}
\begin{split}
    \mathcal{L}_{\text{CE}}&(\tilde{x}_{\textsf{en}},\tilde{x}_{\textsf{non}})
    = \\
    &- \frac{1}{T}\sum_{t=1}^{T}\log P_{\text{NLLB}}(\tilde{x}_{\textsf{en}}^t \mid \tilde{x}_{\textsf{en}}^{<t}, \tilde{x}_{\textsf{non}})
\end{split}
\end{align}
We define the alignment reward $\mathcal{R}_{\text{align}}$ as the NLLB model probability:

\begin{equation}
    \mathcal{R}_{\text{align}}(\tilde{x}_{\textsf{en}}, \tilde{x}_{\textsf{non}}) = \sigma\big(-{\mathcal{L}_{\text{CE}}(\tilde{x}_{\textsf{en}}, \tilde{x}_{\textsf{non}})} \big)
\end{equation}
Higher $\mathcal{R}_{\text{align}}$ indicates greater probability of translating the non-English counterfactual $\tilde{x}_{\textsf{non}}$ back to the English anchor $\tilde{x}_{\textsf{en}}$, thus reflecting stronger cross-lingual semantic alignment. Subsequently, we add $\mathcal{R}_{\text{align}}$ to the total reward $\mathcal{R}_{\text{total}}$ (\S\ref{subsubsec:total_reward}).

\subsection{Results and Discussion}
Table \ref{tab:automatic_evaluation_ablation_align} shows that although counterfactual quality in terms of \textit{validity} and \textit{fluency} in some languages is enhanced (Swahili and Russian on \data{SIB200}), the alignment generally harms the other languages. In particular, \textit{minimality} is noticeably degraded after multilingual alignment. Although one could consistently use the best-performing language as the dominant one, prior knowledge is needed to determine the optimal language on a specific dimension by evaluating DG-CFs, which is \textit{task- and model-specific}. Furthermore, as we reported in \S\ref{subsec:automatic_evaluation}, no single language outperforms others across all evaluation dimensions. These factors make multilingual alignment largely \textbf{infeasible} and \textbf{ineffective}. Therefore, we discard the use of multilingual alignment.

\begin{table*}[t!]
  \centering
  \footnotesize
  \setlength{\tabcolsep}{5.5pt}
  \renewcommand{\arraystretch}{1.08}

  \resizebox{2\columnwidth}{!}{%
  \begin{tabular}{@{}r l ccccc ccccc@{}}
    \toprule
    & & \multicolumn{5}{c}{\textbf{\data{SIB200}}}
      & \multicolumn{5}{c}{\textbf{\data{Taxi1500}}} \\
    \cmidrule(lr){3-7}\cmidrule(lr){8-12}
    \textbf{Setting} & \textbf{Lang}
      & \textbf{SLFR} \(\uparrow\) & \textbf{HLFR} \(\uparrow\) & \textbf{PPL} \(\downarrow\) & \textbf{SS} \(\uparrow\) & \textbf{TS} \(\downarrow\)
      & \textbf{SLFR} \(\uparrow\) & \textbf{HLFR} \(\uparrow\) & \textbf{PPL} \(\downarrow\) & \textbf{SS} \(\uparrow\) & \textbf{TS} \(\downarrow\)  \\
    \midrule


    \multirow{7}{*}{\textbf{w/ align}} & \textsf{en} & \cell{0.566}{0.171}{-} & \cell{0.384}{0.172}{-} & \cell{36.39}{1.48}{-} & \cell{0.690}{0.027}{+} & \cell{0.274}{0.034}{-} & \cell{0.676}{0.045}{-} & \cell{0.523}{0.009}{-} & \cell{24.27}{0.78}{-} & \cell{0.691}{0.001}{+} & \cell{0.477}{0.009}{+} \\
                                 & \textsf{de} & \cell{0.697}{0.030}{-} & \cell{0.596}{0.010}{+} & \cell{26.48}{1.28}{-} & \cell{0.687}{0.018}{-} & \cell{0.280}{0.021}{+} & \cell{0.748}{0.036}{+} & \cell{0.477}{0.028}{-} & \cell{23.07}{0.20}{+} & \cell{0.688}{0.030}{+} & \cell{0.498}{0.046}{-} \\
                                 & \textsf{zh} & \cell{0.697}{0.000}{0} & \cell{0.626}{0.000}{0} & \cell{34.73}{0.19}{-} & \cell{0.668}{0.035}{-} & \cell{0.589}{0.023}{+} & \cell{0.667}{0.009}{+} & \cell{0.405}{0.027}{+} & \cell{34.20}{0.57}{+} & \cell{0.707}{0.002}{+} & \cell{0.846}{0.019}{-} \\
                                 & \textsf{ar} & \cell{0.636}{0.122}{-} & \cell{0.525}{0.121}{-} & \cell{29.72}{0.80}{-} & \cell{0.626}{0.027}{-} & \cell{0.295}{0.014}{+} & \cell{0.775}{0.018}{+} & \cell{0.441}{0.000}{0} & \cell{40.06}{2.26}{+} & \cell{0.645}{0.019}{+} & \cell{0.612}{0.026}{-} \\
                                 & \textsf{vi} & \cell{0.697}{0.020}{-} & \cell{0.636}{0.080}{+} & \cell{23.83}{0.72}{+} & \cell{0.702}{0.005}{-} & \cell{0.307}{0.003}{+} & \cell{0.712}{0.072}{-} & \cell{0.477}{0.037}{-} & \cell{18.28}{0.30}{+} & \cell{0.611}{0.026}{-} & \cell{0.620}{0.019}{+} \\
                                 & \textsf{sw} & \cell{0.576}{0.010}{+} & \cell{0.182}{0.081}{-} & \cell{20.21}{0.48}{+} & \cell{0.774}{0.019}{-} & \cell{0.231}{0.524}{-} & \cell{0.261}{0.009}{-} & \cell{0.063}{0.018}{-} & \cell{16.48}{0.30}{-} & \cell{0.829}{0.000}{0} & \cell{0.776}{0.474}{-} \\
                                 & \textsf{ru} & \cell{0.687}{0.030}{+} & \cell{0.535}{0.030}{+} & \cell{21.29}{0.24}{-} & \cell{0.684}{0.025}{-} & \cell{0.306}{0.023}{+} & \cell{0.559}{0.054}{-} & \cell{0.378}{0.018}{-} & \cell{20.10}{0.13}{-} & \cell{0.718}{0.012}{-} & \cell{0.593}{0.002}{+} \\

    \bottomrule
  \end{tabular}
  }

  \caption{Ablation study with align reward.}
  \label{tab:automatic_evaluation_ablation_align}
\end{table*}

\section{Counterfactual Examples}
\label{app:cf_examples}
\begin{figure*}[t]
  \centering
  \includegraphics[width=\linewidth]{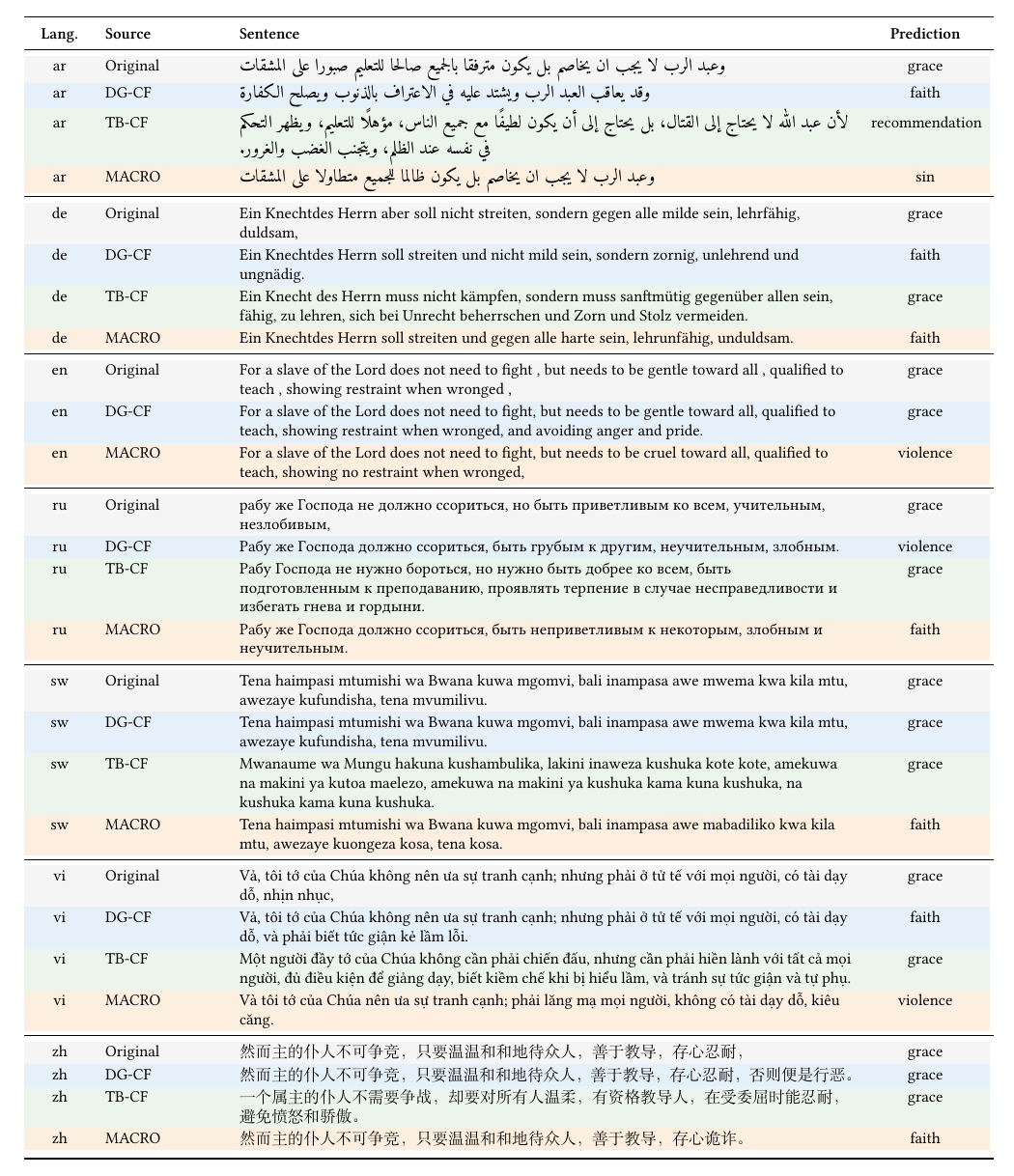}

  \captionof{table}{
    Multilingual counterfactual examples from the \data{TAXI1500} task.
    The figure shows original Bible verses and their labels together with counterfactual samples generated by three methods (DG-CF, TB-CF, and \our{}) across multiple languages using \lm{Qwen3-8B}.
  }
  \label{fig:taxi1500-macro-example}
\end{figure*}

\begin{figure*}[t]
  \centering
  \includegraphics[width=\linewidth]{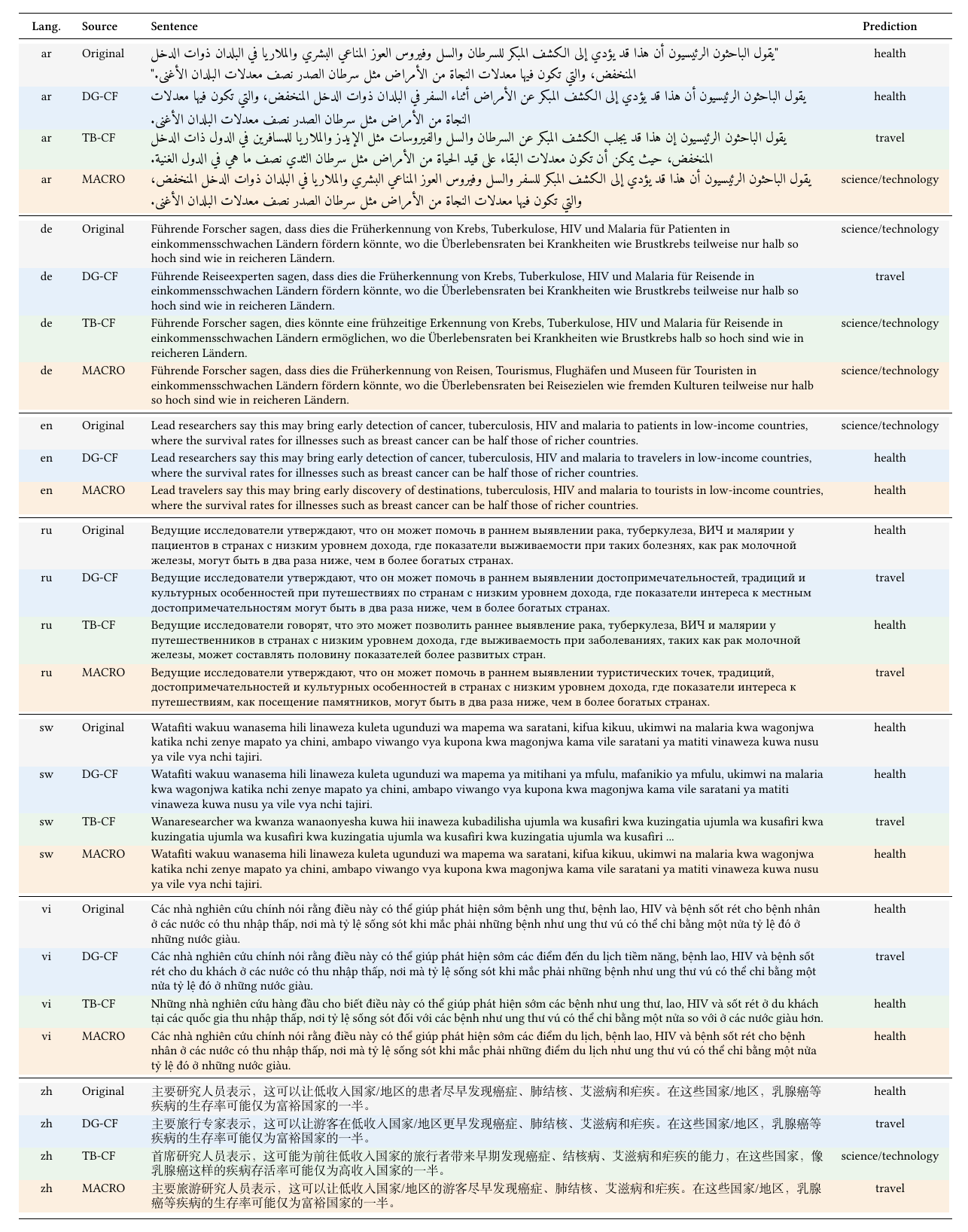}

  \captionof{table}{
    Multilingual counterfactual examples from the \data{SIB200} task.
    The figure shows original sentences and their labels together with counterfactual samples generated by three methods (DG-CF, TB-CF, and \our{}) across multiple languages using \lm{Qwen3-8B}.
  }
  \label{fig:sib200-macro-example}
\end{figure*}

We present multilingual examples from \data{TAXI1500} in Table~\ref{fig:taxi1500-macro-example} and from \data{SIB200} in Table~\ref{fig:sib200-macro-example}. These examples illustrate the original inputs and counterfactuals produced by DG-CF, TB-CF, and \our{}.

\begin{figure*}[t]
  \centering
  \includegraphics[width=\linewidth]{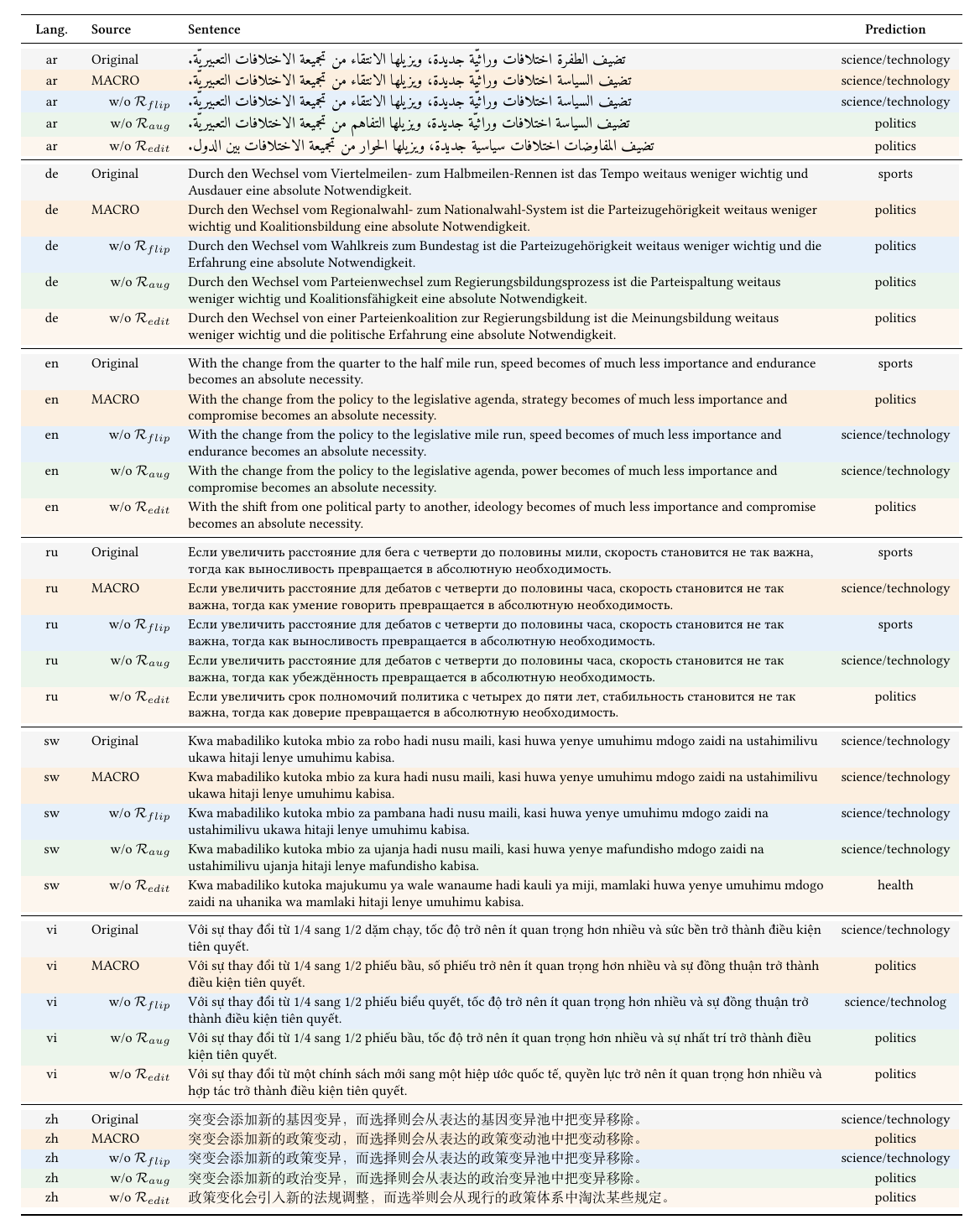}
  \captionof{table}{
    Multilingual counterfactual examples from the \data{SIB200} task.
    The figure shows original sentences and their labels together with counterfactual samples generated by three methods (\our{} and its different reward ablations) across multiple languages using \lm{Qwen3-8B}.
  }
  \label{fig:sib200-macro-ablation-example}
\end{figure*}

\begin{figure*}[t]
  \centering
  \includegraphics[width=\linewidth]{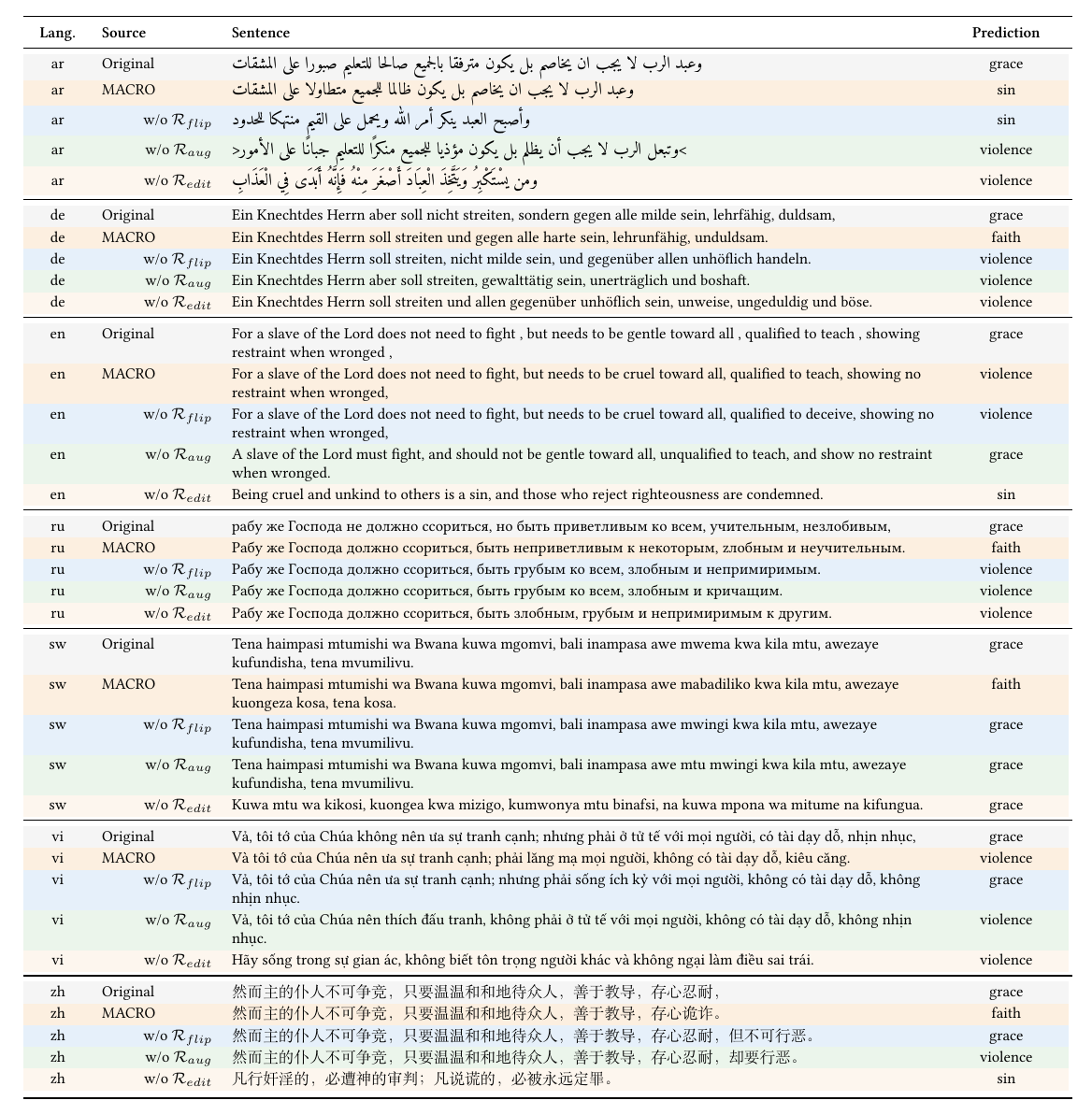}
  \captionof{table}{
    Multilingual counterfactual examples from the \data{TAXI1500} task.
    The figure shows original sentences and their labels together with counterfactual samples generated by three methods (\our{} and its different reward ablations) across multiple languages using \lm{Qwen3-8B}.
  }
  \label{fig:taxi1500-macro-ablation-example}
\end{figure*}

\section{Additional Counterfactual Evaluation Metrics}
\label{app:additional_metrics}
Table~\ref{tab:hlfr_ts_dgcf_macro_side_by_side} reports the additional automatic evaluation metrics of the experiments in Table~\ref{tab:automatic_evaluation_sib200_taxi1500_main}, namely hard label flipping rate and textual similarity. These two metrics complement the main results by further evaluating the success of target-label flipping and the degree of input text modification.

\subsection{Hard Label Flipping Rate}
\label{subsec:hlfr}
\textbf{Hard Label Flipping Rate (HLFR)} measures validity, defined as the fraction of SCEs that successfully change the model prediction to the targeted label $\hat{y}$. The same model $\mathcal{M}$ is used both for generating SCE $\tilde{x}$ and for evaluating whether the predicted label is shifted. Formally, on $N$ evaluation instances, HLFR is computed as:
\begin{align*}
\mathrm{HLFR}
= \frac{1}{N}\sum_{i=1}^{N}\mathbb{I}\big[\mathcal{M}(\tilde{x}_{i}) = \hat{y}\big]
\end{align*}
where $\mathbb{I}[\cdot]$ is an indicator function.

\subsection{Textual Similarity}
\label{subsec:ts}
\textbf{Textual Similarity (TS)} measures the extent of modifications made to the original input $x$. Following \citet{dehghanighobadi-etal-2025-llms, wang-etal-2025-fitcf, bhattacharjee-etal-2024-zero}, we employ a normalized Levenshtein distance $d$, which captures all applied edits. Formally, over $N$ evaluation instances, TS is computed as:
\begin{align*}
    \mathrm{TS} = \frac{1}{N}\sum_{i=1}^{N}\frac{d(x_i,\tilde{x}_i)}{|x_i|}
\end{align*}
where $d(x_i, \tilde{x}_i)$ is the Levenshtein distance between the original input $x_i$ and its SCE $\tilde{x}_i$, and $|x_i|$ is the length of the original input used for normalization.

\begin{table*}[t!]
\centering
\renewcommand*{\arraystretch}{1}
\begin{tabular}{cc|cc|cc||cc|cc}
\toprule[1.5pt]
\multicolumn{2}{c|}{\textbf{Dataset}} & \multicolumn{4}{c||}{\textbf{\data{SIB200}}} & \multicolumn{4}{c}{\textbf{\data{TAXI1500}}}\\
\cmidrule(l){1-10}
\multirow{2}{*}{\rotatebox[origin=c]{90}{\scriptsize{\textbf{Model}}}} & \textbf{Lang-} & \multicolumn{2}{c|}{\textbf{HLFR} ($\uparrow$)} & \multicolumn{2}{c||}{\textbf{TS} ($\downarrow$)} & \multicolumn{2}{c|}{\textbf{HLFR} ($\uparrow$)} & \multicolumn{2}{c}{\textbf{TS} ($\downarrow$)}\\
& \textbf{uage} & \textbf{DG-CF} & \textbf{\sys{Macro}} & \textbf{DG-CF} & \textbf{\sys{Macro}} & \textbf{DG-CF} & \textbf{\sys{Macro}} & \textbf{DG-CF} & \textbf{\sys{Macro}}\\
\midrule
\centering \multirow{7}{*}{\rotatebox[origin=c]{90}{\lm{Gemma3-4B}}} & \textsf{en} & 0.559 & \textbf{0.576} & 0.529 & \textbf{0.427} & 0.434 & \textbf{\uwave{0.514}} & 0.670 & \textbf{0.606}\\
 & \textsf{de} & 0.499 & \textbf{0.545} & \uwave{0.389} & \textbf{\uwave{0.355}} & 0.405 & \textbf{0.450} & \uwave{0.580} & \textbf{\uwave{0.551}}\\
 & \textsf{zh} & \uwave{0.629} & \textbf{\uwave{0.778}} & \textbf{0.467} & 0.666 & \textbf{\uwave{0.405}} & 0.369 & \textbf{0.807} & 0.961\\
 & \textsf{ar} & 0.548 & \textbf{0.646} & 0.401 & \textbf{0.399} & 0.342 & \textbf{0.351} & 0.716 & \textbf{0.659}\\
 & \textsf{vi} & 0.508 & \textbf{0.586} & 0.456 & \textbf{0.437} & \textbf{0.377} & 0.369 & 0.631 & \textbf{0.610}\\
 & \textsf{sw} & 0.549 & \textbf{0.606} & \textbf{0.446} & 0.473 & 0.357 & \textbf{0.360} & 0.647 & \textbf{0.609}\\
 & \textsf{ru} & 0.514 & \textbf{0.606} & 0.459 & \textbf{0.435} & 0.394 & \textbf{0.432} & \textbf{0.650} & 0.666\\
\midrule

\centering \multirow{7}{*}{\rotatebox[origin=c]{90}{\lm{Qwen3-4B}}} & \textsf{en} & 0.485 & \textbf{0.618} & \textbf{0.251} & 0.294 & \uwave{0.485} & \textbf{\uwave{0.505}} & 0.480 & \textbf{0.439}\\
 & \textsf{de} & 0.544 & \textbf{0.578} & 0.276 & \textbf{0.263} & 0.284 & \textbf{0.333} & 0.444 & \textbf{0.439}\\
 & \textsf{zh} & \textbf{\uwave{0.765}} & \uwave{0.662} & \textbf{0.459} & 0.574 & 0.309 & \textbf{0.387} & \textbf{0.589} & 0.769\\
 & \textsf{ar} & 0.569 & \textbf{0.574} & \textbf{0.255} & 0.310 & 0.342 & \textbf{0.378} & 0.615 & \textbf{0.556}\\
 & \textsf{vi} & 0.627 & \textbf{0.637} & \textbf{0.318} & 0.337 & \textbf{0.356} & 0.351 & 0.485 & \textbf{0.522}\\
 & \textsf{sw} & 0.127 & \textbf{0.137} & 0.434 & \textbf{\uwave{0.242}} & \textbf{0.107} & 0.090 & \textbf{\uwave{0.312}} & \uwave{0.320}\\
 & \textsf{ru} & 0.373 & \textbf{0.564} & \textbf{\uwave{0.185}} & 0.273 & 0.315 & \textbf{0.414} & 0.526 & \textbf{0.518}\\
\midrule

\centering \multirow{7}{*}{\rotatebox[origin=c]{90}{\lm{Qwen3-8B}}} & \textsf{en} & 0.489 & \textbf{0.556} & \textbf{0.231} & 0.308 & 0.387 & \textbf{\uwave{0.532}} & \uwave{0.474} & \textbf{\uwave{0.468}}\\
 & \textsf{de} & 0.568 & \textbf{0.586} & 0.263 & \textbf{\uwave{0.259}} & 0.342 & \textbf{0.505} & 0.562 & \textbf{0.544}\\
 & \textsf{zh} & 0.534 & \textbf{0.626} & \textbf{0.320} & 0.566 & 0.360 & \textbf{0.378} & \textbf{0.689} & 0.865\\
 & \textsf{ar} & 0.505 & \textbf{\uwave{0.646}} & \textbf{\uwave{0.169}} & 0.281 & 0.387 & \textbf{0.441} & \textbf{0.622} & 0.638\\
 & \textsf{vi} & \textbf{\uwave{0.614}} & 0.556 & 0.308 & \textbf{0.304} & \uwave{0.405} & \textbf{0.514} & 0.621 & \textbf{0.601}\\
 & \textsf{sw} & 0.244 & \textbf{0.263} & \textbf{0.324} & 0.755 & 0.009 & \textbf{0.081} & \textbf{0.693} & 1.250\\
 & \textsf{ru} & \textbf{0.572} & 0.505 & 0.295 & \textbf{0.283} & 0.270 & \textbf{0.396} & \textbf{0.566} & 0.591\\
\midrule

 \centering \multirow{7}{*}{\rotatebox[origin=c]{90}{\lm{Gemma3-12B}}} & \textsf{en} & 0.763 & \textbf{0.788} & 0.471 & \textbf{0.377} & 0.736 & \textbf{\uwave{0.757}} & 0.731 & \textbf{0.657}\\
 & \textsf{de} & 0.703 & \textbf{0.784} & 0.472 & \textbf{0.352} & 0.678 & \textbf{0.694} & 0.686 & \textbf{\uwave{0.631}}\\
 & \textsf{zh} & \uwave{0.789} & \textbf{0.869} & \textbf{0.554} & 0.709 & 0.605 & \textbf{0.631} & \textbf{0.850} & 0.973\\
 & \textsf{ar} & \uwave{0.789} & \textbf{\uwave{0.889}} & 0.467 & \textbf{0.407} & \textbf{0.672} & 0.631 & 0.699 & \textbf{0.670}\\
 & \textsf{vi} & 0.709 & \textbf{0.737} & 0.508 & \textbf{0.401} & 0.650 & \textbf{0.676} & 0.707 & \textbf{0.655}\\
 & \textsf{sw} & 0.733 & \textbf{0.747} & \uwave{0.446} & \textbf{\uwave{0.358}} & 0.578 & \textbf{0.640} & 0.653 & \textbf{0.634}\\
 & \textsf{ru} & 0.691 & \textbf{0.717} & 0.494 & \textbf{0.378} & 0.643 & \textbf{0.685} & 0.743 & \textbf{0.693}\\

\bottomrule[1.5pt]
\end{tabular}%
\caption{HLFR and TS automatic evaluation results on \data{SIB200} and \data{TAXI1500}. \textbf{Boldface} denotes the stronger method per language and metric between \textbf{DG-CF} and \textbf{\sys{Macro}}, while the \uwave{wavy underline} highlights the best score among the seven languages for the same model, method, and metric.}
\label{tab:hlfr_ts_dgcf_macro_side_by_side}
\end{table*}


\begin{figure*}[t]
    \centering
    \begin{subfigure}[t]{0.95\linewidth}
        \centering
        \includegraphics[width=\linewidth]{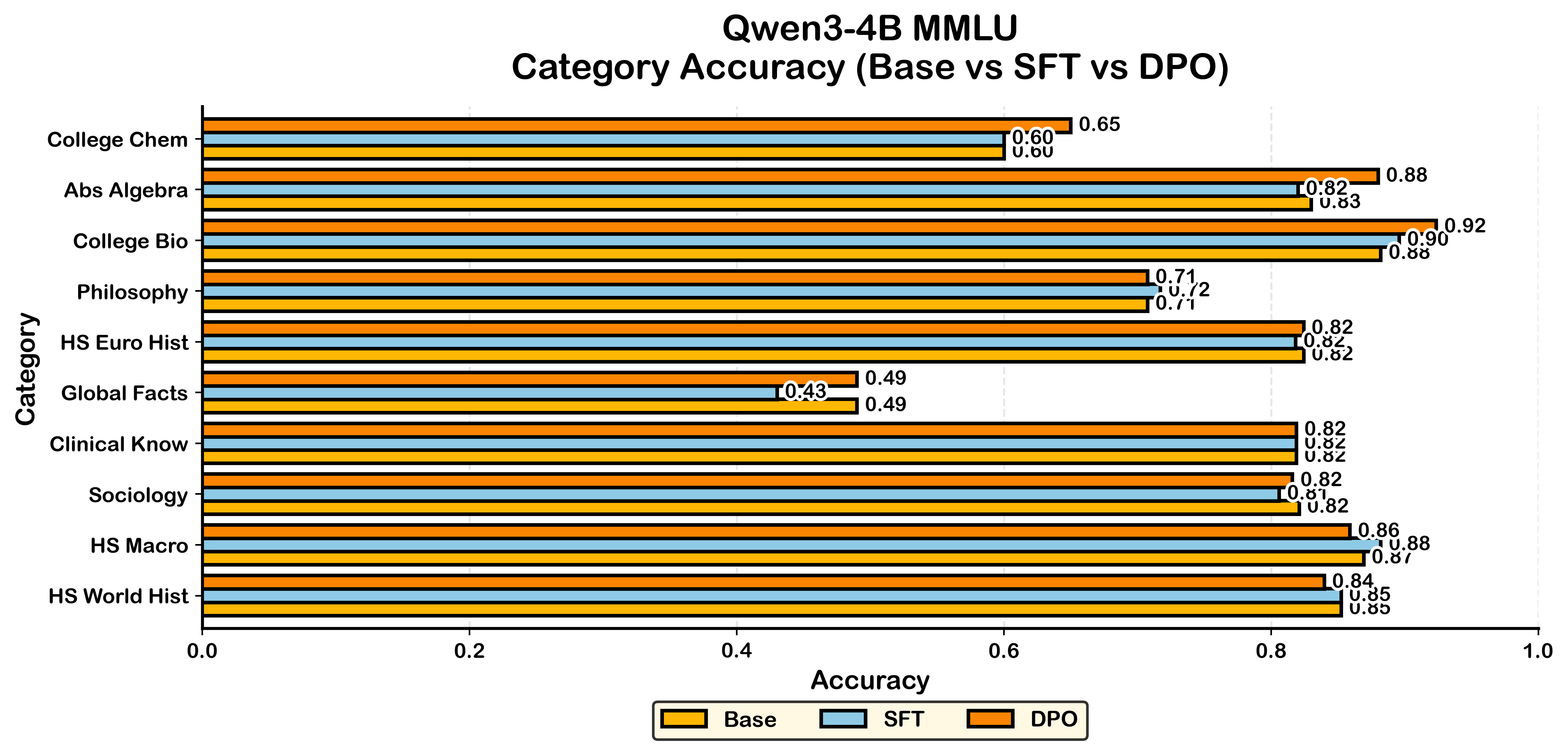}
        \caption{Category-wise performance on \data{MMLU} with \lm{Qwen3-4B}.}
        \label{fig:mmlu_qwen}
    \end{subfigure}

    \vspace{0.5em}

    \begin{subfigure}[t]{0.95\linewidth}
        \centering
        \includegraphics[width=\linewidth]{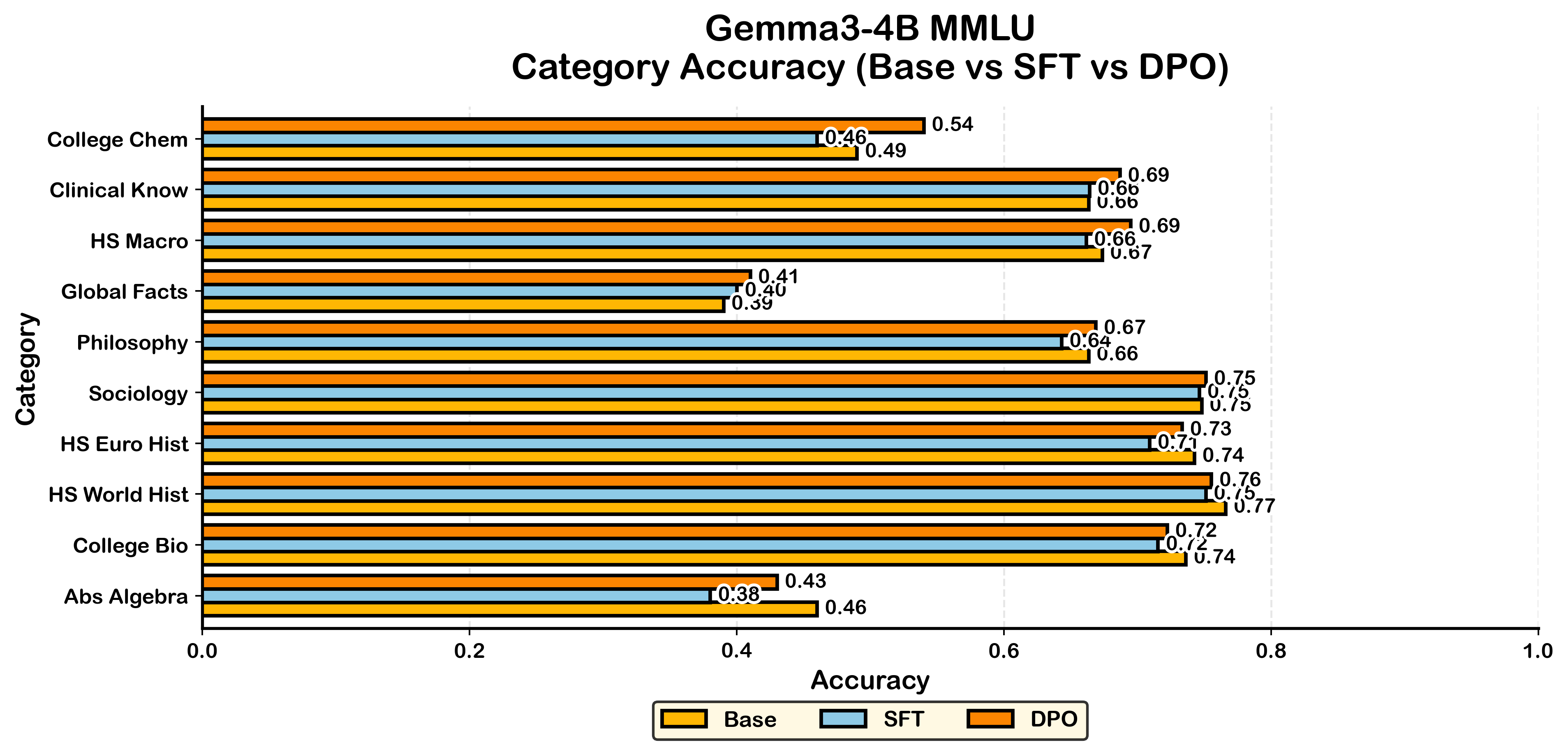}
        \caption{Category-wise performance on \data{MMLU} with \lm{Gemma3-4B}.}
        \label{fig:mmlu_gemma}
    \end{subfigure}

    \caption{Impact of \our on multilingual general capability measured on \data{MMLU}. }
    \label{fig:mmlu}
\end{figure*}


\begin{figure*}[t]
    \centering
    \begin{subfigure}[t]{0.95\textwidth}
        \centering
        \includegraphics[width=\linewidth]{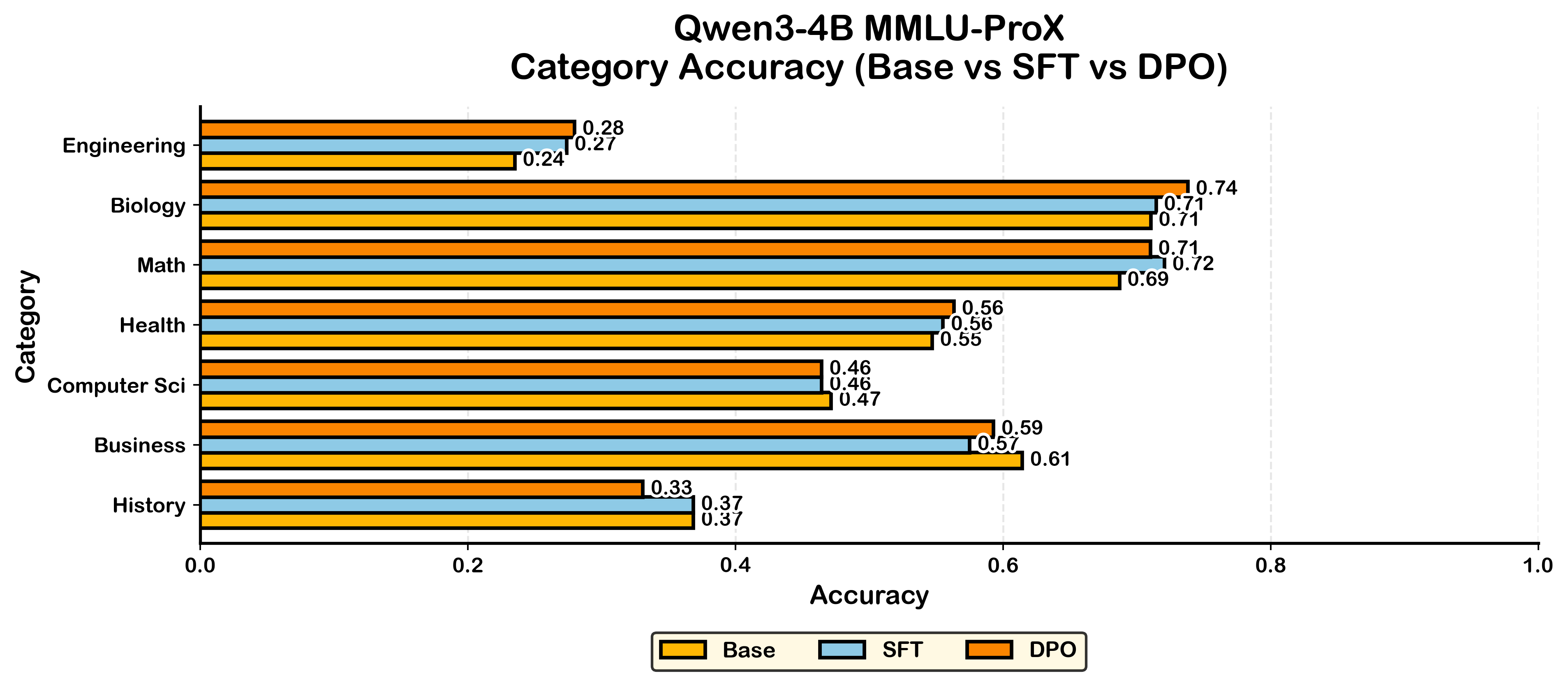}
        \caption{Category-wise performance aggregated over the languages used in post-training on \data{MMLU-ProX} with \lm{Qwen3-4B}.}
        \label{fig:mmlu_prox_category_qwen}
    \end{subfigure}

    \vspace{0.5em}

    \begin{subfigure}[t]{0.95\textwidth}
        \centering
        \includegraphics[width=\linewidth]{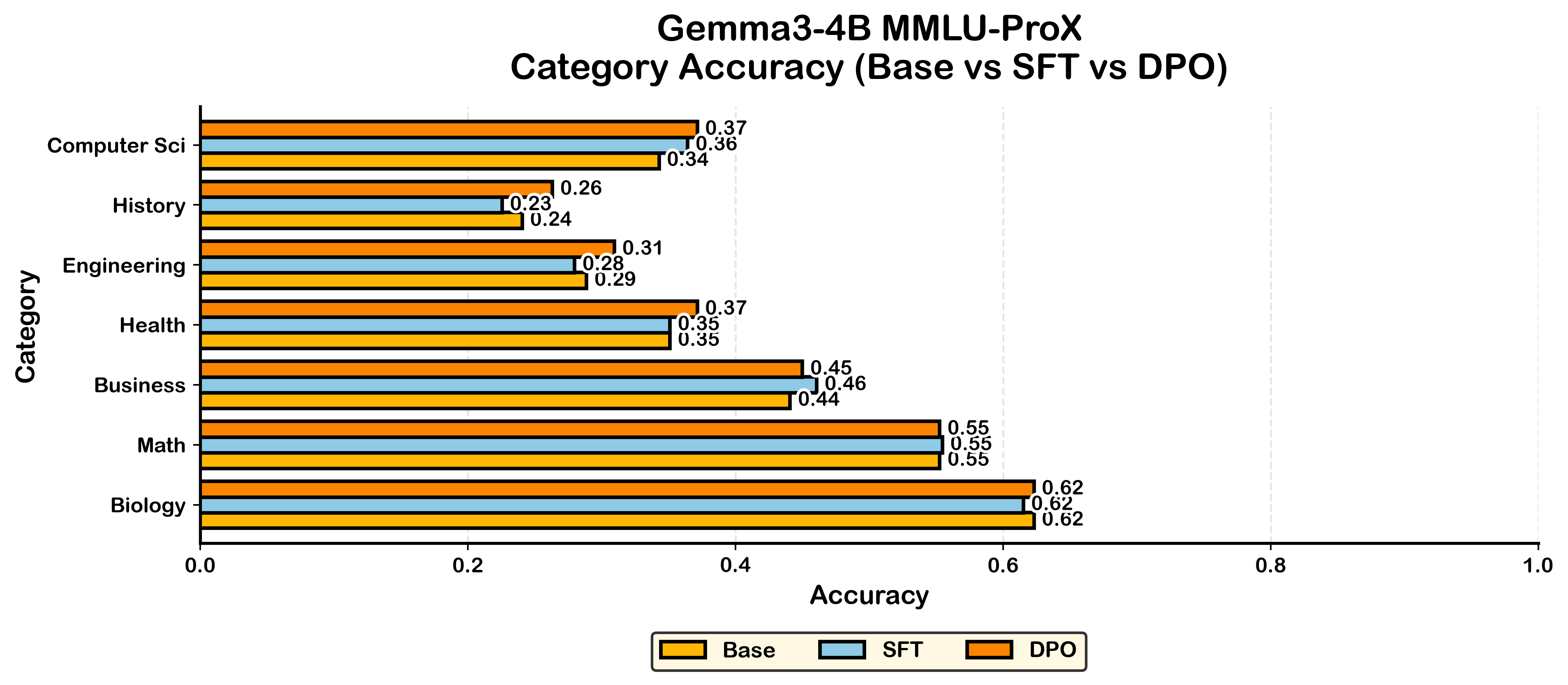}
        \caption{Category-wise performance aggregated over the languages used in post-training on \data{MMLU-ProX} with \lm{Gemma3-4B}.}
        \label{fig:mmlu_prox_category_gemma}
    \end{subfigure}

    \caption{Impact of \our on reasoning capability measured on \data{MMLU-ProX} from the category perspective . Subfigures~(a) and~(b) present the category-wise performance of \lm{Qwen3-4B} and \lm{Gemma3-4B}, respectively.}
    \label{fig:mmlu_prox_category}
\end{figure*}

\begin{figure*}[t]
    \centering
    \begin{subfigure}[t]{0.95\textwidth}
        \centering
        \includegraphics[width=\linewidth]{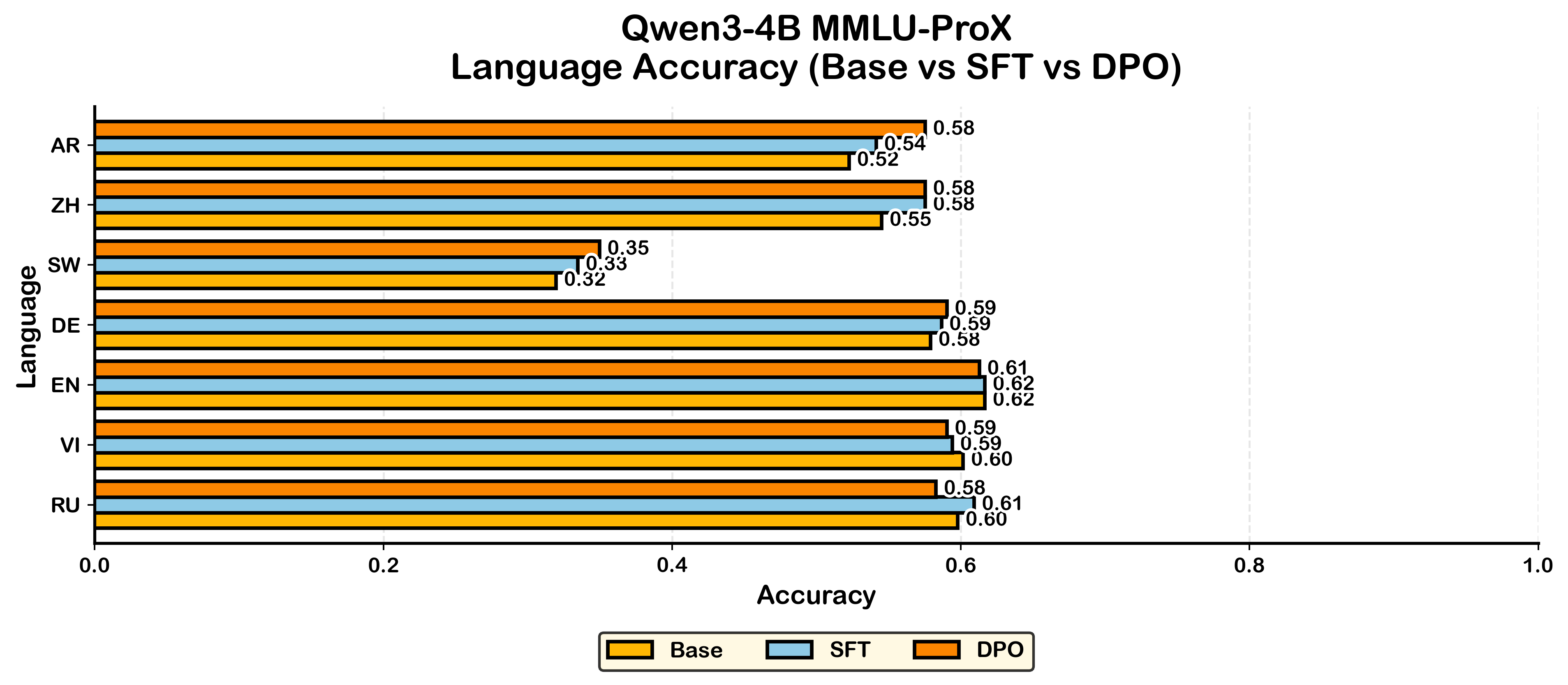}
        \caption{Language-wise performance on \data{MMLU-ProX} with \lm{Qwen3-4B}.}
        \label{fig:mmlu_prox_language_qwen}
    \end{subfigure}

    \vspace{0.5em}

    \begin{subfigure}[t]{0.95\textwidth}
        \centering
        \includegraphics[width=\linewidth]{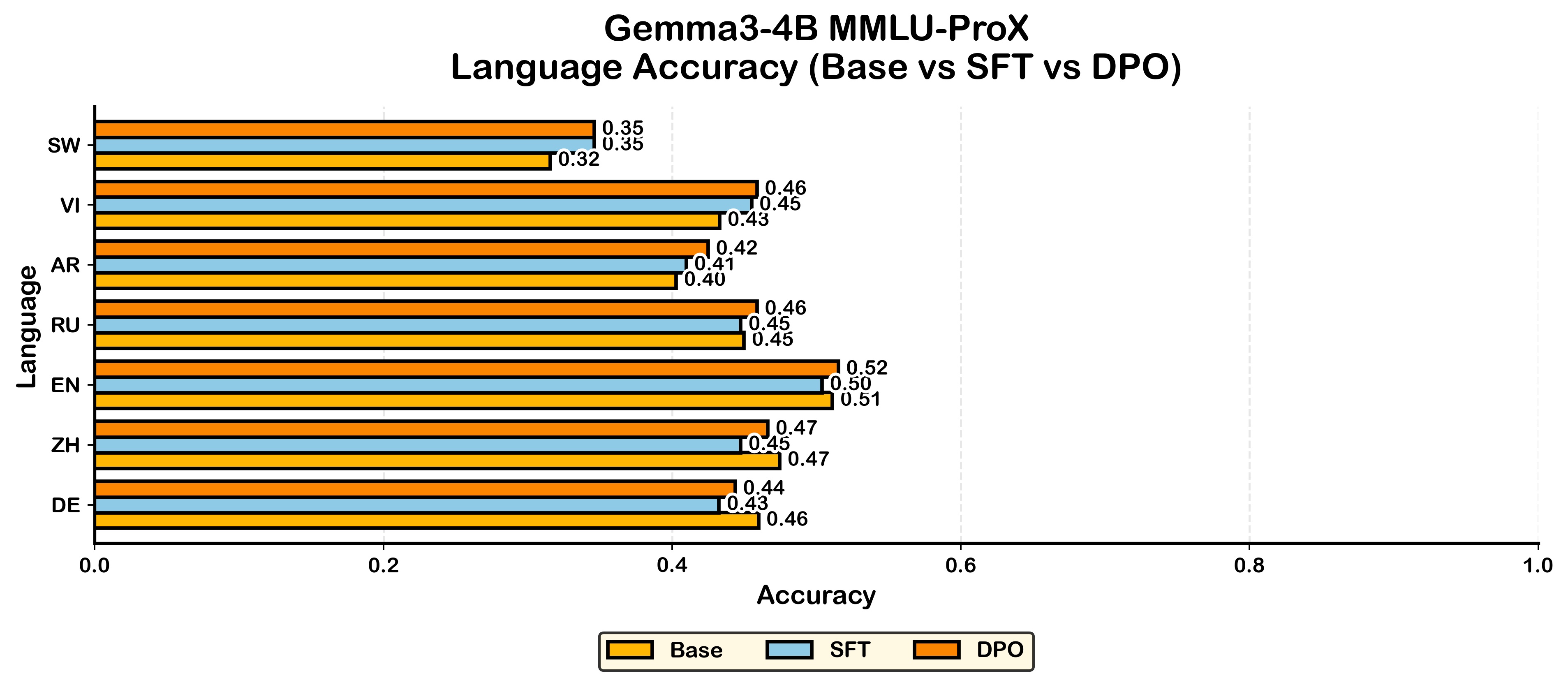}
        \caption{Language-wise performance on \data{MMLU-ProX} with \lm{Gemma3-4B}.}
        \label{fig:mmlu_prox_language_gemma}
    \end{subfigure}

    \caption{Impact of \our on cross-lingual generalization measured on \data{MMLU-ProX} from the language perspective. Subfigures~(a) and~(b) present the language-wise performance of \lm{Qwen3-4B} and \lm{Gemma3-4B}, respectively.}
    \label{fig:mmlu_prox_language}
\end{figure*}

\section{The Impact of Direct Preference Optimization}
\label{app:impact}
\subsection{Impact on Model General Capability}
We explore the impact of DPO training by \our on general model capabilities. We focus on two perspectives -- reasoning and multilingual capability -- for which we employ the \data{MMLU-ProX} \cite{xuan-etal-2025-mmlu} and \data{MMLU} \cite{hendrycks2021measuring} datasets. \data{MMLU} is a widely used benchmark for measuring multitask language understanding across 57 subjects. In contrast, \data{MMLU-ProX} extends the more challenging \data{MMLU-Pro} benchmark, which features reasoning-focused questions and ten answer choices instead of four.

Figures \ref{fig:mmlu}, \ref{fig:mmlu_prox_category} and \ref{fig:mmlu_prox_language} demonstrate that, in general, models after applying \our can largely maintain or even enhance their reasoning and multilingual capabilities. 
Table \ref{tab:general_capability} further confirms the same trend with overall performance: \our (DPO) consistently preserves or slightly improves performance relative to the base model on both \data{MMLU} and \data{MMLU-ProX}. In contrast, SFT does not show the same level of consistency, especially for \lm{Gemma3-4B} on \data{MMLU}.
This is crucial, as we aim to improve the quality of self-generated counterfactual explanations while keeping general capabilities intact.

\begin{table}[t]
    \centering
    \small
    \setlength{\tabcolsep}{4pt}
    \begin{tabular}{lccc ccc}
        \toprule
        & \multicolumn{3}{c}{\data{MMLU}} & \multicolumn{3}{c}{\data{MMLU-ProX}} \\
        \cmidrule(lr){2-4} \cmidrule(lr){5-7}
        Model & Base & SFT & DPO & Base & SFT & DPO \\
        \midrule
        \lm{Qwen3-4B}  & 79.38 & 79.33 & \textbf{79.78} & 54.03 & 55.10 & \textbf{55.37} \\
        \lm{Gemma3-4B} & 67.13 & 64.88 & \textbf{67.26} & 43.38 & 43.45 & \textbf{44.47} \\
        \bottomrule
    \end{tabular}
    \caption{Overall accuracy (\%) on \data{MMLU} and selected \data{MMLU-ProX} categories for the base, SFT, and DPO.}
    \label{tab:general_capability}
\end{table}

\subsection{Impact on Cross-lingual Generalizability}
To further assess whether the benefits of preference alignment transfer beyond the training languages, we evaluate \lm{Gemma3-4B} and \lm{Qwen3-4B} on six unseen languages: Bulgarian, Greek, Spanish, Hindi, Thai, and Turkish. This analysis complements our in-domain findings, where \our generally improves counterfactual quality and shows cross-lingual generalizability.

In general, Table \ref{tab:zero_shot_unseen_languages} suggests that \our generalizes beyond the languages included in the training set, but its gains are more consistent on \textit{validity} than on \textit{fluency} or \textit{minimality}. In other words, \our appears to learn a more transferable notion of \emph{what kinds of edits are effective} for flipping predictions, whereas \emph{how naturally and minimally those edits are performed} remains more sensitive to model family and dataset.


\section{Trade-off}
\label{app:trade-off}
\begin{figure*}
    \centering
    \includegraphics[width=\linewidth]{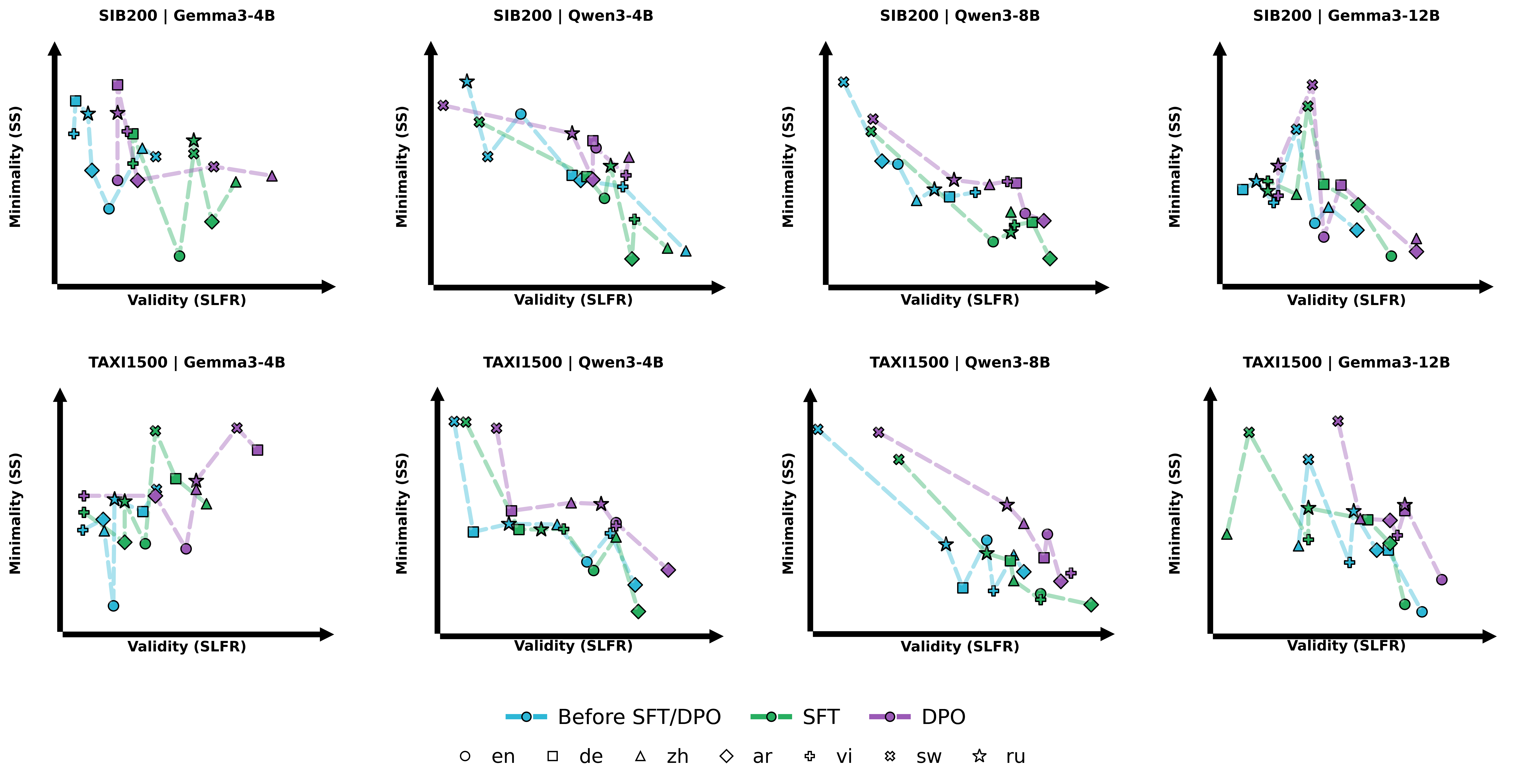}
    \caption{The validity-minimality trade-off across languages across  all models on \data{SIB200} and \data{Taxi1500}.}
    \label{fig:tradeoff_all}
\end{figure*}

Figure~\ref{fig:tradeoff_all} presents the \textit{validity-minimality} trade-off across all models and languages on \data{SIB200} and \data{Taxi1500}.

\section{Translation-based Counterfactual}
Table \ref{tab:dgcf_tbcf_with_deltas} illustrates the automatic evaluation results of translation-based counterfactuals generated by TB-CF. Although TB-CF yields higher \textit{validity} than those of DG-CF or \our, they require excessive modifications. These drastic edits, driven by translation artifacts and cross-linguistic syntactic differences, cause TB-CF to drift too far from the source text. Consequently, they fail to isolate the specific features that influence model decisions and should no longer be viewed as reliable counterfactual explanations.

\begin{table*}[t!]
\centering
\renewcommand*{\arraystretch}{1}
\resizebox{\textwidth}{!}{%
\begin{tabular}{cc|cc|cc|cc|cc|cc|cc}
\toprule[1.5pt]
\multicolumn{2}{c|}{\textbf{Dataset}} & \multicolumn{12}{c}{\textbf{\data{SIB200}}}\\
\cmidrule(l){1-14}
\multirow{2}{*}{\rotatebox[origin=c]{90}{\scriptsize{\textbf{Model}}}} & \textbf{Lang-} & \multicolumn{2}{c|}{\textbf{SLFR} ($\uparrow$)} & \multicolumn{2}{c|}{\textbf{HLFR} ($\uparrow$)} & \multicolumn{2}{c|}{\textbf{PPL} ($\downarrow$)} & \multicolumn{2}{c|}{\textbf{SS} ($\uparrow$)} & \multicolumn{2}{c|}{\textbf{TS} ($\downarrow$)} & \multicolumn{2}{c}{\textbf{$\mathcal{R}_{\text{edit}}$} ($\uparrow$)}\\
 & \textbf{uage} & \textbf{DG-CF} & \textbf{TB-CF} & \textbf{DG-CF} & \textbf{TB-CF} & \textbf{DG-CF} & \textbf{TB-CF} & \textbf{DG-CF} & \textbf{TB-CF} & \textbf{DG-CF} & \textbf{TB-CF} & \textbf{DG-CF} & \textbf{TB-CF}\\
\midrule
\centering \multirow{7}{*}{\rotatebox[origin=c]{90}{\lm{Gemma3-4B}}} & \textsf{en} & 0.627 & - & 0.559 & - & 48.76 & - & 0.613 & - & 0.529 & - & 0.471 & -\\
 & \textsf{de} & 0.592 & \textbf{\cell{0.737}{0.145}{+}} & \textbf{0.499} & \cell{0.414}{0.085}{-} & 28.54 & \textbf{\cell{11.95}{16.59}{-}} & \textbf{0.754} & \cell{0.267}{0.487}{-} & \textbf{0.389} & \cell{7.640}{7.251}{+} & \textbf{0.611} & \cell{-6.640}{7.251}{-}\\
 & \textsf{zh} & 0.662 & \textbf{\cell{0.687}{0.025}{+}} & \textbf{0.629} & \cell{0.313}{0.316}{-} & 35.09 & \textbf{\cell{15.68}{19.41}{-}} & \textbf{0.692} & \cell{0.074}{0.618}{-} & \textbf{0.467} & \cell{22.770}{22.303}{+} & \textbf{0.533} & \cell{-21.770}{22.303}{-}\\
 & \textsf{ar} & 0.609 & \textbf{\cell{0.657}{0.048}{+}} & \textbf{0.548} & \cell{0.293}{0.255}{-} & 27.74 & \textbf{\cell{19.29}{8.45}{-}} & \textbf{0.663} & \cell{0.073}{0.590}{-} & \textbf{0.401} & \cell{8.045}{7.644}{+} & \textbf{0.599} & \cell{-7.045}{7.644}{-}\\
 & \textsf{vi} & 0.590 & \textbf{\cell{0.677}{0.087}{+}} & \textbf{0.508} & \cell{0.434}{0.074}{-} & 18.88 & \textbf{\cell{10.80}{8.08}{-}} & \textbf{0.711} & \cell{0.317}{0.394}{-} & \textbf{0.456} & \cell{7.959}{7.503}{+} & \textbf{0.544} & \cell{-6.959}{7.503}{-}\\
 & \textsf{sw} & 0.676 & \textbf{\cell{0.687}{0.011}{+}} & \textbf{0.549} & \cell{0.424}{0.125}{-} & 18.71 & \textbf{\cell{13.59}{5.12}{-}} & \textbf{0.681} & \cell{0.427}{0.254}{-} & \textbf{0.446} & \cell{7.620}{7.174}{+} & \textbf{0.554} & \cell{-6.620}{7.174}{-}\\
 & \textsf{ru} & 0.605 & \textbf{\cell{0.677}{0.072}{+}} & \textbf{0.514} & \cell{0.404}{0.110}{-} & 20.26 & \textbf{\cell{12.04}{8.22}{-}} & \textbf{0.737} & \cell{0.288}{0.449}{-} & \textbf{0.459} & \cell{8.039}{7.580}{+} & \textbf{0.541} & \cell{-7.039}{7.580}{-}\\
\midrule
\centering \multirow{7}{*}{\rotatebox[origin=c]{90}{\lm{Qwen3-4B}}} & \textsf{en} & 0.531 & - & 0.482 & - & 58.52 & - & 0.801 & - & 0.224 & - & 0.776 & -\\
 & \textsf{de} & \textbf{0.709} & \cell{0.596}{0.113}{-} & \textbf{0.612} & \cell{0.455}{0.157}{-} & 45.15 & \textbf{\cell{41.40}{3.75}{-}} & 0.658 & \textbf{\cell{0.733}{0.075}{+}} & \textbf{0.306} & \cell{0.562}{0.256}{+} & \textbf{0.694} & \cell{0.438}{0.256}{-}\\
 & \textsf{zh} & 0.767 & \textbf{\cell{0.768}{0.001}{+}} & \textbf{0.724} & \cell{0.374}{0.350}{-} & \textbf{43.75} & \cell{44.39}{0.64}{+} & \textbf{0.598} & \cell{0.078}{0.520}{-} & \textbf{0.438} & \cell{1.211}{0.773}{+} & \textbf{0.562} & \cell{-0.211}{0.773}{-}\\
 & \textsf{ar} & 0.646 & \textbf{\cell{0.768}{0.122}{+}} & \textbf{0.591} & \cell{0.374}{0.217}{-} & 40.77 & \textbf{\cell{40.49}{0.28}{-}} & \textbf{0.704} & \cell{0.084}{0.620}{-} & \textbf{0.251} & \cell{0.941}{0.690}{+} & \textbf{0.749} & \cell{0.059}{0.690}{-}\\
 & \textsf{vi} & \textbf{0.678} & \cell{0.475}{0.203}{-} & \textbf{0.578} & \cell{0.394}{0.184}{-} & 26.15 & \textbf{\cell{20.38}{5.77}{-}} & 0.649 & \textbf{\cell{0.733}{0.084}{+}} & \textbf{0.320} & \cell{0.579}{0.259}{+} & \textbf{0.680} & \cell{0.421}{0.259}{-}\\
 & \textsf{sw} & 0.512 & \textbf{\cell{0.566}{0.054}{+}} & 0.137 & \textbf{\cell{0.141}{0.004}{+}} & 42.85 & \textbf{\cell{18.69}{24.16}{-}} & \textbf{0.714} & \cell{0.491}{0.223}{-} & \textbf{0.397} & \cell{1.889}{1.492}{+} & \textbf{0.603} & \cell{-0.889}{1.492}{-}\\
 & \textsf{ru} & 0.496 & \textbf{\cell{0.505}{0.009}{+}} & 0.423 & \textbf{\cell{0.444}{0.021}{+}} & \textbf{23.73} & \cell{32.23}{8.50}{+} & \textbf{0.793} & \cell{0.731}{0.062}{-} & \textbf{0.199} & \cell{0.612}{0.413}{+} & \textbf{0.801} & \cell{0.388}{0.413}{-}\\
\midrule
\centering \multirow{7}{*}{\rotatebox[origin=c]{90}{\lm{Qwen3-8B}}} & \textsf{en} & 0.594 & - & 0.489 & - & 53.24 & - & 0.731 & - & 0.231 & - & 0.769 & -\\
 & \textsf{de} & \textbf{0.652} & \cell{0.596}{0.056}{-} & \textbf{0.568} & \cell{0.495}{0.073}{-} & 32.09 & \textbf{\cell{30.57}{1.52}{-}} & \textbf{0.686} & \cell{0.661}{0.025}{-} & \textbf{0.263} & \cell{0.558}{0.295}{+} & \textbf{0.737} & \cell{0.442}{0.295}{-}\\
 & \textsf{zh} & 0.615 & \textbf{\cell{0.717}{0.102}{+}} & \textbf{0.534} & \cell{0.404}{0.130}{-} & 38.59 & \textbf{\cell{34.98}{3.61}{-}} & \textbf{0.681} & \cell{0.083}{0.598}{-} & \textbf{0.320} & \cell{1.235}{0.915}{+} & \textbf{0.680} & \cell{-0.235}{0.915}{-}\\
 & \textsf{ar} & 0.576 & \textbf{\cell{0.758}{0.182}{+}} & \textbf{0.505} & \cell{0.404}{0.101}{-} & \textbf{34.88} & \cell{34.94}{0.06}{+} & \textbf{0.735} & \cell{0.080}{0.655}{-} & \textbf{0.169} & \cell{0.951}{0.782}{+} & \textbf{0.831} & \cell{0.049}{0.782}{-}\\
 & \textsf{vi} & \textbf{0.681} & \cell{0.586}{0.095}{-} & \textbf{0.614} & \cell{0.515}{0.099}{-} & 22.66 & \textbf{\cell{19.40}{3.26}{-}} & \textbf{0.692} & \cell{0.656}{0.036}{-} & \textbf{0.308} & \cell{0.583}{0.275}{+} & \textbf{0.692} & \cell{0.417}{0.275}{-}\\
 & \textsf{sw} & 0.533 & \textbf{\cell{0.556}{0.023}{+}} & \textbf{0.244} & \cell{0.242}{0.002}{-} & 22.93 & \textbf{\cell{22.40}{0.53}{-}} & \textbf{0.844} & \cell{0.521}{0.323}{-} & \textbf{0.324} & \cell{1.512}{1.188}{+} & \textbf{0.676} & \cell{-0.512}{1.188}{-}\\
 & \textsf{ru} & \textbf{0.635} & \cell{0.596}{0.039}{-} & \textbf{0.572} & \cell{0.455}{0.117}{-} & \textbf{22.08} & \cell{22.43}{0.35}{+} & \textbf{0.696} & \cell{0.662}{0.034}{-} & \textbf{0.295} & \cell{0.609}{0.314}{+} & \textbf{0.705} & \cell{0.391}{0.314}{-}\\
\midrule
\centering \multirow{7}{*}{\rotatebox[origin=c]{90}{\lm{Gemma3-12B}}} & \textsf{en} & 0.830 & - & 0.763 & - & 53.90 & - & 0.530 & - & 0.471 & - & 0.529 & -\\
 & \textsf{de} & 0.767 & \textbf{\cell{0.788}{0.021}{+}} & \textbf{0.703} & \cell{0.667}{0.036}{-} & 28.13 & \textbf{\cell{13.32}{14.81}{-}} & \textbf{0.574} & \cell{0.239}{0.335}{-} & \textbf{0.472} & \cell{7.754}{7.282}{+} & \textbf{0.528} & \cell{-6.754}{7.282}{-}\\
 & \textsf{zh} & \textbf{0.842} & \cell{0.828}{0.014}{-} & \textbf{0.789} & \cell{0.576}{0.213}{-} & 31.61 & \textbf{\cell{16.96}{14.65}{-}} & \textbf{0.551} & \cell{0.073}{0.478}{-} & \textbf{0.554} & \cell{22.704}{22.150}{+} & \textbf{0.446} & \cell{-21.704}{22.150}{-}\\
 & \textsf{ar} & \textbf{0.867} & \cell{0.818}{0.049}{-} & \textbf{0.789} & \cell{0.556}{0.233}{-} & 25.48 & \textbf{\cell{25.19}{0.29}{-}} & \textbf{0.521} & \cell{0.063}{0.458}{-} & \textbf{0.467} & \cell{8.081}{7.614}{+} & \textbf{0.533} & \cell{-7.081}{7.614}{-}\\
 & \textsf{vi} & \textbf{0.794} & \cell{0.768}{0.026}{-} & \textbf{0.709} & \cell{0.657}{0.052}{-} & 18.54 & \textbf{\cell{12.51}{6.03}{-}} & \textbf{0.557} & \cell{0.255}{0.302}{-} & \textbf{0.508} & \cell{8.144}{7.636}{+} & \textbf{0.492} & \cell{-7.144}{7.636}{-}\\
 & \textsf{sw} & 0.814 & \textbf{\cell{0.828}{0.014}{+}} & \textbf{0.733} & \cell{0.697}{0.036}{-} & 14.86 & \textbf{\cell{12.45}{2.41}{-}} & \textbf{0.653} & \cell{0.399}{0.254}{-} & \textbf{0.446} & \cell{7.464}{7.018}{+} & \textbf{0.554} & \cell{-6.464}{7.018}{-}\\
 & \textsf{ru} & \textbf{0.779} & \cell{0.758}{0.021}{-} & \textbf{0.691} & \cell{0.626}{0.065}{-} & 21.03 & \textbf{\cell{15.23}{5.80}{-}} & \textbf{0.585} & \cell{0.240}{0.345}{-} & \textbf{0.494} & \cell{7.769}{7.275}{+} & \textbf{0.506} & \cell{-6.769}{7.275}{-}\\
\midrule[1.2pt]
\multicolumn{2}{c|}{\textbf{Dataset}} & \multicolumn{12}{c}{\textbf{\data{TAXI1500}}}\\
\cmidrule(l){1-14}
\multirow{2}{*}{\rotatebox[origin=c]{90}{\scriptsize{\textbf{Model}}}} & \textbf{Lang-} & \multicolumn{2}{c|}{\textbf{SLFR} ($\uparrow$)} & \multicolumn{2}{c|}{\textbf{HLFR} ($\uparrow$)} & \multicolumn{2}{c|}{\textbf{PPL} ($\downarrow$)} & \multicolumn{2}{c|}{\textbf{SS} ($\uparrow$)} & \multicolumn{2}{c|}{\textbf{TS} ($\downarrow$)} & \multicolumn{2}{c}{\textbf{$\mathcal{R}_{\text{edit}}$} ($\uparrow$)}\\
 & \textbf{uage} & \textbf{DG-CF} & \textbf{TB-CF} & \textbf{DG-CF} & \textbf{TB-CF} & \textbf{DG-CF} & \textbf{TB-CF} & \textbf{DG-CF} & \textbf{TB-CF} & \textbf{DG-CF} & \textbf{TB-CF} & \textbf{DG-CF} & \textbf{TB-CF}\\
\midrule
\centering \multirow{7}{*}{\rotatebox[origin=c]{90}{\lm{Gemma3-4B}}} & \textsf{en} & 0.531 & - & 0.434 & - & 36.24 & - & 0.507 & - & 0.670 & - & 0.330 & -\\
 & \textsf{de} & 0.557 & \textbf{\cell{0.874}{0.317}{+}} & \textbf{0.405} & \cell{0.324}{0.081}{-} & 33.93 & \textbf{\cell{11.59}{22.34}{-}} & \textbf{0.639} & \cell{0.293}{0.346}{-} & \textbf{0.580} & \cell{8.018}{7.438}{+} & \textbf{0.420} & \cell{-7.018}{7.438}{-}\\
 & \textsf{zh} & 0.523 & \textbf{\cell{0.892}{0.369}{+}} & \textbf{0.405} & \cell{0.315}{0.090}{-} & 39.48 & \textbf{\cell{14.64}{24.84}{-}} & \textbf{0.612} & \cell{0.291}{0.421}{-} & \textbf{0.807} & \cell{26.717}{25.910}{+} & \textbf{0.193} & \cell{-25.717}{25.910}{-}\\
 & \textsf{ar} & 0.522 & \textbf{\cell{0.919}{0.397}{+}} & \textbf{0.342} & \cell{0.333}{0.009}{-} & 41.88 & \textbf{\cell{17.79}{24.09}{-}} & \textbf{0.628} & \cell{0.327}{0.301}{-} & \textbf{0.716} & \cell{11.583}{10.867}{+} & \textbf{0.284} & \cell{-10.583}{10.867}{-}\\
 & \textsf{vi} & 0.504 & \textbf{\cell{0.838}{0.334}{+}} & \textbf{0.377} & \cell{0.270}{0.107}{-} & 18.47 & \textbf{\cell{10.46}{8.01}{-}} & \textbf{0.613} & \cell{0.295}{0.318}{-} & \textbf{0.631} & \cell{7.942}{7.311}{+} & \textbf{0.369} & \cell{-6.942}{7.311}{-}\\
 & \textsf{sw} & 0.569 & \textbf{\cell{0.865}{0.296}{+}} & \textbf{0.357} & \cell{0.351}{0.006}{-} & 25.59 & \textbf{\cell{14.78}{10.81}{-}} & \textbf{0.670} & \cell{0.503}{0.167}{-} & \textbf{0.647} & \cell{7.537}{6.890}{+} & \textbf{0.353} & \cell{-6.537}{6.890}{-}\\
 & \textsf{ru} & 0.532 & \textbf{\cell{0.865}{0.333}{+}} & \textbf{0.394} & \cell{0.288}{0.106}{-} & 21.49 & \textbf{\cell{12.73}{8.76}{-}} & \textbf{0.656} & \cell{0.317}{0.339}{-} & \textbf{0.650} & \cell{9.932}{9.282}{+} & \textbf{0.350} & \cell{-8.932}{9.282}{-}\\
\midrule
\centering \multirow{7}{*}{\rotatebox[origin=c]{90}{\lm{Qwen3-4B}}} & \textsf{en} & 0.623 & - & 0.485 & - & 32.64 & - & 0.614 & - & 0.480 & - & 0.520 & -\\
 & \textsf{de} & \textbf{0.486} & \textbf{\cell{0.486}{0.000}{0}} & 0.284 & \textbf{\cell{0.342}{0.058}{+}} & 40.17 & \textbf{\cell{37.80}{2.37}{-}} & \textbf{0.662} & \cell{0.589}{0.073}{-} & \textbf{0.444} & \cell{0.731}{0.287}{+} & \textbf{0.556} & \cell{0.269}{0.287}{-}\\
 & \textsf{zh} & 0.587 & \textbf{\cell{0.622}{0.035}{+}} & \textbf{0.309} & \cell{0.306}{0.003}{-} & 45.84 & \textbf{\cell{35.72}{10.12}{-}} & \textbf{0.674} & \cell{0.600}{0.074}{-} & \textbf{0.589} & \cell{0.927}{0.338}{+} & \textbf{0.411} & \cell{0.073}{0.338}{-}\\
 & \textsf{ar} & \textbf{0.681} & \cell{0.649}{0.032}{-} & \textbf{0.342} & \cell{0.279}{0.063}{-} & 86.15 & \textbf{\cell{55.98}{30.17}{-}} & \textbf{0.577} & \cell{0.563}{0.014}{-} & \textbf{0.615} & \cell{0.839}{0.224}{+} & \textbf{0.385} & \cell{0.161}{0.224}{-}\\
 & \textsf{vi} & 0.651 & \textbf{\cell{0.676}{0.025}{+}} & \textbf{0.356} & \cell{0.288}{0.068}{-} & 25.67 & \textbf{\cell{19.18}{6.49}{-}} & \textbf{0.660} & \cell{0.565}{0.095}{-} & \textbf{0.485} & \cell{0.814}{0.329}{+} & \textbf{0.515} & \cell{0.186}{0.329}{-}\\
 & \textsf{sw} & 0.463 & \textbf{\cell{0.505}{0.042}{+}} & 0.107 & \textbf{\cell{0.117}{0.010}{+}} & 25.61 & \textbf{\cell{14.90}{10.71}{-}} & \textbf{0.840} & \cell{0.553}{0.287}{-} & \textbf{0.312} & \cell{1.765}{1.453}{+} & \textbf{0.688} & \cell{-0.765}{1.453}{-}\\
 & \textsf{ru} & 0.529 & \textbf{\cell{0.631}{0.102}{+}} & 0.315 & \textbf{\cell{0.324}{0.009}{+}} & 27.73 & \textbf{\cell{25.04}{2.69}{-}} & \textbf{0.675} & \cell{0.620}{0.055}{-} & \textbf{0.526} & \cell{0.848}{0.322}{+} & \textbf{0.474} & \cell{0.152}{0.322}{-}\\
\midrule
\centering \multirow{7}{*}{\rotatebox[origin=c]{90}{\lm{Qwen3-8B}}} & \textsf{en} & 0.559 & - & 0.387 & - & 30.95 & - & 0.682 & - & 0.474 & - & 0.526 & -\\
 & \textsf{de} & \textbf{0.495} & \cell{0.432}{0.063}{-} & \textbf{0.342} & \cell{0.252}{0.090}{-} & \textbf{25.85} & \cell{26.46}{0.61}{+} & 0.617 & \textbf{\cell{0.646}{0.029}{+}} & \textbf{0.562} & \cell{0.727}{0.165}{+} & \textbf{0.438} & \cell{0.273}{0.165}{-}\\
 & \textsf{zh} & \textbf{0.631} & \cell{0.577}{0.054}{-} & \textbf{0.360} & \cell{0.324}{0.036}{-} & 38.92 & \textbf{\cell{33.37}{5.55}{-}} & \textbf{0.662} & \cell{0.639}{0.023}{-} & \textbf{0.689} & \cell{0.933}{0.244}{+} & \textbf{0.311} & \cell{0.067}{0.244}{-}\\
 & \textsf{ar} & 0.658 & \textbf{\cell{0.667}{0.009}{+}} & \textbf{0.387} & \cell{0.333}{0.054}{-} & 50.29 & \textbf{\cell{41.30}{8.99}{-}} & \textbf{0.639} & \cell{0.600}{0.039}{-} & \textbf{0.622} & \cell{0.808}{0.186}{+} & \textbf{0.378} & \cell{0.192}{0.186}{-}\\
 & \textsf{vi} & \textbf{0.577} & \cell{0.514}{0.063}{-} & \textbf{0.405} & \cell{0.306}{0.099}{-} & 18.83 & \textbf{\cell{16.36}{2.47}{-}} & \textbf{0.613} & \cell{0.603}{0.010}{-} & \textbf{0.621} & \cell{0.849}{0.228}{+} & \textbf{0.379} & \cell{0.151}{0.228}{-}\\
 & \textsf{sw} & 0.108 & \textbf{\cell{0.180}{0.072}{+}} & 0.009 & \textbf{\cell{0.054}{0.045}{+}} & 18.96 & \textbf{\cell{18.90}{0.06}{-}} & \textbf{0.833} & \cell{0.610}{0.223}{-} & \textbf{0.693} & \cell{1.241}{0.548}{+} & \textbf{0.307} & \cell{-0.241}{0.548}{-}\\
 & \textsf{ru} & \textbf{0.450} & \cell{0.378}{0.072}{-} & 0.270 & \textbf{\cell{0.288}{0.018}{+}} & 18.73 & \textbf{\cell{18.33}{0.40}{-}} & \textbf{0.676} & \cell{0.671}{0.005}{-} & \textbf{0.566} & \cell{0.847}{0.281}{+} & \textbf{0.434} & \cell{0.153}{0.281}{-}\\
\midrule
\centering \multirow{7}{*}{\rotatebox[origin=c]{90}{\lm{Gemma3-12B}}} & \textsf{en} & 0.832 & - & 0.736 & - & 41.48 & - & 0.437 & - & 0.731 & - & 0.269 & -\\
 & \textsf{de} & 0.791 & \textbf{\cell{0.937}{0.146}{+}} & \textbf{0.678} & \cell{0.667}{0.011}{-} & 26.50 & \textbf{\cell{13.01}{13.49}{-}} & \textbf{0.537} & \cell{0.257}{0.280}{-} & \textbf{0.686} & \cell{8.387}{7.701}{+} & \textbf{0.314} & \cell{-7.387}{7.701}{-}\\
 & \textsf{zh} & 0.682 & \textbf{\cell{0.838}{0.156}{+}} & 0.605 & \textbf{\cell{0.640}{0.035}{+}} & 27.90 & \textbf{\cell{17.40}{10.50}{-}} & \textbf{0.544} & \cell{0.257}{0.287}{-} & \textbf{0.850} & \cell{26.798}{25.948}{+} & \textbf{0.150} & \cell{-25.798}{25.948}{-}\\
 & \textsf{ar} & 0.777 & \textbf{\cell{0.802}{0.025}{+}} & \textbf{0.672} & \cell{0.631}{0.041}{-} & 37.25 & \textbf{\cell{22.22}{15.03}{-}} & \textbf{0.537} & \cell{0.298}{0.239}{-} & \textbf{0.699} & \cell{11.946}{11.247}{+} & \textbf{0.301} & \cell{-10.946}{11.247}{-}\\
 & \textsf{vi} & 0.744 & \textbf{\cell{0.892}{0.148}{+}} & \textbf{0.650} & \cell{0.559}{0.091}{-} & 16.18 & \textbf{\cell{12.08}{4.10}{-}} & \textbf{0.517} & \cell{0.247}{0.270}{-} & \textbf{0.707} & \cell{8.459}{7.752}{+} & \textbf{0.293} & \cell{-7.459}{7.752}{-}\\
 & \textsf{sw} & 0.694 & \textbf{\cell{0.919}{0.225}{+}} & 0.578 & \textbf{\cell{0.766}{0.188}{+}} & 17.46 & \textbf{\cell{13.54}{3.92}{-}} & \textbf{0.684} & \cell{0.476}{0.208}{-} & \textbf{0.653} & \cell{7.483}{6.830}{+} & \textbf{0.347} & \cell{-6.483}{6.830}{-}\\
 & \textsf{ru} & 0.749 & \textbf{\cell{0.883}{0.134}{+}} & \textbf{0.643} & \cell{0.604}{0.039}{-} & 20.49 & \textbf{\cell{15.75}{4.74}{-}} & \textbf{0.600} & \cell{0.271}{0.329}{-} & \textbf{0.743} & \cell{9.984}{9.241}{+} & \textbf{0.257} & \cell{-8.984}{9.241}{-}\\
\bottomrule[1.5pt]
\end{tabular}%
}
\caption{Automatic evaluation on \data{SIB200} and \data{TAXI1500}. We report \textbf{DG-CF} and \textbf{TB-CF} on SLFR, HLFR, PPL, SS, TS, and $\mathcal{R}_{\text{edit}}$. \textbf{Boldface} denotes the better method for each language and metric.}
\label{tab:dgcf_tbcf_with_deltas}
\end{table*}

\section{Cross-lingual Edit Similarity}
\begin{figure*}[t]
    \centering
    \begin{subfigure}[b]{\textwidth}
    \centering
    \includegraphics[width=\textwidth]{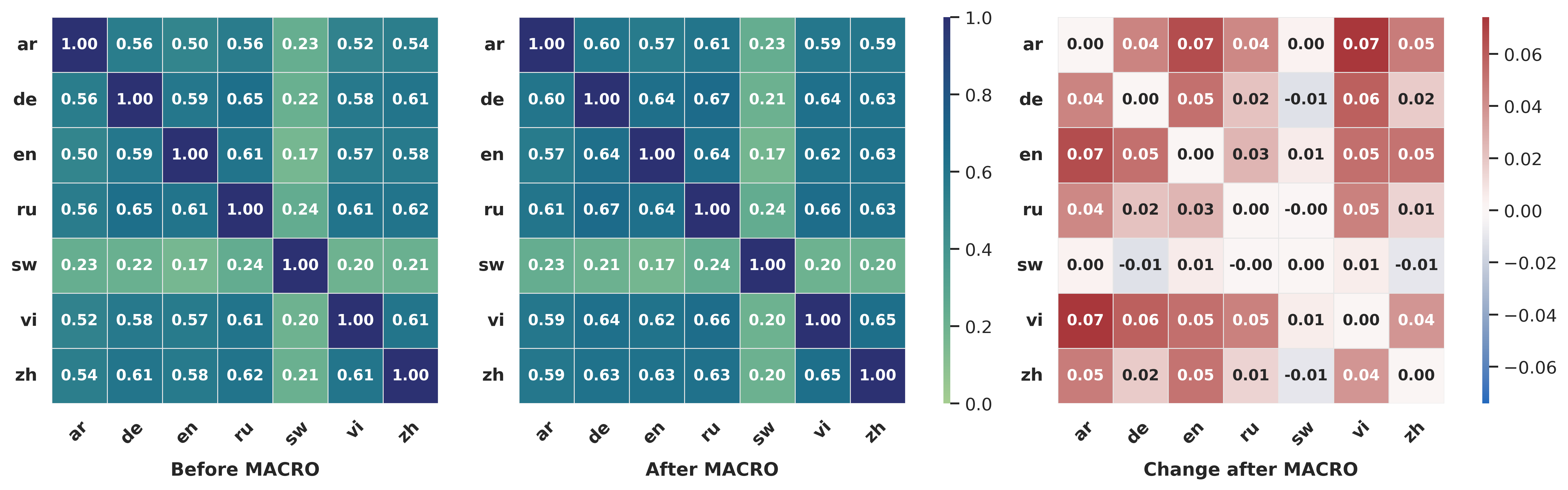}
    \subcaption{Cross-lingual Edit Similarity on \data{TAXI1500}, computed from counterfactuals in different languages.}
    \label{fig:ces_compare_dgcf+macro+diff_taxi1500_test}
    \end{subfigure}
    \begin{subfigure}[b]{\textwidth}
    \centering
    \includegraphics[width=\textwidth]{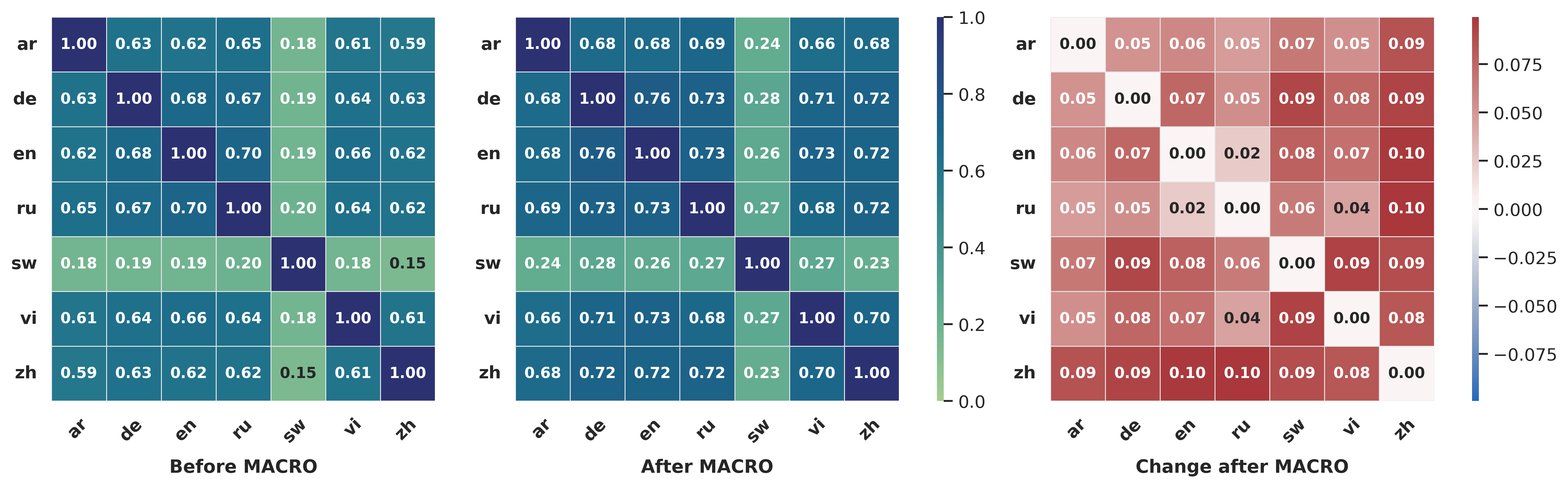}
    \caption{Cross-lingual Edit Similarity on \data{SIB200}, computed from counterfactuals in different languages.}
    \label{fig:ces_compare_dgcf+macro+diff_sib200_test}
    \end{subfigure}
    \caption{Cross-lingual edit similarity scores.}
    \label{fig:ces_score}
\end{figure*}

Figure \ref{fig:ces_compare_dgcf+macro+diff_taxi1500_test} shows the cross-lingual edit similarity scores across all language pairs on \data{Taxi1500}. The trend of \data{Taxi1500} is consistent with that of \data{SIB200} (Figure~\ref{fig:ces_compare_dgcf+macro+diff_sib200_test}, \S\ref{subsec:ces}). The cross-lingual edit consistency between Swahili and other languages is noticeably lower than that of other language pairs. Generally, after applying \our, the cross-lingual edit similarity increases.

\section{Ablation Study}
\subsection{Ablation Examples}
\label{app:ablation_examples}
We present qualitative multilingual counterfactual examples for \data{SIB200} and \data{TAXI1500} in Figures~\ref{fig:sib200-macro-ablation-example} and \ref{fig:taxi1500-macro-ablation-example}, respectively.

\newcommand{\wos}{w/o}

\begin{table*}[t!]
  \centering
  \footnotesize
  \setlength{\tabcolsep}{5.5pt}
  \renewcommand{\arraystretch}{1.08}

  \resizebox{2\columnwidth}{!}{%
  \begin{tabular}{@{}r l ccccc ccccc@{}}
    \toprule
    & & \multicolumn{5}{c}{\textbf{\data{SIB200}}}
      & \multicolumn{5}{c}{\textbf{\data{Taxi1500}}} \\
    \cmidrule(lr){3-7}\cmidrule(lr){8-12}
    \textbf{Setting} & \textbf{Lang}
      & \textbf{SLFR} \(\uparrow\) & \textbf{HLFR} \(\uparrow\) & \textbf{PPL} \(\downarrow\) & \textbf{SS} \(\uparrow\) & \textbf{TS} \(\downarrow\)
      & \textbf{SLFR} \(\uparrow\) & \textbf{HLFR} \(\uparrow\) & \textbf{PPL} \(\downarrow\) & \textbf{SS} \(\uparrow\) & \textbf{TS} \(\downarrow\)  \\
    \midrule

    \multirow{7}{*}{\textbf{\wos\ \Rflip}} & \textsf{en} & \cell{0.525}{0.212}{-} & \cell{0.384}{0.172}{-} & \cell{35.95}{1.92}{-} & \cell{0.705}{0.042}{+} & \cell{0.261}{0.047}{-} & \cell{0.685}{0.036}{-} & \cell{0.468}{0.064}{-} & \cell{24.65}{0.40}{-} & \cell{0.707}{0.017}{+} & \cell{0.452}{0.016}{-} \\
                             & \textsf{de} & \cell{0.586}{0.141}{-} & \cell{0.485}{0.101}{-} & \cell{27.86}{0.10}{+} & \cell{0.698}{0.007}{-} & \cell{0.260}{0.001}{+} & \cell{0.712}{0.000}{0} & \cell{0.414}{0.091}{-} & \cell{24.70}{1.83}{+} & \cell{0.706}{0.048}{+} & \cell{0.493}{0.051}{-} \\
                             & \textsf{zh} & \cell{0.576}{0.121}{-} & \cell{0.485}{0.141}{-} & \cell{35.48}{0.56}{+} & \cell{0.694}{0.009}{-} & \cell{0.558}{0.008}{-} & \cell{0.640}{0.018}{-} & \cell{0.405}{0.027}{+} & \cell{35.83}{2.20}{+} & \cell{0.752}{0.047}{+} & \cell{0.819}{0.046}{-} \\
                             & \textsf{ar} & \cell{0.657}{0.101}{-} & \cell{0.495}{0.151}{-} & \cell{30.31}{0.21}{-} & \cell{0.668}{0.015}{+} & \cell{0.270}{0.011}{-} & \cell{0.757}{0.000}{0} & \cell{0.459}{0.018}{+} & \cell{47.18}{9.38}{+} & \cell{0.658}{0.032}{+} & \cell{0.566}{0.072}{-} \\
                             & \textsf{vi} & \cell{0.657}{0.060}{-} & \cell{0.485}{0.071}{-} & \cell{23.59}{0.48}{+} & \cell{0.707}{0.000}{0} & \cell{0.284}{0.020}{-} & \cell{0.748}{0.036}{-} & \cell{0.495}{0.019}{-} & \cell{19.00}{1.02}{+} & \cell{0.675}{0.038}{+} & \cell{0.562}{0.039}{-} \\
                             & \textsf{sw} & \cell{0.515}{0.051}{-} & \cell{0.212}{0.051}{-} & \cell{20.22}{0.49}{+} & \cell{0.794}{0.001}{+} & \cell{0.217}{0.538}{-} & \cell{0.279}{0.009}{+} & \cell{0.081}{0.000}{0} & \cell{17.10}{0.32}{+} & \cell{0.841}{0.012}{+} & \cell{0.347}{0.903}{-} \\
                             & \textsf{ru} & \cell{0.697}{0.040}{+} & \cell{0.566}{0.061}{+} & \cell{21.77}{0.24}{+} & \cell{0.705}{0.004}{-} & \cell{0.286}{0.003}{+} & \cell{0.550}{0.063}{-} & \cell{0.351}{0.045}{-} & \cell{20.91}{0.68}{+} & \cell{0.758}{0.028}{+} & \cell{0.541}{0.050}{-} \\
        \midrule

    \multirow{7}{*}{\textbf{\wos\ \Raug}} & \textsf{en} & \cell{0.626}{0.111}{-} & \cell{0.455}{0.101}{-} & \cell{35.03}{2.84}{-} & \cell{0.702}{0.039}{+} & \cell{0.263}{0.045}{-} & \cell{0.676}{0.045}{-} & \cell{0.477}{0.055}{-} & \cell{25.13}{0.08}{+} & \cell{0.730}{0.040}{+} & \cell{0.435}{0.033}{-} \\
                            & \textsf{de} & \cell{0.636}{0.091}{-} & \cell{0.545}{0.041}{-} & \cell{27.17}{0.59}{-} & \cell{0.691}{0.014}{-} & \cell{0.266}{0.007}{+} & \cell{0.739}{0.027}{+} & \cell{0.360}{0.145}{-} & \cell{23.92}{1.05}{+} & \cell{0.711}{0.053}{+} & \cell{0.475}{0.069}{-} \\
                            & \textsf{zh} & \cell{0.646}{0.051}{-} & \cell{0.576}{0.050}{-} & \cell{34.73}{0.19}{-} & \cell{0.670}{0.033}{-} & \cell{0.586}{0.020}{+} & \cell{0.595}{0.063}{-} & \cell{0.315}{0.063}{-} & \cell{36.91}{3.28}{+} & \cell{0.747}{0.042}{+} & \cell{0.822}{0.043}{-} \\
                            & \textsf{ar} & \cell{0.677}{0.081}{-} & \cell{0.566}{0.080}{-} & \cell{29.07}{1.45}{-} & \cell{0.672}{0.019}{+} & \cell{0.285}{0.004}{+} & \cell{0.703}{0.054}{-} & \cell{0.423}{0.018}{-} & \cell{42.12}{4.32}{+} & \cell{0.674}{0.048}{+} & \cell{0.572}{0.066}{-} \\
                            & \textsf{vi} & \cell{0.657}{0.060}{-} & \cell{0.576}{0.020}{+} & \cell{23.33}{0.22}{+} & \cell{0.682}{0.025}{-} & \cell{0.302}{0.002}{-} & \cell{0.721}{0.063}{-} & \cell{0.432}{0.082}{-} & \cell{18.90}{0.92}{+} & \cell{0.653}{0.016}{+} & \cell{0.576}{0.025}{-} \\
                            & \textsf{sw} & \cell{0.525}{0.041}{-} & \cell{0.202}{0.061}{-} & \cell{19.30}{0.43}{-} & \cell{0.783}{0.010}{-} & \cell{0.214}{0.541}{-} & \cell{0.297}{0.027}{+} & \cell{0.072}{0.009}{-} & \cell{17.01}{0.23}{+} & \cell{0.859}{0.030}{+} & \cell{0.336}{0.914}{-} \\
                            & \textsf{ru} & \cell{0.646}{0.011}{-} & \cell{0.525}{0.020}{+} & \cell{21.07}{0.46}{-} & \cell{0.703}{0.006}{-} & \cell{0.287}{0.004}{+} & \cell{0.631}{0.018}{+} & \cell{0.369}{0.027}{-} & \cell{22.08}{1.85}{+} & \cell{0.743}{0.013}{+} & \cell{0.532}{0.059}{-} \\
     
    \midrule

    \multirow{7}{*}{\textbf{\wos\ \Redit}} & \textsf{en} & \cell{0.929}{0.192}{+} & \cell{0.818}{0.262}{+} & \cell{28.91}{8.96}{-} & \cell{0.348}{0.315}{-} & \cell{0.665}{0.357}{+} & \cell{0.847}{0.126}{+} & \cell{0.667}{0.135}{+} & \cell{17.99}{7.06}{-} & \cell{0.431}{0.259}{-} & \cell{0.748}{0.280}{+} \\
                             & \textsf{de} & \cell{0.929}{0.202}{+} & \cell{0.828}{0.242}{+} & \cell{21.06}{6.70}{-} & \cell{0.395}{0.310}{-} & \cell{0.626}{0.367}{+} & \cell{0.856}{0.144}{+} & \cell{0.712}{0.207}{+} & \cell{17.14}{5.73}{-} & \cell{0.431}{0.227}{-} & \cell{0.781}{0.237}{+} \\
                             & \textsf{zh} & \cell{0.939}{0.242}{+} & \cell{0.909}{0.283}{+} & \cell{26.51}{8.41}{-} & \cell{0.398}{0.305}{-} & \cell{0.956}{0.390}{+} & \cell{0.793}{0.135}{+} & \cell{0.613}{0.235}{+} & \cell{20.54}{13.09}{-} & \cell{0.442}{0.263}{-} & \cell{1.218}{0.353}{+} \\
                             & \textsf{ar} & \cell{0.929}{0.171}{+} & \cell{0.848}{0.202}{+} & \cell{21.87}{8.65}{-} & \cell{0.331}{0.322}{-} & \cell{0.621}{0.340}{+} & \cell{0.910}{0.153}{+} & \cell{0.658}{0.217}{+} & \cell{25.11}{12.69}{-} & \cell{0.416}{0.210}{-} & \cell{0.945}{0.307}{+} \\
                             & \textsf{vi} & \cell{0.899}{0.182}{+} & \cell{0.869}{0.313}{+} & \cell{15.49}{7.62}{-} & \cell{0.393}{0.314}{-} & \cell{0.667}{0.363}{+} & \cell{0.847}{0.063}{+} & \cell{0.640}{0.126}{+} & \cell{12.22}{5.76}{-} & \cell{0.399}{0.238}{-} & \cell{0.826}{0.225}{+} \\
                             & \textsf{sw} & \cell{0.717}{0.151}{+} & \cell{0.303}{0.040}{+} & \cell{21.60}{1.87}{+} & \cell{0.633}{0.160}{-} & \cell{0.510}{0.245}{-} & \cell{0.387}{0.117}{+} & \cell{0.189}{0.108}{+} & \cell{18.41}{1.63}{+} & \cell{0.727}{0.102}{-} & \cell{0.755}{0.495}{-} \\
                             & \textsf{ru} & \cell{0.960}{0.303}{+} & \cell{0.848}{0.343}{+} & \cell{16.50}{5.03}{-} & \cell{0.370}{0.339}{-} & \cell{0.661}{0.378}{+} & \cell{0.757}{0.144}{+} & \cell{0.568}{0.172}{+} & \cell{13.83}{6.40}{-} & \cell{0.457}{0.273}{-} & \cell{0.927}{0.336}{+} \\


    \bottomrule
  \end{tabular}
  }

  \caption{Automatic evaluation results for different reward ablations on \data{SIB200} and \data{Taxi1500} using \lm{Qwen3-8B}. Each cell reports the absolute metric value, and the number in parentheses indicates the absolute change relative to the full reward \our configuration.}

  \label{tab:automatic_evaluation_ablation}
\end{table*}

\begin{table*}[t!]
  \centering
  \footnotesize
  \setlength{\tabcolsep}{5.5pt}
  \renewcommand{\arraystretch}{1.08}

  \resizebox{2\columnwidth}{!}{%
  \begin{tabular}{@{}r l ccccc ccccc@{}}
    \toprule
    & & \multicolumn{5}{c}{\textbf{\data{SIB200}}}
      & \multicolumn{5}{c}{\textbf{\data{Taxi1500}}} \\
    \cmidrule(lr){3-7}\cmidrule(lr){8-12}
    \textbf{Setting} & \textbf{Lang}
      & \textbf{SLFR} \((\uparrow)\) & \textbf{HLFR} \((\uparrow)\) & \textbf{PPL} \((\downarrow)\) & \textbf{SS} \((\uparrow)\) & \textbf{TS} \((\downarrow)\)
      & \textbf{SLFR} \((\uparrow)\) & \textbf{HLFR} \((\uparrow)\) & \textbf{PPL} \((\downarrow)\) & \textbf{SS} \((\uparrow)\) & \textbf{TS} \((\downarrow)\)  \\
    \midrule


    \multirow{7}{*}{\textbf{\wos\ \Rflip}} 
        & \textsf{en} & \cell{0.588}{0.048}{-} & \cell{0.505}{0.071}{-} & \cell{32.81}{1.08}{+} & \cell{0.679}{0.029}{+} & \cell{0.381}{0.046}{-} & \cell{0.532}{0.063}{-} & \cell{0.441}{0.073}{-} & \cell{23.69}{0.15}{+} & \cell{0.608}{0.021}{+} & \cell{0.593}{0.013}{-} \\
        & \textsf{de} & \cell{0.515}{0.121}{-} & \cell{0.436}{0.109}{-} & \cell{25.29}{0.85}{+} & \cell{0.783}{0.008}{+} & \cell{0.317}{0.038}{-} & \cell{0.586}{0.072}{-} & \cell{0.423}{0.027}{-} & \cell{24.86}{1.82}{+} & \cell{0.713}{0.012}{-} & \cell{0.539}{0.012}{-} \\
        & \textsf{zh} & \cell{0.632}{0.166}{-} & \cell{0.559}{0.219}{-} & \cell{31.30}{1.23}{+} & \cell{0.740}{0.084}{+} & \cell{0.608}{0.058}{-} & \cell{0.532}{0.072}{-} & \cell{0.351}{0.018}{-} & \cell{29.89}{2.53}{-} & \cell{0.700}{0.030}{+} & \cell{0.943}{0.018}{-} \\
        & \textsf{ar} & \cell{0.647}{0.010}{-} & \cell{0.559}{0.087}{-} & \cell{28.18}{1.93}{+} & \cell{0.698}{0.048}{+} & \cell{0.362}{0.037}{-} & \cell{0.486}{0.082}{-} & \cell{0.306}{0.045}{-} & \cell{35.27}{4.78}{+} & \cell{0.667}{0.006}{+} & \cell{0.627}{0.032}{-} \\
        & \textsf{vi} & \cell{0.539}{0.107}{-} & \cell{0.456}{0.130}{-} & \cell{19.34}{0.64}{+} & \cell{0.754}{0.040}{+} & \cell{0.391}{0.046}{-} & \cell{0.414}{0.091}{-} & \cell{0.333}{0.036}{-} & \cell{17.48}{1.17}{+} & \cell{0.679}{0.018}{+} & \cell{0.576}{0.034}{-} \\
        & \textsf{sw} & \cell{0.623}{0.114}{-} & \cell{0.520}{0.086}{-} & \cell{17.55}{0.04}{-} & \cell{0.746}{0.078}{+} & \cell{0.379}{0.094}{-} & \cell{0.622}{0.018}{-} & \cell{0.405}{0.045}{+} & \cell{18.15}{1.72}{-} & \cell{0.732}{0.024}{-} & \cell{0.637}{0.028}{+} \\
        & \textsf{ru} & \cell{0.574}{0.062}{-} & \cell{0.475}{0.131}{-} & \cell{20.10}{0.96}{+} & \cell{0.778}{0.040}{+} & \cell{0.348}{0.087}{-} & \cell{0.514}{0.090}{-} & \cell{0.369}{0.063}{-} & \cell{19.15}{0.20}{+} & \cell{0.735}{0.053}{+} & \cell{0.622}{0.044}{-} \\
    \midrule

    \multirow{7}{*}{\textbf{\wos\ \Raug}} 
        & \textsf{en} & \cell{0.564}{0.072}{-} & \cell{0.490}{0.086}{-} & \cell{32.41}{0.68}{+} & \cell{0.663}{0.013}{+} & \cell{0.395}{0.032}{-} & \cell{0.577}{0.018}{-} & \cell{0.486}{0.028}{-} & \cell{23.56}{0.02}{+} & \cell{0.613}{0.026}{+} & \cell{0.584}{0.022}{-} \\
        & \textsf{de} & \cell{0.564}{0.072}{-} & \cell{0.456}{0.089}{-} & \cell{25.55}{1.11}{+} & \cell{0.753}{0.022}{-} & \cell{0.366}{0.011}{+} & \cell{0.622}{0.036}{-} & \cell{0.459}{0.009}{+} & \cell{23.62}{0.58}{+} & \cell{0.721}{0.004}{-} & \cell{0.530}{0.021}{-} \\
        & \textsf{zh} & \cell{0.701}{0.097}{-} & \cell{0.642}{0.136}{-} & \cell{30.12}{0.05}{+} & \cell{0.721}{0.065}{+} & \cell{0.672}{0.006}{+} & \cell{0.595}{0.009}{-} & \cell{0.432}{0.063}{+} & \cell{32.68}{0.26}{+} & \cell{0.693}{0.023}{+} & \cell{0.940}{0.021}{-} \\
        & \textsf{ar} & \cell{0.642}{0.015}{-} & \cell{0.549}{0.097}{-} & \cell{27.38}{1.13}{+} & \cell{0.694}{0.044}{+} & \cell{0.382}{0.017}{-} & \cell{0.541}{0.027}{-} & \cell{0.369}{0.018}{+} & \cell{33.00}{2.51}{+} & \cell{0.662}{0.001}{+} & \cell{0.659}{0.000}{0} \\
        & \textsf{vi} & \cell{0.574}{0.072}{-} & \cell{0.510}{0.076}{-} & \cell{18.85}{0.15}{+} & \cell{0.741}{0.027}{+} & \cell{0.420}{0.017}{-} & \cell{0.486}{0.019}{-} & \cell{0.396}{0.027}{+} & \cell{17.32}{1.01}{+} & \cell{0.662}{0.001}{+} & \cell{0.593}{0.017}{-} \\
        & \textsf{sw} & \cell{0.730}{0.007}{-} & \cell{0.623}{0.017}{+} & \cell{17.89}{0.30}{+} & \cell{0.728}{0.060}{+} & \cell{0.397}{0.076}{-} & \cell{0.541}{0.099}{-} & \cell{0.324}{0.036}{-} & \cell{17.43}{2.44}{-} & \cell{0.749}{0.007}{-} & \cell{0.622}{0.013}{+} \\
        & \textsf{ru} & \cell{0.588}{0.048}{-} & \cell{0.515}{0.091}{-} & \cell{20.30}{1.16}{+} & \cell{0.771}{0.033}{+} & \cell{0.382}{0.053}{-} & \cell{0.523}{0.081}{-} & \cell{0.414}{0.018}{-} & \cell{19.27}{0.32}{+} & \cell{0.712}{0.030}{+} & \cell{0.632}{0.034}{-} \\
    \midrule

    \multirow{7}{*}{\textbf{\wos\ \Redit}} 
        & \textsf{en} & \cell{0.789}{0.153}{+} & \cell{0.750}{0.174}{+} & \cell{29.29}{2.44}{-} & \cell{0.506}{0.144}{-} & \cell{0.689}{0.262}{+} & \cell{0.676}{0.081}{+} & \cell{0.559}{0.045}{+} & \cell{22.51}{1.03}{-} & \cell{0.441}{0.146}{-} & \cell{0.732}{0.126}{+} \\
        & \textsf{de} & \cell{0.730}{0.094}{+} & \cell{0.686}{0.141}{+} & \cell{22.97}{1.47}{-} & \cell{0.667}{0.108}{-} & \cell{0.538}{0.183}{+} & \cell{0.676}{0.018}{+} & \cell{0.505}{0.055}{+} & \cell{23.45}{0.41}{+} & \cell{0.585}{0.140}{-} & \cell{0.686}{0.135}{+} \\
        & \textsf{zh} & \cell{0.755}{0.043}{-} & \cell{0.730}{0.048}{-} & \cell{27.40}{2.67}{-} & \cell{0.634}{0.022}{-} & \cell{0.830}{0.164}{+} & \cell{0.676}{0.072}{+} & \cell{0.514}{0.145}{+} & \cell{23.94}{8.48}{-} & \cell{0.552}{0.118}{-} & \cell{1.367}{0.406}{+} \\
        & \textsf{ar} & \cell{0.799}{0.142}{+} & \cell{0.725}{0.079}{+} & \cell{24.34}{1.91}{-} & \cell{0.567}{0.083}{-} & \cell{0.576}{0.177}{+} & \cell{0.703}{0.135}{+} & \cell{0.523}{0.172}{+} & \cell{27.42}{3.07}{-} & \cell{0.568}{0.093}{-} & \cell{0.926}{0.267}{+} \\
        & \textsf{vi} & \cell{0.716}{0.070}{+} & \cell{0.672}{0.086}{+} & \cell{15.77}{2.93}{-} & \cell{0.633}{0.081}{-} & \cell{0.626}{0.189}{+} & \cell{0.568}{0.063}{+} & \cell{0.468}{0.099}{+} & \cell{14.56}{1.75}{-} & \cell{0.557}{0.104}{-} & \cell{0.733}{0.123}{+} \\
        & \textsf{sw} & \cell{0.833}{0.096}{+} & \cell{0.721}{0.115}{+} & \cell{16.17}{1.42}{-} & \cell{0.672}{0.004}{+} & \cell{0.554}{0.081}{+} & \cell{0.658}{0.018}{+} & \cell{0.432}{0.072}{+} & \cell{16.99}{2.88}{-} & \cell{0.684}{0.072}{-} & \cell{0.816}{0.207}{+} \\
        & \textsf{ru} & \cell{0.770}{0.134}{+} & \cell{0.706}{0.100}{+} & \cell{17.75}{1.39}{-} & \cell{0.660}{0.078}{-} & \cell{0.607}{0.172}{+} & \cell{0.631}{0.027}{+} & \cell{0.514}{0.082}{+} & \cell{16.73}{2.22}{-} & \cell{0.624}{0.058}{-} & \cell{0.781}{0.115}{+} \\

    \bottomrule
  \end{tabular}
  }

  \caption{Automatic evaluation results for different reward ablations on \data{SIB200} and \data{Taxi1500} using \lm{Gemma3-4B}. Each cell reports the absolute metric value, and the number in parentheses indicates the absolute change relative to the full reward \our configuration.}
  \label{tab:automatic_evaluation_ablation_gemma4}
\end{table*}

\subsection{Ablation Results}
\label{subsec:ablation_full_app}
We report ablation study results to analyze the role of each reward component in \our. 
Specifically, we remove \Rflip, \Raug, and \Redit one at a time and compare the resulting models against the full reward setting. 
Table~\ref{tab:automatic_evaluation_ablation} and Table~\ref{tab:automatic_evaluation_ablation_gemma4} report the automatic evaluation results on \data{SIB200} and \data{Taxi1500} using \lm{Qwen3-8B} and \lm{Gemma3-4B}.
Overall, removing \Rflip and \Raug most consistently reduces flip rates, while removing \Redit increases flip success at the cost of larger edits and lower semantic similarity.

\subsection{Additional Analysis}
\label{subsec:additional_analysis}
To systematically compare how each language responds to component removal, based on Table~\ref{tab:automatic_evaluation_ablation}, we compute the \textit{relative degradation} for each metric as $\Delta_{\text{rel}} = \delta / v_{\text{full}}$, where $\delta$ is the absolute change reported in parentheses and $v_{\text{full}}$ is the corresponding full-model value. We then aggregate these relative changes into three dimensions, i.e., \textit{validity} (average across SLFR and HLFR (Appendix~\ref{subsec:hlfr})), \textit{fluency} (PPL), and \textit{minimality} (average across SS and TS (Appendix~\ref{subsec:ts})), averaged over both datasets. A positive value indicates degradation (validity dropped, PPL increased, or minimality worsened), while a negative value indicates improvement. Tables~\ref{tab:flip}-\ref{tab:aug} present the per-setting results, and Table~\ref{tab:avg} shows the overall average across all three settings.

\begin{table*}[t!]
\centering
\renewcommand*{\arraystretch}{0.85}
\resizebox{\textwidth}{!}{%
\begin{tabular}{cc|cc|cc|cc||cc|cc|cc}
\toprule[1.5pt]
\multicolumn{2}{c|}{\textbf{Dataset}} & \multicolumn{6}{c||}{\textbf{\data{SIB200}}} & \multicolumn{6}{c}{\textbf{\data{TAXI1500}}}\\
\cmidrule(l){1-14}
\multirow{2}{*}{\rotatebox[origin=c]{90}{\scriptsize{\textbf{Model}}}} & \textbf{Lang-}
& \multicolumn{2}{c|}{\textbf{SLFR} (\(\uparrow\))}
& \multicolumn{2}{c|}{\textbf{PPL} (\(\downarrow\))}
& \multicolumn{2}{c||}{\textbf{SS} (\(\uparrow\))}
& \multicolumn{2}{c|}{\textbf{SLFR} (\(\uparrow\))}
& \multicolumn{2}{c|}{\textbf{PPL} (\(\downarrow\))}
& \multicolumn{2}{c}{\textbf{SS} (\(\uparrow\))}\\
 & \textbf{uage}
 & \textbf{DG-CF} & \textbf{\our}
 & \textbf{DG-CF} & \textbf{\our}
 & \textbf{DG-CF} & \textbf{\our}
 & \textbf{DG-CF} & \textbf{\our}
 & \textbf{DG-CF} & \textbf{\our}
 & \textbf{DG-CF} & \textbf{\our}\\
\midrule

\centering \multirow{6}{*}{\rotatebox[origin=c]{90}{\lm{Gemma3-4B}}} & \textsf{bg} & \bestresult{0.632} & 0.627 & \bestresult{14.78} & 15.61 & 0.722 & \bestresult{0.768} & - & - & - & - & - & - \\
 & \textsf{el} & 0.627 & \bestresult{0.672} & \bestresult{9.65} & 10.69 & \bestresult{0.691} & 0.685 & - & - & - & - & - & - \\
 & \textsf{es} & \bestresult{0.588} & 0.559 & \bestresult{21.17} & 23.92 & 0.709 & \bestresult{0.750} & 0.595 & \bestresult{0.622} & 19.33 & \bestresult{19.00} & 0.658 & \bestresult{0.702} \\
 & \textsf{hi} & 0.676 & \bestresult{0.740} & \bestresult{4.52} & 4.77 & \bestresult{0.682} & 0.681 & 0.640 & \bestresult{0.649} & \bestresult{5.92} & 5.99 & 0.635 & \bestresult{0.719} \\
 & \textsf{th} & 0.618 & \bestresult{0.637} & \bestresult{6.60} & 7.26 & 0.674 & \bestresult{0.683} & 0.595 & \bestresult{0.631} & 8.94 & \bestresult{8.57} & 0.574 & \bestresult{0.636} \\
 & \textsf{tr} & 0.642 & \bestresult{0.657} & \bestresult{16.20} & 18.68 & 0.676 & \bestresult{0.710} & - & - & - & - & - & - \\

\midrule

\centering \multirow{6}{*}{\rotatebox[origin=c]{90}{\lm{Qwen3-4B}}} & \textsf{bg} & 0.446 & \bestresult{0.657} & \bestresult{17.52} & 19.36 & \bestresult{0.859} & 0.813 & - & - & - & - & - & - \\
 & \textsf{el} & 0.446 & \bestresult{0.515} & \bestresult{12.41} & 12.55 & \bestresult{0.806} & 0.800 & - & - & - & - & - & - \\
 & \textsf{es} & 0.544 & \bestresult{0.711} & \bestresult{23.53} & 26.21 & \bestresult{0.793} & 0.740 & 0.505 & \bestresult{0.586} & \bestresult{19.05} & 21.93 & 0.665 & \bestresult{0.743} \\
 & \textsf{hi} & 0.412 & \bestresult{0.534} & \bestresult{5.22} & 5.32 & \bestresult{0.822} & 0.792 & \bestresult{0.387} & 0.270 & \bestresult{6.41} & 6.94 & 0.676 & \bestresult{0.793} \\
 & \textsf{th} & 0.583 & \bestresult{0.613} & \bestresult{7.98} & 8.31 & \bestresult{0.748} & 0.703 & 0.595 & \bestresult{0.622} & \bestresult{8.97} & 10.00 & 0.619 & \bestresult{0.688} \\
 & \textsf{tr} & 0.583 & \bestresult{0.632} & \bestresult{20.70} & 22.75 & \bestresult{0.762} & 0.736 & - & - & - & - & - & - \\

\bottomrule[1.5pt]
\end{tabular}%
}
\caption{Zero-shot evaluation results on \data{SIB200} and \data{TAXI1500} for languages unseen during training. We compare DG-CF and \our across three metrics: soft label flipping rate (SLFR (\(\uparrow\))), perplexity (PPL (\(\downarrow\))), and semantic similarity (SS (\(\uparrow\))). \bestlegend denote the best results for each language-metric pair, respectively.}
\label{tab:zero_shot_unseen_languages}
\end{table*}

\begin{table}[ht]
\centering
\begin{tabular}{lccc}
\toprule
\textbf{Lang} & \textbf{Validity} & \textbf{Fluency} & \textbf{Minimality} \\
\midrule
en & $+19.18$ & $-3.33$ & $-6.87$ \\
de & $+14.28$ & $-3.63$ & $-4.52$ \\
zh & $+14.43$ & $-3.59$ & $-3.68$ \\
ar & $+10.12$ & $+12.06$ & $-5.65$ \\
vi & $+6.43$ & $-3.55$ & $-4.76$ \\
sw & $+7.88$ & $-2.10$ & $-36.27$ \\
ru & $+9.20$ & $-2.12$ & $-3.48$ \\
\bottomrule
\end{tabular}
\caption{Relative degradation (\%) under w/o $\mathcal{R}_{\text{flip}}$.}
\label{tab:flip}
\end{table}

\begin{table}[ht]
\centering

\begin{tabular}{lccc}
\toprule
\textbf{Lang} & \textbf{Validity} & \textbf{Fluency} & \textbf{Minimality} \\
\midrule
en & $+12.45$ & $+2.35$ & $-8.34$ \\
de & $+12.94$ & $-3.17$ & $-6.34$ \\
zh & $+10.39$ & $+4.61$ & $-4.85$ \\
ar & $+8.57$ & $+8.34$ & $-5.58$ \\
vi & $+8.93$ & $-2.81$ & $-2.78$ \\
sw & $+12.47$ & $-4.46$ & $-37.42$ \\
ru & $+4.81$ & $-4.90$ & $-3.50$ \\
\bottomrule
\end{tabular}
\caption{Relative degradation (\%) under w/o $\mathcal{R}_{\text{aug}}$.}
\label{tab:aug}
\end{table}

\begin{table}[ht]
\centering
\begin{tabular}{lccc}
\toprule
\textbf{Lang} & \textbf{Validity} & \textbf{Fluency} & \textbf{Minimality} \\
\midrule
en & $-29.01$ & $+54.75$ & $+65.20$ \\
de & $-32.57$ & $+48.44$ & $+65.93$ \\
zh & $-32.17$ & $+111.56$ & $+47.60$ \\
ar & $-30.81$ & $+83.80$ & $+63.19$ \\
vi & $-28.56$ & $+93.00$ & $+59.66$ \\
sw & $-54.64$ & $+9.60$ & $+78.83$ \\
ru & $-33.85$ & $+65.00$ & $+62.64$ \\
\bottomrule
\end{tabular}
\caption{Relative degradation (\%) under w/o $\mathcal{R}_{\text{edit}}$.}
\label{tab:edit}
\end{table}

\begin{table}[t]
\centering
\begin{tabular}{lcccc}
\toprule
\textbf{Lang} & $\overline{|\Delta_{\text{val}}|}$ & $\overline{|\Delta_{\text{flu}}|}$ & $\overline{|\Delta_{\text{min}}|}$ & \textbf{Overall} \\
\midrule
sw & $25.00$ & $5.39$ & $50.84$ & $27.07$ \\
zh & $19.00$ & $39.92$ & $18.71$ & $25.87$ \\
ar & $16.50$ & $34.73$ & $24.81$ & $25.35$ \\
vi & $14.64$ & $33.12$ & $22.40$ & $23.39$ \\
en & $20.21$ & $20.15$ & $26.80$ & $22.39$ \\
de & $19.93$ & $18.41$ & $25.60$ & $21.31$ \\
ru & $15.95$ & $24.00$ & $23.20$ & $21.05$ \\
\bottomrule
\end{tabular}
\caption{Average absolute relative change (\%) across all three ablation settings, and overall sensitivity (mean of three dimensions). Higher values indicate greater sensitivity to component removal.}
\label{tab:avg}
\end{table}

\subsubsection{Language-Specific Observation}
\paragraph{Analysis of w/o $\mathcal{R}_{\text{flip}}$ (Table~\ref{tab:flip}).}
Removing the flip reward degrades \textit{validity} across all languages, with English suffering the most ($+19.18\%$), followed by Chinese and German. Vietnamese and Swahili are the least affected ($+6.43\%$ and $+7.88\%$). \textit{Minimality} improves modestly for most languages, but Swahili shows a dramatic degradation of $36.27\%$, which reveals that the full model's Swahili counterfactuals involve relatively large edit distances, and without flip pressure, the model collapses to near-paraphrases.

\paragraph{Analysis of w/o $\mathcal{R}_{\text{aug}}$ (Table~\ref{tab:aug}).}
The augmentation reward primarily affects \textit{validity}, with all languages showing consistent degradation. English, German, and Swahili are the most affected. Russian is the least affected ($+4.81\%$), suggesting that its counterfactuals already cross the decision boundary with sufficient margin without augmentation. Minimality is moderately degraded. Arabic again stands out with $+8.34\%$ fluency degradation, further evidencing its unique sensitivity to any modification of the reward structure. Swahili again shows the anomalous minimality deterioration ($-37.42\%$).

\paragraph{Analysis of w/o $\mathcal{R}_{\text{edit}}$ (Table~\ref{tab:edit}).}
This setting produces the most extreme degradation patterns. \textit{Validity} improves substantially for all languages (the edit constraint was \textit{restraining} flip success), with Swahili showing the largest gain ($54.64\%$)---its SLFR and HLFR roughly double without the edit constraint. However, this comes at severe cost to \textit{fluency} and \textit{minimality}. Chinese suffers the worst fluency degradation ($+111.56\%$, i.e., PPL more than doubles), followed by Vietnamese and Arabic. Swahili's fluency degradation is remarkably contained ($+9.60\%$), likely because its baseline PPL is already high due to limited pre-training coverage, leaving less room for relative worsening. \textit{Minimality} degrades dramatically across all languages, with Swahili showing the worst ($+78.83\%$), followed by German and English. Chinese has the lowest minimality degradation ($+47.60\%$) under this setting.

\begin{table}[t]
\centering
\small
\setlength{\tabcolsep}{4pt}
\begin{tabular}{ll ccc}
\toprule[1.5pt]
\textbf{Setting} & \textbf{Method} & \textbf{SLFR} $\uparrow$ & \textbf{PPL} $\downarrow$ & \textbf{SS} $\uparrow$ \\
\midrule
\multirow{3}{*}{\shortstack[l]{\lm{Gemma3-4B}\\\data{SIB200}}}
 & DG-CF & 0.62 & 28.3 & 0.69 \\
 & \our w/o $\mathcal{R}_{\text{aug}}$ & 0.62 & 24.6 & \bestresult{0.72} \\
 & \our & \bestresult{0.68} & \bestresult{24.0} & 0.69 \\
\midrule
\multirow{3}{*}{\shortstack[l]{\lm{Gemma3-4B}\\\data{TAXI1500}}}
 & DG-CF & 0.53 & 31.0 & 0.62 \\
 & \our w/o $\mathcal{R}_{\text{aug}}$ & 0.56 & 23.8 & \bestresult{0.69} \\
 & \our & \bestresult{0.60} & \bestresult{23.5} & 0.68 \\
\midrule
\multirow{3}{*}{\shortstack[l]{\lm{Qwen3-8B}\\\data{SIB200}}}
 & DG-CF & 0.61 & 32.4 & \bestresult{0.72} \\
 & \our w/o $\mathcal{R}_{\text{aug}}$ & 0.63 & \bestresult{27.1} & 0.70 \\
 & \our & \bestresult{0.70} & 27.9 & 0.70 \\
\midrule
\multirow{3}{*}{\shortstack[l]{\lm{Qwen3-8B}\\\data{TAXI1500}}}
 & DG-CF & 0.50 & 28.9 & 0.67 \\
 & \our w/o $\mathcal{R}_{\text{aug}}$ & 0.62 & 26.6 & \bestresult{0.73} \\
 & \our & \bestresult{0.64} & \bestresult{24.9} & 0.70 \\
\bottomrule[1.5pt]
\end{tabular}
\caption{\our without $\mathcal{R}_{\text{aug}}$, the configuration applicable to black-box models, compared with DG-CF and the full method. All values are averaged over the seven languages. \bestlegend marks the best result per metric.}
\label{tab:blackbox}
\end{table}

\subsubsection{Overall Language Sensitivity}
\textbf{Swahili} emerges as the most sensitive language overall, in particular its extreme minimality and validity sensitivity. This oscillation between extremes, unique to Swahili, likely stems from its limited pre-training representation, causing the model's behavior to be highly contingent on the specific reward configuration. In contrast, Swahili's fluency sensitivity is the lowest ($5.39\%$), indicating that its generation quality is relatively stable---possibly because the baseline PPL is already high and leaves little room for relative change. \textbf{Chinese} and \textbf{Arabic} rank second and third overall, but their sensitivity profiles are qualitatively different from Swahili's. Both are dominated by fluency sensitivity. \textbf{English} and \textbf{German} display the most balanced sensitivity profiles, with all three dimensions falling within a relatively narrow range. English has the highest validity sensitivity among high-resource languages, confirming that its label-flip success depends most critically on the collaboration among all three rewards. German tracks closely behind English in all dimensions. \textbf{Russian} is the least sensitive language overall ($21.05\%$), with the lowest validity sensitivity ($15.95\%$). This suggests that Russian counterfactuals are inherently close to the decision boundary and relatively well-formed, making each individual reward component less critical. \textbf{Vietnamese} occupies the middle ground, with elevated fluency sensitivity ($33.12\%$) comparable to Chinese and Arabic, but more moderate validity and minimality profiles.

\section{Counterfactual Data Augmentation}
\label{app:cda}

Beyond serving as explanations, SCEs have practical value as training data, where counterfactual data augmentation (CDA) improves downstream task performance and robustness \cite{qiu-etal-2024-paircfr, nguyen-etal-2024-llms, wang-etal-2025-truth}. We test whether \our's SCEs serve this purpose, using \lm{Gemma3-12B}.

\subsection{Setup}

Let $\mathcal{D}_{\text{train}}^{(\ell)} = \{(x_i^{(\ell)}, y_i^{*(\ell)})\}_{i=1}^{N}$ denote the training split in language $\ell$, where $y_i^{*(\ell)}$ is the gold label, and let $\mathcal{D}_{\text{train}} = \cup_{\ell \in \mathcal{L}} \mathcal{D}_{\text{train}}^{(\ell)}$. SFT$_{\text{orig}}$ is trained on $\mathcal{D}_{\text{train}}$ alone, whereas CDA is trained on the original data together with the corresponding counterfactuals,
\begin{align*}
\mathcal{D}_{\text{CDA}} = \mathcal{D}_{\text{train}} \cup \{(\tilde{x}_i^{(\ell)}, \hat{y}_i^{(\ell)}) \mid \mathcal{M}(\tilde{x}_i^{(\ell)}) = \hat{y}_i^{(\ell)}\},
\end{align*}
i.e., only counterfactuals that reach their target label are added, at most one per instance. We construct $\mathcal{D}_{\text{CDA}}$ separately from \our, DG-CF, and SFT counterfactuals, keeping everything else fixed, and train all models with LoRA for two epochs using the configuration of Table~\ref{tab:training-hparams}. All models are evaluated on the original test set. Zero-shot denotes the untuned model $\mathcal{M}$ evaluated on identical test set.

\begin{table}[t]
\centering
\small
\setlength{\tabcolsep}{5pt}
\begin{tabular}{llccc}
\toprule
\textbf{Model} & \textbf{Temp.} & \textbf{SLFR} ($\uparrow$) & \textbf{HLFR} ($\uparrow$) & \textbf{Uniq.} \\
\midrule
 \multirow{3}{*}{\lm{Qwen3-4B}} & 0.7 & 89.8 & 12.6 & 3.78 \\
  & 1.1 & 89.0 & 14.8 & 5.30 \\
  & 1.5 & 89.8 & 14.8 & 6.48 \\
\cmidrule(l){1-5}
 \multirow{3}{*}{\lm{Gemma3-4B}} & 0.7 & 62.0 & 53.0 & 7.02 \\
  & 1.1 & 59.2 & 52.4 & 8.46 \\
  & 1.5 & 57.8 & 52.4 & 8.98 \\
\bottomrule
\end{tabular}
\caption{Temperature sweep on English \data{SIB200}, sampling ten candidates per input. SLFR and HLFR are reported in \%. Uniq.\ is the average number of distinct candidates among the ten.}
\label{tab:temperature}
\end{table}
\begin{table}[t]
\centering
\small
\begin{tabular}{cc|cc}
\toprule
\textbf{Model} & \textbf{Language} & \data{SIB200} & \data{TAXI1500} \\
\midrule
 \multirow{8}{*}{\rotatebox[origin=c]{90}{\lm{Qwen3-4B}}} & \textsf{en} & 77.9 & 93.7 \\
  & \textsf{de} & 92.7 & 91.3 \\
  & \textsf{zh} & 96.6 & 95.2 \\
  & \textsf{ar} & 96.4 & 98.8 \\
  & \textsf{vi} & 93.9 & 97.1 \\
  & \textsf{sw} & 97.0 & 97.2 \\
  & \textsf{ru} & 80.0 & 94.5 \\
  & \textbf{Avg} & \textbf{90.6} & \textbf{95.4} \\
\midrule
 \multirow{8}{*}{\rotatebox[origin=c]{90}{\lm{Qwen3-8B}}} & \textsf{en} & 76.2 & 96.7 \\
  & \textsf{de} & 80.5 & 98.7 \\
  & \textsf{zh} & 77.7 & 92.3 \\
  & \textsf{ar} & 91.3 & 97.0 \\
  & \textsf{vi} & 87.9 & 94.3 \\
  & \textsf{sw} & 78.9 & 99.7 \\
  & \textsf{ru} & 78.5 & 96.5 \\
  & \textbf{Avg} & \textbf{81.6} & \textbf{96.5} \\
\midrule
 \multirow{8}{*}{\rotatebox[origin=c]{90}{\lm{Gemma3-4B}}} & \textsf{en} & 89.7 & 83.4 \\
  & \textsf{de} & 87.6 & 90.6 \\
  & \textsf{zh} & 92.3 & 87.2 \\
  & \textsf{ar} & 92.2 & 85.2 \\
  & \textsf{vi} & 91.3 & 80.9 \\
  & \textsf{sw} & 94.0 & 93.5 \\
  & \textsf{ru} & 92.3 & 85.9 \\
  & \textbf{Avg} & \textbf{91.3} & \textbf{86.7} \\
\midrule
 \multirow{8}{*}{\rotatebox[origin=c]{90}{\lm{Gemma3-12B}}} & \textsf{en} & 98.6 & 95.0 \\
  & \textsf{de} & 99.1 & 93.7 \\
  & \textsf{zh} & 98.0 & 90.0 \\
  & \textsf{ar} & 99.7 & 95.8 \\
  & \textsf{vi} & 99.0 & 89.8 \\
  & \textsf{sw} & 99.3 & 93.4 \\
  & \textsf{ru} & 98.4 & 89.7 \\
  & \textbf{Avg} & \textbf{98.9} & \textbf{92.5} \\
\bottomrule
\end{tabular}
\caption{Retention rate (\%) of training inputs, i.e., the fraction with at least one flipping candidate among the ten sampled.}
\label{tab:retention}
\end{table}
\begin{table}[t]
\centering
\small
\setlength{\tabcolsep}{5pt}
\begin{tabular}{lcc}
\toprule
\textbf{Model} & \textbf{Retained} & \textbf{Discarded} \\
\midrule
 \lm{Qwen3-4B}   & 8.11 & 4.47 \\
 \lm{Qwen3-8B}   & 8.28 & 5.99 \\
 \lm{Gemma3-4B}  & 9.29 & 8.52 \\
 \lm{Gemma3-12B} & 9.75 & 9.60 \\
\bottomrule
\end{tabular}
\caption{Average number of distinct candidates among the ten sampled, for retained and discarded inputs, pooled over both datasets and all languages.}
\label{tab:retention_diversity}
\end{table}

\subsection{Results}

Table~\ref{tab:cda_gemma12} shows that, for \lm{Gemma3-12B}, including counterfactuals generated by \our consistently outperforms both SFT$_{\text{orig}}$ and CDA from DG-CF and SFT on both datasets when averaged over all languages. Among the three CDA conditions, \our is the only one that outperforms SFT$_{\text{orig}}$ on both datasets, whereas DG-CF falls below it on almost every language and SFT is inconsistent across datasets. Couterfactual data augmentation thus helps only when the added counterfactuals are of sufficient quality, indicating that \our's SCEs are more reliable as training data than those of either baseline.

\section{Applicability to Black-box Models}
\label{app:blackbox}
\our requires logit access only for $\mathcal{R}_{\text{aug}}$, since $\mathcal{R}_{\text{flip}}$ needs only the predicted label and $\mathcal{R}_{\text{edit}}$ only the original input and its counterfactual, both of which are obtainable through a black-box model API. \our without $\mathcal{R}_{\text{aug}}$ (Tables~\ref{tab:automatic_evaluation_ablation} and \ref{tab:automatic_evaluation_ablation_gemma4}) can therefore be applied to explain black-box models, and Table~\ref{tab:blackbox} summarizes this configuration against DG-CF and \our with full scores.

The w/o $\mathcal{R}_{\text{aug}}$ ablation improves over DG-CF on all three metrics in most settings. Relative to the \our with full scores it loses \textit{validity}, as expected from the role of $\mathcal{R}_{\text{aug}}$ (\S\ref{subsec:ablation_study}), while \textit{minimality} is comparable or better. Preference optimization for multilingual SCE generation is thus deployable without logit access, with $\mathcal{R}_{\text{aug}}$ acting as an optional refinement.

\begin{figure*}[h]
  \centering
  \includegraphics[width=\textwidth]{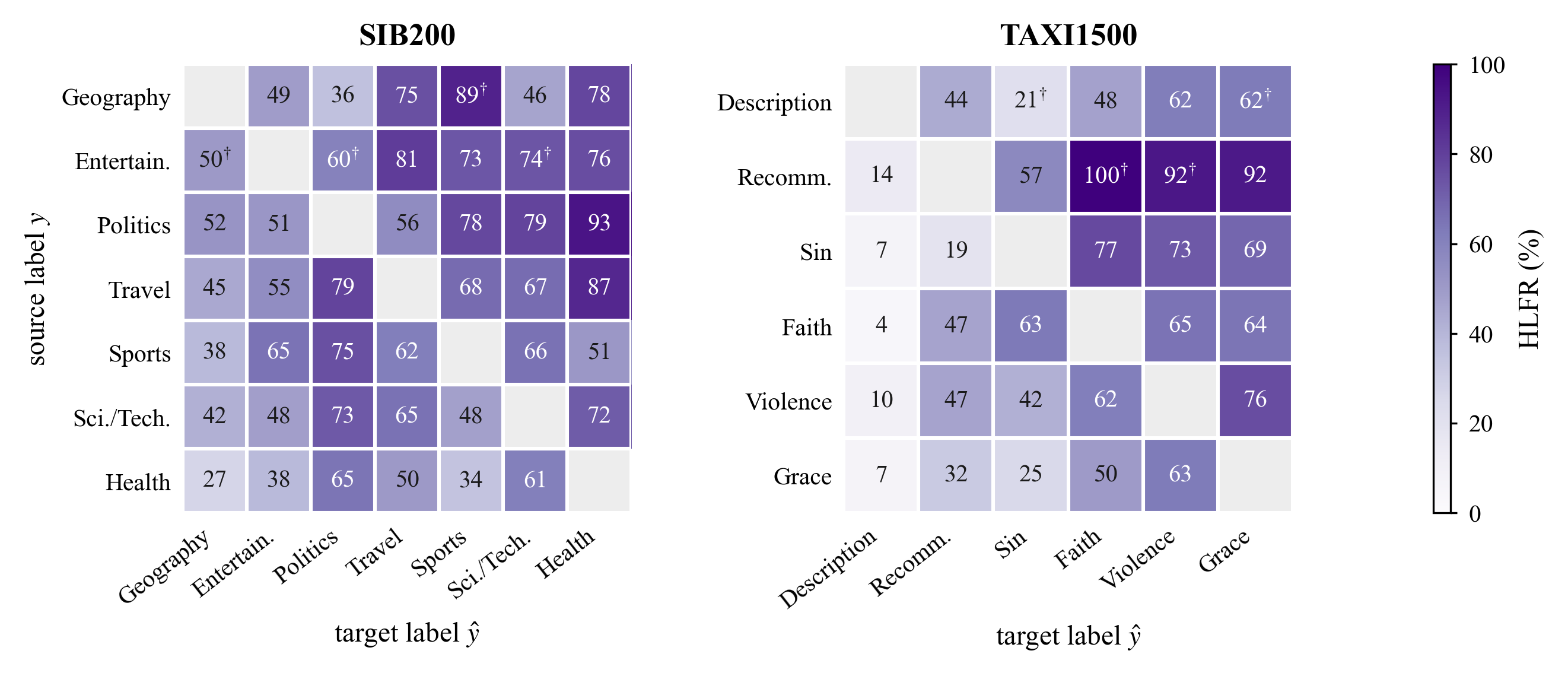}
  \caption{Hard label flipping rate (\%) for every source--target label pair, pooled over four models and seven languages.}
  \label{fig:transition_heatmap}
\end{figure*}

\section{Target-Label Transition Difficulty}
\label{app:transition}
Label transitions differ substantially in difficulty. Figure~\ref{fig:transition_heatmap} reports HLFR (\S\ref{subsec:hlfr}) for every source--target pair, pooled over all four models and seven languages. The matrix is strongly asymmetric, as the two directions of a pair differ by 32\% on average on \data{SIB200} and 82\% on \data{TAXI1500} in relative terms. Flip difficulty is thus a property of the ordered pair rather than of the semantic distance between two classes.

\section{Sampling Temperature Sensitivity}
\label{app:temperature}
We sample ten candidates per input at a temperature of $1.1$ in all experiments (Appendix~\ref{app:training}). To check how much this choice matters, we sweep the temperature over $\{0.7, 1.1, 1.5\}$ on English \data{SIB200} with two models. Table~\ref{tab:temperature} shows that flip rates move by at most a few points and in no consistent direction across the two models, whereas the candidates become steadily more distinct as the temperature rises. Temperature thus mainly affects how many distinct candidates the ranking stage can choose from, rather than whether they flip the prediction.

\section{Preference Pair Retention}
\label{app:retention}
An input enters DPO training only if at least one of its ten sampled candidates flips the model prediction (\S\ref{subsec:dataset_optimization}). Training could therefore be biased toward inputs that are easy to flip. Table~\ref{tab:retention} shows that 91.8\% of inputs are retained overall. Retention differs more across models and datasets than across languages, and it is not lower for the lower-resource ones: Arabic or Swahili is the most retained language for every model and dataset, while English is among the least retained in most cases. The discarded part is thus small and does not fall mainly on the languages where such a bias would matter most.
Table~\ref{tab:retention_diversity} shows what the discarded inputs look like. For the two \lm{Qwen3} family models, they produce far fewer distinct candidates than the retained ones, so the ten samples were nearly identical and left nothing for the ranking stage to compare. For the \lm{Gemma3} family models this gap is small, so their discarded inputs are more likely to be genuinely difficult.

\begin{table}[ht]
\centering
\small
\setlength{\tabcolsep}{3pt}
\begin{tabular}{cc ccc @{\hspace{1em}} ccc}
\toprule[1.5pt]
\multirow{2}{*}{\rotatebox[origin=c]{90}{\scriptsize{\textbf{Model}}}} & \multirow{2}{*}{\textbf{Lang.}} & \multicolumn{3}{c}{\data{SIB200}} & \multicolumn{3}{c}{\data{TAXI1500}} \\
\cmidrule(lr){3-5} \cmidrule(lr){6-8}
 & & \textbf{Base} & \textbf{\our} & $\Delta$ & \textbf{Base} & \textbf{\our} & $\Delta$ \\
\midrule
 \multirow{8}{*}{\rotatebox[origin=c]{90}{\lm{Qwen3-4B}}} & \textsf{en} & 80.4 & 81.4 & $+1.2$ & 36.0 & 37.8 & $+5.0$ \\
  & \textsf{de} & 80.4 & 80.9 & $+0.6$ & 29.7 & 33.3 & $+12.1$ \\
  & \textsf{zh} & 82.8 & 83.3 & $+0.6$ & 38.7 & 41.4 & $+7.0$ \\
  & \textsf{ar} & 82.4 & 81.4 & $-1.2$ & 44.1 & 45.0 & $+2.0$ \\
  & \textsf{vi} & 76.5 & 77.0 & $+0.7$ & 36.0 & 37.8 & $+5.0$ \\
  & \textsf{sw} & 57.4 & 57.8 & $+0.7$ & 18.9 & 21.6 & $+14.3$ \\
  & \textsf{ru} & 82.8 & 83.3 & $+0.6$ & 35.1 & 37.8 & $+7.7$ \\
  & \textbf{Avg} & \textbf{77.5} & \textbf{77.9} & $\mathbf{+0.4}$ & \textbf{34.1} & \textbf{36.4} & $\mathbf{+6.8}$ \\
\midrule
 \multirow{8}{*}{\rotatebox[origin=c]{90}{\lm{Qwen3-8B}}} & \textsf{en} & 82.8 & 82.4 & $-0.5$ & 25.2 & 30.6 & $+21.4$ \\
  & \textsf{de} & 78.4 & 79.4 & $+1.3$ & 21.6 & 23.4 & $+8.3$ \\
  & \textsf{zh} & 77.9 & 78.4 & $+0.6$ & 43.2 & 45.0 & $+4.2$ \\
  & \textsf{ar} & 79.9 & 80.4 & $+0.6$ & 48.6 & 47.7 & $-1.9$ \\
  & \textsf{vi} & 82.4 & 82.4 & $\pm0.0$ & 24.3 & 30.6 & $+25.9$ \\
  & \textsf{sw} & 60.3 & 63.2 & $+4.8$ & 15.3 & 17.1 & $+11.8$ \\
  & \textsf{ru} & 81.4 & 81.9 & $+0.6$ & 24.3 & 27.0 & $+11.1$ \\
  & \textbf{Avg} & \textbf{77.6} & \textbf{78.3} & $\mathbf{+0.9}$ & \textbf{28.9} & \textbf{31.6} & $\mathbf{+9.3}$ \\
\midrule
 \multirow{8}{*}{\rotatebox[origin=c]{90}{\lm{Gemma3-4B}}} & \textsf{en} & 80.9 & 81.9 & $+1.2$ & 38.7 & 40.5 & $+4.7$ \\
  & \textsf{de} & 84.3 & 84.3 & $\pm0.0$ & 49.5 & 50.5 & $+2.0$ \\
  & \textsf{zh} & 81.4 & 81.4 & $\pm0.0$ & 40.5 & 41.4 & $+2.2$ \\
  & \textsf{ar} & 83.3 & 83.3 & $\pm0.0$ & 45.0 & 45.0 & $\pm0.0$ \\
  & \textsf{vi} & 80.4 & 80.9 & $+0.6$ & 44.1 & 45.0 & $+2.0$ \\
  & \textsf{sw} & 80.9 & 80.9 & $\pm0.0$ & 52.3 & 51.4 & $-1.7$ \\
  & \textsf{ru} & 81.4 & 81.4 & $\pm0.0$ & 47.7 & 46.8 & $-1.9$ \\
  & \textbf{Avg} & \textbf{81.8} & \textbf{82.0} & $\mathbf{+0.3}$ & \textbf{45.4} & \textbf{45.8} & $\mathbf{+0.9}$ \\
\midrule
 \multirow{8}{*}{\rotatebox[origin=c]{90}{\lm{Gemma3-12B}}} & \textsf{en} & 87.7 & 86.8 & $-1.0$ & 50.5 & 51.4 & $+1.8$ \\
  & \textsf{de} & 89.7 & 89.7 & $\pm0.0$ & 51.4 & 50.5 & $-1.8$ \\
  & \textsf{zh} & 85.8 & 86.3 & $+0.6$ & 42.3 & 42.3 & $\pm0.0$ \\
  & \textsf{ar} & 88.2 & 88.7 & $+0.6$ & 59.5 & 57.7 & $-3.0$ \\
  & \textsf{vi} & 89.2 & 88.7 & $-0.6$ & 46.8 & 46.8 & $\pm0.0$ \\
  & \textsf{sw} & 85.3 & 84.8 & $-0.6$ & 48.6 & 49.5 & $+1.9$ \\
  & \textsf{ru} & 88.7 & 88.2 & $-0.6$ & 44.1 & 44.1 & $\pm0.0$ \\
  & \textbf{Avg} & \textbf{87.8} & \textbf{87.6} & $\mathbf{-0.2}$ & \textbf{49.0} & \textbf{48.9} & $\mathbf{-0.3}$ \\
\bottomrule[1.5pt]
\end{tabular}
\caption{Classification accuracy (\%) on the original test inputs before and after \our training. $\Delta$ is the relative change in \%.}
\label{tab:cls_before_after}
\end{table}
\section{Impact on Task Classification Accuracy}
\label{app:cls_degradation}
Aligning a model to flip its own predictions could in principle distort the decision boundary that the counterfactuals are meant to explain. To assess this, we evaluate the models trained with \our on the original test set, and compare them with the corresponding base models on the same inputs. Table~\ref{tab:cls_before_after} indicates no systematic degradation. Averaged over all models, datasets, and languages, accuracy increases by $2.7\%$ in relative terms. \lm{Gemma3} family models change negligibly, whereas \lm{Qwen3-4B} and \lm{Qwen3-8B} improve by $6.8\%$ and $9.3\%$ on \data{TAXI1500}. The decision boundary that \our explains is therefore preserved rather than distorted, and in some settings even refined.

\end{document}